\documentclass[10pt,twocolumn,letterpaper]{article}

\usepackage{cvpr}              

\usepackage[accsupp]{axessibility}
\usepackage{pgf}
\usepackage{multirow}
\usepackage{pifont}
\usepackage{pgfplots}
\pgfplotsset{compat=1.17}
\usepackage{pgfplotstable}
\usepackage{tikz}
\usepackage{csvsimple}
\usepackage{graphicx}
\usepackage{xifthen}
\usepackage{arydshln}
\usepackage{adjustbox}
\usepackage{balance}
\usepackage{tcolorbox}
\usepackage{etoolbox}
\usetikzlibrary{shapes,calc,positioning, decorations.text, arrows.meta, calc, shadows.blur, shadings, fit}

\definecolor{cvprblue}{rgb}{0.21,0.49,0.74}
\usepackage[pagebackref,breaklinks,colorlinks,allcolors=cvprblue]{hyperref}

\def\paperID{10138} 
\def\confName{CVPR}
\def\confYear{2025}

\title{ILIAS: Instance-Level Image retrieval At Scale}

\author{Giorgos Kordopatis-Zilos \quad Vladan Stojni\'{c} \quad Anna Manko \quad Pavel \v{S}uma  \\ Nikolaos-Antonios Ypsilantis \quad Nikos Efthymiadis \quad Zakaria Laskar \\ Ji\v{r}\'{i} Matas \quad Ond\v{r}ej Chum \quad Giorgos Tolias \\ \\
VRG, FEE, Czech Technical University in Prague \vspace{-7pt}
}

\begin{document}

\maketitle

\def\Lqtrain{\ensuremath{L_q^{\text{\tiny \raisebox{2pt}{train}}}}\xspace}
\def\Lxtrain{\ensuremath{L_x^{\text{\tiny \raisebox{2pt}{train}}}}\xspace}
\def\Lqtest{\ensuremath{L_q^{\text{\tiny \raisebox{2pt}{test}}}}\xspace}
\def\Lxtest{\ensuremath{L_x^{\text{\tiny \raisebox{2pt}{test}}}}\xspace}
\def\Lxasmk{\ensuremath{L^{\text{\tiny \raisebox{2pt}{asmk}}}_x}\xspace}

\newcommand{\ione}{i\hspace{-.05em}+\hspace{-.07em}1}

\newcommand{\mypartight}[1]{\noindent {\bf #1}}
\newcommand{\myparagraph}[1]{\vspace{3pt}\noindent\textbf{#1}\xspace}

\newcommand{\optional}[1]{{#1}}
\newcommand{\alert}[1]{{\color{red}{#1}}}
\newcommand{\gt}[1]{{\color{purple}{GT: #1}}}
\newcommand{\gtt}[1]{{\color{purple}{#1}}}
\newcommand{\gtr}[2]{{\color{purple}\st{#1} {#2}}}

\newcommand{\gkz}[1]{{\color{cyan}{GKZ: #1}}}
\newcommand{\gkzt}[1]{{\color{cyan}{#1}}}
\newcommand{\gkzr}[2]{{\color{cyan}\st{#1} {#2}}}

\newcommand{\ps}[1]{{\color{brown}{PS: #1}}}
\newcommand{\pst}[1]{{\color{brown}{#1}}}
\newcommand{\psr}[2]{{\color{brown}\st{#1} {#2}}}

\newcommand{\am}[1]{{\color{orange}{AM: #1}}}
\newcommand{\amt}[1]{{\color{orange}{#1}}}
\newcommand{\amr}[2]{{\color{orange}\st{#1} {#2}}}

\newcommand{\och}[1]{{\color{blue}{OCh: #1}}}
\newcommand{\ocht}[1]{{\color{blue}{#1}}}
\newcommand{\ochr}[2]{{\color{blue}\st{#1} {#2}}}

\newcommand{\gray}[1]{{\color{gray}{#1}}}

\definecolor{appleblue}{RGB}{0,122,255}

\newcolumntype{Y}{>{\centering\arraybackslash}p{4em}}

\def\roxf{$\mathcal{R}$Oxford\xspace}
\def\rox{$\mathcal{R}$Oxf\xspace}
\def\ro{$\mathcal{R}$O\xspace}
\def\rpar{$\mathcal{R}$Paris\xspace}
\def\rpa{$\mathcal{R}$Par\xspace}
\def\rp{$\mathcal{R}$P\xspace}
\def\rdis{$\mathcal{R}$1M\xspace}

\newcommand\resnet[3]{\ensuremath{\prescript{#2}{}{\mathtt{R}}{#1}_{\scriptscriptstyle #3}}\xspace}

\newcommand{\ours}{\mbox{ILIAS}\xspace} %
\newcommand{\miniours}{\textit{mini}-ILIAS\xspace} %

\newcommand{\stddev}[1]{\scriptsize{$\pm#1$}}

\newcommand{\diffup}[1]{{\color{OliveGreen}{($\uparrow$ #1)}}}
\newcommand{\diffdown}[1]{{\color{BrickRed}{($\downarrow$ #1)}}}

\def\nmsp{\hspace{-6pt}}
\def\nssp{\hspace{-3pt}}
\def\nxssp{\hspace{-1pt}}
\def\zsp{\hspace{0pt}}
\def\xssp{\hspace{1pt}}
\def\ssp{\hspace{3pt}}
\def\msp{\hspace{6pt}}
\def\mlsp{\hspace{9pt}}
\def\lsp{\hspace{12pt}}
\def\xlsp{\hspace{20pt}}

\newcommand{\head}[1]{{\smallskip\noindent\bf #1}}
\newcommand{\equ}[1]{(\ref{equ:#1})\xspace}

\newcommand{\nn}[1]{\ensuremath{\text{NN}_{#1}}\xspace}
\def\l1{\ensuremath{\ell_1}\xspace}
\def\l2{\ensuremath{\ell_2}\xspace}

\newcommand{\tran}{^\top}
\newcommand{\mtran}{^{-\top}}
\newcommand{\zcol}{\mathbf{0}}
\newcommand{\zrow}{\zcol\tran}

\newcommand{\ind}{\mathds{1}}
\newcommand{\expect}{\mathbb{E}}
\newcommand{\nat}{\mathbb{N}}
\newcommand{\zahl}{\mathbb{Z}}
\newcommand{\real}{\mathbb{R}}
\newcommand{\proj}{\mathbb{P}}
\newcommand{\prob}{\mathbf{Pr}}

\newcommand{\mif}{\textrm{if }}
\newcommand{\other}{\textrm{otherwise}}
\newcommand{\minimize}{\textrm{minimize }}
\newcommand{\maximize}{\textrm{maximize }}

\newcommand{\id}{\operatorname{id}}
\newcommand{\const}{\operatorname{const}}
\newcommand{\sgn}{\operatorname{sgn}}
\newcommand{\erf}{\operatorname{erf}}
\newcommand{\var}{\operatorname{Var}}
\newcommand{\mean}{\operatorname{mean}}
\newcommand{\trace}{\operatorname{tr}}
\newcommand{\diag}{\operatorname{diag}}
\newcommand{\vect}{\operatorname{vec}}
\newcommand{\cov}{\operatorname{cov}}

\newcommand{\softmax}{\operatorname{softmax}}
\newcommand{\clip}{\operatorname{clip}}

\newcommand{\defn}{\mathrel{:=}}
\newcommand{\peq}{\mathrel{+\!=}}
\newcommand{\meq}{\mathrel{-\!=}}

\newcommand{\floor}[1]{\left\lfloor{#1}\right\rfloor}
\newcommand{\ceil}[1]{\left\lceil{#1}\right\rceil}
\newcommand{\inner}[1]{\left\langle{#1}\right\rangle}
\newcommand{\norm}[1]{\left\|{#1}\right\|}
\newcommand{\frob}[1]{\norm{#1}_F}
\newcommand{\card}[1]{\left|{#1}\right|\xspace}
\newcommand{\diff}{\mathrm{d}}
\newcommand{\der}[3][]{\frac{d^{#1}#2}{d#3^{#1}}}
\newcommand{\pder}[3][]{\frac{\partial^{#1}{#2}}{\partial{#3^{#1}}}}
\newcommand{\ipder}[3][]{\partial^{#1}{#2}/\partial{#3^{#1}}}
\newcommand{\dder}[3]{\frac{\partial^2{#1}}{\partial{#2}\partial{#3}}}

\newcommand{\wb}[1]{\overline{#1}}
\newcommand{\wt}[1]{\widetilde{#1}}

\newcommand{\cA}{\mathcal{A}}
\newcommand{\cB}{\mathcal{B}}
\newcommand{\cC}{\mathcal{C}}
\newcommand{\cD}{\mathcal{D}}
\newcommand{\cE}{\mathcal{E}}
\newcommand{\cF}{\mathcal{F}}
\newcommand{\cG}{\mathcal{G}}
\newcommand{\cH}{\mathcal{H}}
\newcommand{\cI}{\mathcal{I}}
\newcommand{\cJ}{\mathcal{J}}
\newcommand{\cK}{\mathcal{K}}
\newcommand{\cL}{\mathcal{L}}
\newcommand{\cM}{\mathcal{M}}
\newcommand{\cN}{\mathcal{N}}
\newcommand{\cO}{\mathcal{O}}
\newcommand{\cP}{\mathcal{P}}
\newcommand{\cQ}{\mathcal{Q}}
\newcommand{\cR}{\mathcal{R}}
\newcommand{\cS}{\mathcal{S}}
\newcommand{\cT}{\mathcal{T}}
\newcommand{\cU}{\mathcal{U}}
\newcommand{\cV}{\mathcal{V}}
\newcommand{\cW}{\mathcal{W}}
\newcommand{\cX}{\mathcal{X}}
\newcommand{\cY}{\mathcal{Y}}
\newcommand{\cZ}{\mathcal{Z}}

\newcommand{\vA}{\mathbf{A}}
\newcommand{\vB}{\mathbf{B}}
\newcommand{\vC}{\mathbf{C}}
\newcommand{\vD}{\mathbf{D}}
\newcommand{\vE}{\mathbf{E}}
\newcommand{\vF}{\mathbf{F}}
\newcommand{\vG}{\mathbf{G}}
\newcommand{\vH}{\mathbf{H}}
\newcommand{\vI}{\mathbf{I}}
\newcommand{\vJ}{\mathbf{J}}
\newcommand{\vK}{\mathbf{K}}
\newcommand{\vL}{\mathbf{L}}
\newcommand{\vM}{\mathbf{M}}
\newcommand{\vN}{\mathbf{N}}
\newcommand{\vO}{\mathbf{O}}
\newcommand{\vP}{\mathbf{P}}
\newcommand{\vQ}{\mathbf{Q}}
\newcommand{\vR}{\mathbf{R}}
\newcommand{\vS}{\mathbf{S}}
\newcommand{\vT}{\mathbf{T}}
\newcommand{\vU}{\mathbf{U}}
\newcommand{\vV}{\mathbf{V}}
\newcommand{\vW}{\mathbf{W}}
\newcommand{\vX}{\mathbf{X}}
\newcommand{\vY}{\mathbf{Y}}
\newcommand{\vZ}{\mathbf{Z}}

\newcommand{\va}{\mathbf{a}}
\newcommand{\vb}{\mathbf{b}}
\newcommand{\vc}{\mathbf{c}}
\newcommand{\vd}{\mathbf{d}}
\newcommand{\ve}{\mathbf{e}}
\newcommand{\vf}{\mathbf{f}}
\newcommand{\vg}{\mathbf{g}}
\newcommand{\vh}{\mathbf{h}}
\newcommand{\vi}{\mathbf{i}}
\newcommand{\vj}{\mathbf{j}}
\newcommand{\vk}{\mathbf{k}}
\newcommand{\vl}{\mathbf{l}}
\newcommand{\vm}{\mathbf{m}}
\newcommand{\vn}{\mathbf{n}}
\newcommand{\vo}{\mathbf{o}}
\newcommand{\vp}{\mathbf{p}}
\newcommand{\vq}{\mathbf{q}}
\newcommand{\vr}{\mathbf{r}}
\newcommand{\Vs}{\mathbf{s}}
\newcommand{\vt}{\mathbf{t}}
\newcommand{\vu}{\mathbf{u}}
\newcommand{\vv}{\mathbf{v}}
\newcommand{\vw}{\mathbf{w}}
\newcommand{\vx}{\mathbf{x}}
\newcommand{\vy}{\mathbf{y}}
\newcommand{\vz}{\mathbf{z}}

\newcommand{\vone}{\mathbf{1}}
\newcommand{\vzero}{\mathbf{0}}

\newcommand{\valpha}{{\boldsymbol{\alpha}}}
\newcommand{\vbeta}{{\boldsymbol{\beta}}}
\newcommand{\vgamma}{{\boldsymbol{\gamma}}}
\newcommand{\vdelta}{{\boldsymbol{\delta}}}
\newcommand{\vepsilon}{{\boldsymbol{\epsilon}}}
\newcommand{\vzeta}{{\boldsymbol{\zeta}}}
\newcommand{\veta}{{\boldsymbol{\eta}}}
\newcommand{\vtheta}{{\boldsymbol{\theta}}}
\newcommand{\viota}{{\boldsymbol{\iota}}}
\newcommand{\vkappa}{{\boldsymbol{\kappa}}}
\newcommand{\vlambda}{{\boldsymbol{\lambda}}}
\newcommand{\vmu}{{\boldsymbol{\mu}}}
\newcommand{\vnu}{{\boldsymbol{\nu}}}
\newcommand{\vxi}{{\boldsymbol{\xi}}}
\newcommand{\vomikron}{{\boldsymbol{\omikron}}}
\newcommand{\vpi}{{\boldsymbol{\pi}}}
\newcommand{\vrho}{{\boldsymbol{\rho}}}
\newcommand{\vsigma}{{\boldsymbol{\sigma}}}
\newcommand{\vtau}{{\boldsymbol{\tau}}}
\newcommand{\vupsilon}{{\boldsymbol{\upsilon}}}
\newcommand{\vphi}{{\boldsymbol{\phi}}}
\newcommand{\vchi}{{\boldsymbol{\chi}}}
\newcommand{\vpsi}{{\boldsymbol{\psi}}}
\newcommand{\vomega}{{\boldsymbol{\omega}}}

\newcommand{\rLambda}{\mathrm{\Lambda}}
\newcommand{\rSigma}{\mathrm{\Sigma}}

\makeatletter
\DeclareRobustCommand\onedot{\futurelet\@let@token\@onedot}
\def\@onedot{\ifx\@let@token.\else.\null\fi\xspace}
\def\eg{\emph{e.g}\onedot} \def\Eg{\emph{E.g}\onedot}
\def\ie{\emph{i.e}\onedot} \def\Ie{\emph{I.e}\onedot}
\def\vs{\emph{vs\onedot}}
\def\cf{\emph{cf}\onedot} \def\Cf{\emph{C.f}\onedot}
\def\etc{\emph{etc}\onedot} \def\vs{\emph{vs}\onedot}
\def\wrt{w.r.t\onedot} \def\dof{d.o.f\onedot}
\def\etal{\emph{et al}\onedot}
\makeatother

\newcommand\rurl[1]{%
  \href{https://#1}{\nolinkurl{#1}}%
}

\newcommand{\bentarrow}[1][]{%
  \begin{tikzpicture}[#1]%
    \draw (0,0.7ex) -- (0,0) -- (0.75em,0);
    \draw (0.55em,0.2em) -- (0.75em,0) -- (0.55em,-0.2em);
  \end{tikzpicture}%
}

\definecolor{higha}{HTML}{009b10} 
\definecolor{lowa}{HTML}{ec462e}  
\definecolor{mediuma}{HTML}{FFFFFF} 

\newcommand*{\opacitya}{50} 
\newcommand*{\minvalcolora}{2.5} 
\newcommand*{\midvalcolora}{11.6} 
\newcommand*{\maxvalcolora}{37.3} 
\newcommand{\grca}[1]{
    \ifdim #1pt < \midvalcolora pt
        \pgfmathparse{(#1-\minvalcolora)/(\midvalcolora-\minvalcolora)}
        \let\normalizedval\pgfmathresult
    
        \pgfmathparse{100*(\normalizedval)^(2.0)} 
        \xdef\tempa{\pgfmathresult}
        \pgfmathparse{min(100,max(0,\tempa))}
        \xdef\tempa{\pgfmathresult}
    
        \cellcolor{mediuma!\tempa!lowa!\opacitya} #1
    \else
        \pgfmathparse{(#1-\midvalcolora)/(\maxvalcolora-\midvalcolora)}
        \let\normalizedval\pgfmathresult
    
        \pgfmathparse{100*(\normalizedval)^(2.0)}
        \xdef\tempa{\pgfmathresult}
        \pgfmathparse{min(100,max(0,\tempa))}
        \xdef\tempa{\pgfmathresult}
    
        \cellcolor{higha!\tempa!mediuma!\opacitya} #1
    \fi
}

\definecolor{highb}{HTML}{009b10} 
\definecolor{lowb}{HTML}{ec462e}  
\definecolor{mediumb}{HTML}{FFFFFF} 

\newcommand*{\opacityb}{50} 
\newcommand*{\minvalcolorb}{1.8} 
\newcommand*{\midvalcolorb}{8.6} 
\newcommand*{\maxvalcolorb}{31.3} 
\newcommand{\grcb}[1]{
    \ifdim #1pt < \midvalcolorb pt
        \pgfmathparse{(#1-\minvalcolorb)/(\midvalcolorb-\minvalcolorb)}
        \let\normalizedval\pgfmathresult
    
        \pgfmathparse{100*(\normalizedval)^(2.0)} 
        \xdef\tempa{\pgfmathresult}
        \pgfmathparse{min(100,max(0,\tempa))}
        \xdef\tempa{\pgfmathresult}
    
        \cellcolor{mediumb!\tempa!lowb!\opacityb} #1
    \else
        \pgfmathparse{(#1-\midvalcolorb)/(\maxvalcolorb-\midvalcolorb)}
        \let\normalizedval\pgfmathresult
    
        \pgfmathparse{100*(\normalizedval)^(2.0)}
        \xdef\tempa{\pgfmathresult}
        \pgfmathparse{min(100,max(0,\tempa))}
        \xdef\tempa{\pgfmathresult}
    
        \cellcolor{highb!\tempa!mediumb!\opacityb} #1
    \fi
}

\definecolor{highc}{HTML}{009b10} 
\definecolor{lowc}{HTML}{ec462e}  
\definecolor{mediumc}{HTML}{FFFFFF} 

\newcommand*{\opacityc}{50} 
\newcommand*{\minvalcolorc}{1.7} 
\newcommand*{\midvalcolorc}{5.9} 
\newcommand*{\maxvalcolorc}{20.8} 
\newcommand{\grcc}[1]{
    \ifdim #1pt < \midvalcolorc pt
        \pgfmathparse{(#1-\minvalcolorc)/(\midvalcolorc-\minvalcolorc)}
        \let\normalizedval\pgfmathresult
    
        \pgfmathparse{100*(\normalizedval)^(2.0)} 
        \xdef\tempa{\pgfmathresult}
        \pgfmathparse{min(100,max(0,\tempa))}
        \xdef\tempa{\pgfmathresult}
    
        \cellcolor{mediumc!\tempa!lowc!\opacityc} #1
    \else
        \pgfmathparse{(#1-\midvalcolorc)/(\maxvalcolorc-\midvalcolorc)}
        \let\normalizedval\pgfmathresult
    
        \pgfmathparse{100*(\normalizedval)^(2.0)}
        \xdef\tempa{\pgfmathresult}
        \pgfmathparse{min(100,max(0,\tempa))}
        \xdef\tempa{\pgfmathresult}
    
        \cellcolor{highc!\tempa!mediumc!\opacityc} #1
    \fi
}

\definecolor{highd}{HTML}{009b10} 
\definecolor{lowd}{HTML}{ec462e}  
\definecolor{mediumd}{HTML}{FFFFFF} 

\newcommand*{\opacityd}{50} 
\newcommand*{\minvalcolord}{1.5} 
\newcommand*{\midvalcolord}{7.5} 
\newcommand*{\maxvalcolord}{19.8} 
\newcommand{\grcd}[1]{
    \ifdim #1pt < \midvalcolord pt
        \pgfmathparse{(#1-\minvalcolord)/(\midvalcolord-\minvalcolord)}
        \let\normalizedval\pgfmathresult
    
        \pgfmathparse{100*(\normalizedval)^(2.0)} 
        \xdef\tempa{\pgfmathresult}
        \pgfmathparse{min(100,max(0,\tempa))}
        \xdef\tempa{\pgfmathresult}
    
        \cellcolor{mediumd!\tempa!lowd!\opacityd} #1
    \else
        \pgfmathparse{(#1-\midvalcolord)/(\maxvalcolord-\midvalcolord)}
        \let\normalizedval\pgfmathresult
    
        \pgfmathparse{100*(\normalizedval)^(2.0)}
        \xdef\tempa{\pgfmathresult}
        \pgfmathparse{min(100,max(0,\tempa))}
        \xdef\tempa{\pgfmathresult}
    
        \cellcolor{highd!\tempa!mediumd!\opacityd} #1
    \fi
}

\definecolor{highe}{HTML}{009b10} 
\definecolor{lowe}{HTML}{ec462e}  
\definecolor{mediume}{HTML}{FFFFFF} 

\newcommand*{\opacitye}{50} 
\newcommand*{\minvalcolore}{2.3} 
\newcommand*{\midvalcolore}{10.6} 
\newcommand*{\maxvalcolore}{24.7} 
\newcommand{\grce}[1]{
    \ifdim #1pt < \midvalcolore pt
        \pgfmathparse{(#1-\minvalcolore)/(\midvalcolore-\minvalcolore)}
        \let\normalizedval\pgfmathresult
    
        \pgfmathparse{100*(\normalizedval)^(2.0)} 
        \xdef\tempa{\pgfmathresult}
        \pgfmathparse{min(100,max(0,\tempa))}
        \xdef\tempa{\pgfmathresult}
    
        \cellcolor{mediume!\tempa!lowe!\opacitye} #1
    \else
        \pgfmathparse{(#1-\midvalcolore)/(\maxvalcolore-\midvalcolore)}
        \let\normalizedval\pgfmathresult
    
        \pgfmathparse{100*(\normalizedval)^(2.0)}
        \xdef\tempa{\pgfmathresult}
        \pgfmathparse{min(100,max(0,\tempa))}
        \xdef\tempa{\pgfmathresult}
    
        \cellcolor{highe!\tempa!mediume!\opacitye} #1
    \fi
}

\newcommand{\amescol}{cyan}
\newcommand{\sglipcol}{OliveGreen}
\newcommand{\dinocol}{BrickRed}
\newcommand{\globalcol}{BrickRed}

\definecolor{lightred}{RGB}{231, 76, 60}   
\definecolor{lightblue}{RGB}{54, 69, 79}
\definecolor{lightgreen}{RGB}{50, 205, 50} 

\definecolor{appleblue}{RGB}{80,122,255}  
\definecolor{applepink}{RGB}{217,127,174}  
\definecolor{appleteal}{RGB}{85,163,152}  
\definecolor{applepurple}{RGB}{138,107,221}  
\definecolor{applemustard}{RGB}{234,179,77}  
\definecolor{appleorange}{RGB}{255,127,80}  
\definecolor{applegreen}{RGB}{52,199,89}
\definecolor{appleyellow}{RGB}{255,204,0}

\definecolor{cvprblue}{rgb}{0.21,0.49,0.74}
\definecolor{materialPurple}{rgb}{0.615, 0.275, 1} 
\definecolor{plotlyBlue}{rgb}{0, 0.482, 1} 
\definecolor{plotlyGreen}{rgb}{0.157, 0.655, 0.271} 
\definecolor{plotlyRed}{rgb}{0.863, 0.208, 0.271} 
\definecolor{plotlyOrange}{rgb}{1, 0.757, 0.027} 
\definecolor{plotlyYellow}{rgb}{1, 0.843, 0} 
\definecolor{plotlyCyan}{rgb}{0.09, 0.745, 0.812} 
\definecolor{plotlyMagenta}{rgb}{0.917, 0.722, 0.894} 
\definecolor{plotlyTeal}{rgb}{0, 0.545, 0.455} 
\definecolor{plotlyNavy}{rgb}{0.4, 0.063, 0.949} 
\definecolor{plotlyPurple}{rgb}{0.58, 0.404, 0.741} 
\definecolor{plotlyBrown}{rgb}{0.82, 0.604, 0.416} 
\definecolor{plotlyPink}{rgb}{0.89, 0.467, 0.761} 
\definecolor{plotlyGray}{rgb}{0.498, 0.498, 0.498} 
\definecolor{plotlyLightBlue}{rgb}{0.122, 0.467, 0.706} 
\definecolor{plotlyLightOrange}{rgb}{1, 0.733, 0.471} 
\definecolor{plotlyLightGreen}{rgb}{0.596, 0.875, 0.541} 

\definecolor{calendarPurple}{rgb}{0.635, 0.349, 1.000} 
\definecolor{calendarRed}{rgb}{1.000, 0.420, 0.420} 
\definecolor{calendarPinkRed}{rgb}{1.000, 0.396, 0.518} 
\definecolor{calendarLightPink}{rgb}{1.000, 0.663, 0.663} 
\definecolor{calendarLightPurple}{rgb}{0.847, 0.627, 1.000} 
\definecolor{calendarDarkBlue}{rgb}{0.424, 0.435, 1.000} 
\definecolor{calendarBlue}{rgb}{0.000, 0.620, 1.000} 
\definecolor{calendarGreen}{rgb}{0.000, 0.714, 0.427} 
\definecolor{calendarLightMint}{rgb}{0.678, 0.910, 0.827} 
\definecolor{calendarYellow}{rgb}{1.000, 0.922, 0.231} 
\definecolor{calendarAmber}{rgb}{1.000, 0.757, 0.027} 
\definecolor{calendarOrange}{rgb}{1.000, 0.596, 0.000} 
\definecolor{calendarBrown}{rgb}{0.490, 0.353, 0.314} 
\definecolor{calendarGray}{rgb}{0.710, 0.710, 0.710} 
\definecolor{calendarGrayBlue}{rgb}{0.482, 0.604, 0.631} 
\definecolor{calendarLightBlue}{rgb}{0.647, 0.847, 0.867} 
\definecolor{calendarDarkPurple}{rgb}{0.416, 0.051, 0.678} 

\pgfplotsset{
  colormap/magma/.style={%
    /pgfplots/colormap={magma}{%
      rgb=(0.001462, 0.000466, 0.013866)
      rgb=(0.035520, 0.028397, 0.125209)
      rgb=(0.102815, 0.063010, 0.257854)
      rgb=(0.191460, 0.064818, 0.396152)
      rgb=(0.291366, 0.064553, 0.475462)
      rgb=(0.384299, 0.097855, 0.501002)
      rgb=(0.475780, 0.134577, 0.507921)
      rgb=(0.569172, 0.167454, 0.504105)
      rgb=(0.664915, 0.198075, 0.488836)
      rgb=(0.761077, 0.231214, 0.460162)
      rgb=(0.852126, 0.276106, 0.418573)
      rgb=(0.925937, 0.346844, 0.374959)
      rgb=(0.969680, 0.446936, 0.360311)
      rgb=(0.989363, 0.557873, 0.391671)
      rgb=(0.996580, 0.668256, 0.456192)
      rgb=(0.996727, 0.776795, 0.541039)
      rgb=(0.992440, 0.884330, 0.640099)
      rgb=(0.987053, 0.991438, 0.749504)
    },
  },
  colormap/inferno/.style={%
    /pgfplots/colormap={inferno}{%
      rgb=(0.001462, 0.000466, 0.013866)
      rgb=(0.037668, 0.025921, 0.132232)
      rgb=(0.116656, 0.047574, 0.272321)
      rgb=(0.217949, 0.036615, 0.383522)
      rgb=(0.316282, 0.053490, 0.425116)
      rgb=(0.410113, 0.087896, 0.433098)
      rgb=(0.503493, 0.121575, 0.423356)
      rgb=(0.596940, 0.154848, 0.398125)
      rgb=(0.688653, 0.192239, 0.357603)
      rgb=(0.775059, 0.239667, 0.303526)
      rgb=(0.851384, 0.302260, 0.239636)
      rgb=(0.912966, 0.381636, 0.169755)
      rgb=(0.956852, 0.475356, 0.094695)
      rgb=(0.981895, 0.579392, 0.026250)
      rgb=(0.987464, 0.690366, 0.079990)
      rgb=(0.973088, 0.805409, 0.216877)
      rgb=(0.947594, 0.917399, 0.410665)
      rgb=(0.988362, 0.998364, 0.644924)
    },
  },
  colormap/plasma/.style={%
    /pgfplots/colormap={plasma}{%
      rgb=(0.050383, 0.029803, 0.527975)
      rgb=(0.186213, 0.018803, 0.587228)
      rgb=(0.287076, 0.010855, 0.627295)
      rgb=(0.381047, 0.001814, 0.653068)
      rgb=(0.471457, 0.005678, 0.659897)
      rgb=(0.557243, 0.047331, 0.643443)
      rgb=(0.636008, 0.112092, 0.605205)
      rgb=(0.706178, 0.178437, 0.553657)
      rgb=(0.768090, 0.244817, 0.498465)
      rgb=(0.823132, 0.311261, 0.444806)
      rgb=(0.872303, 0.378774, 0.393355)
      rgb=(0.915471, 0.448807, 0.342890)
      rgb=(0.951344, 0.522850, 0.292275)
      rgb=(0.977856, 0.602051, 0.241387)
      rgb=(0.992541, 0.687030, 0.192170)
      rgb=(0.992505, 0.777967, 0.152855)
      rgb=(0.974443, 0.874622, 0.144061)
      rgb=(0.940015, 0.975158, 0.131326)
    },
  },
}

\begin{abstract}
This work introduces \ours, a new test dataset for Instance-Level Image retrieval At Scale.
It is designed to evaluate the ability of current and future foundation models and retrieval techniques to recognize particular objects.
The key benefits over existing datasets include large scale, domain diversity, accurate ground truth, and a performance that is far from saturated.
\ours includes query and positive images for 1,000 object instances, manually collected to capture challenging conditions and diverse domains. Large-scale retrieval is conducted against 100 million distractor images from YFCC100M. To avoid false negatives without extra annotation effort, we include only query objects confirmed to have emerged after 2014, \ie the compilation date of YFCC100M.
An extensive benchmarking is performed with the following observations: 
i) models fine-tuned on specific domains, such as landmarks or products, excel in that domain but fail on \ours
ii) learning a linear adaptation layer using multi-domain class supervision results in  performance improvements, especially for vision-language models
iii) local descriptors in retrieval re-ranking are still a key ingredient, especially in the presence of severe background clutter
iv) the text-to-image performance of the vision-language foundation models is surprisingly close to the corresponding image-to-image case. \\
website: \url{https://vrg.fel.cvut.cz/ilias/}
\end{abstract}
\vspace{-15pt}

\section{Introduction}
The ability to recognize and differentiate every unique object instance in the physical world represents one of the ultimate goals for foundation representation models~\cite{clip,bha+,odm+24,siglip}. 
This work aims to assess this capability through the lens of instance-level image retrieval at a very large scale.
Instance-level image retrieval corresponds to searching for images of particular objects within large collections. 
All images of a particular object form their own instance-level class.
This is an important information retrieval task due to its numerous real-world applications in robotics~\cite{lcj23,sth+24}, e-commerce~\cite{eproduct, zwd+21}, and cultural heritage~\cite{sll+21,ff22}, to name just a few. 
The task faces challenges because of the substantial variations among positive examples, such as illumination/viewpoint~\cite{jc19,tas+15} changes and background clutter~\cite{mbe18,bad+21}.
An additional difficulty is the high similarity among negatives, which is driven by the extremely fine granularity in the class definitions. 
It becomes even more challenging at a real-world scale, where searching through millions or even billions of images requires handling an open-world setup with countless unseen objects spanning diverse and complex domains.

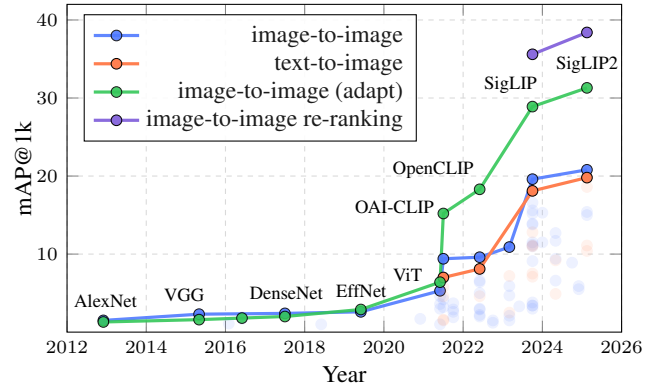
\begin{figure}[t]
  \centering
  \pgfplotsset{every tick label/.append style={font=\tiny}}
\begin{tikzpicture}[define rgb/.code={\definecolor{mycolor}{RGB}{#1}}, rgb color/.style={define rgb={#1},mycolor}]
    \hspace{-3pt}
    \begin{axis}[
        width=1.07\linewidth,
        height=0.71\linewidth,
        xmin=2012.0,
        xmax=2026.0,
        ymin=0,
        ymax=42,
        grid=both,
        grid style={color=lightgray!60, dash pattern=on 2pt off 2pt},
        ytick={10, 20, 30, 40},
        ylabel = {\small mAP@1k},
        xlabel = {\small Year},
        ylabel style= {yshift=-5pt},
        xlabel style= {yshift=2pt},
        yticklabel style={font=\scriptsize},
        xticklabel style=
    {/pgf/number format/1000 sep=,anchor=north, font=\scriptsize},
        legend pos=north west,
        legend style={cells={anchor=east}, font=\small, fill opacity=0.9, row sep=-1pt},
    ]
    
        \addplot[only marks, mark=*, opacity=0.1, mark size=2, color=appleblue, 
                forget plot] 
        table[x=date, y=yfcc100m] {./data/performance_year_original.csv};
        
        \addplot[mark=*, solid, opacity=0.9, mark size=2, line  width=1.2pt, color=appleblue,
            mark options={draw=black, line width=0.35pt},
            filter discard warning=false, 
            unbounded coords=discard,
            y filter/.expression={y < 0 ? nan : y},
            point meta=explicit symbolic, 
            ]
        table[x=date, y=sota, meta=model_short] {./data/performance_year_original.csv}; 
        \addlegendentry{image-to-image};
        
        \addplot[only marks, mark=*, opacity=0.1, mark size=2, color=appleorange, 
            forget plot] 
        table[x=date, y=yfcc100m] {./data/performance_year_text.csv};
        
        \addplot[mark=*, solid, opacity=1, mark size=2, line width=1.2pt, color=appleorange,
            mark options={draw=black, line width=0.35pt},
            filter discard warning=false, 
            unbounded coords=discard,
            y filter/.expression={y < 0 ? nan : y},
            point meta=explicit symbolic,]
        table[x=date, y=sota, meta=model_short] {./data/performance_year_text.csv}; 
        \addlegendentry{text-to-image};

        \addplot[mark=*, solid, opacity=0.9, mark size=2, line width=1.2pt, color=applegreen,
            mark options={draw=black, line width=0.35pt},
            filter discard warning=false, 
            unbounded coords=discard,
            y filter/.expression={y < 0 ? nan : y},
            point meta=explicit symbolic, 
            ]
        table[x=date, y=sota, meta=model_short] {./data/performance_year_proj.csv}; 
        \addlegendentry{image-to-image (adapt)};
        
        \addplot[mark=*, opacity=1, mark size=2, line width=1.2pt, color=applepurple, 
                mark options={draw=black, line width=0.35pt}
                ]
                coordinates {
                             (2023.75, 35.6)
                             (2025.125, 38.4)};
        \addlegendentry{image-to-image re-ranking};
        
        \node[anchor=south east, font=\scriptsize] at (axis cs:2014,2) {AlexNet};
        \node[anchor=south east, font=\scriptsize] at (axis cs:2015.7,2.8) {VGG};
        \node[anchor=south east, font=\scriptsize] at (axis cs:2018.7,3) {DenseNet};
        \node[anchor=south east, font=\scriptsize] at (axis cs:2020.3,3.3) {EffNet};
        \node[anchor=south east, font=\scriptsize] at (axis cs:2021.2,5.3) {ViT};
        \node[anchor=south east, font=\scriptsize] at (axis cs:2021.5,14) {OAI-CLIP};
        \node[anchor=south east, font=\scriptsize] at (axis cs:2022.5,18.5) {OpenCLIP};
        \node[anchor=south east, font=\scriptsize] at (axis cs:2024.1,29.5) {SigLIP};
        \node[anchor=south east, font=\scriptsize] at (axis cs:2026.15,32) {SigLIP2};
        \end{axis}
    \end{tikzpicture}
  \vspace{-24pt}
  \caption{\textbf{Performance timeline on \ours}. Curves indicate best performance in chronological order for {\color{appleblue}\textbf{image-to-image}} and {\color{appleorange}\textbf{text-to-image}} retrieval, showing a significant boost with the release of foundation models. Representations are {\color{applegreen}\textbf{linearly adapted}} via multi-domain learning on UnED~\cite{ycc+23}. {\color{applepurple}\textbf{Re-ranking with local descriptors}} achieves the best results by a significant margin.
  \label{fig:sota_scatter}
  \vspace{-15pt}
  }
\end{figure}

Benchmarking instance-level retrieval under real-world challenges is currently limited by the lack of suitable datasets. Constructing a dataset with instance-level class definitions necessitates huge development effort, reflected by the many shortcomings of existing datasets.
Shortcomings exist in several key aspects, such as dataset size~\cite{wj15}, domain diversity~\cite{rit+18,sxj+15}, and ground-truth accuracy~\cite{wac+20}, which suffers from both false positives and false negatives.
Popular datasets are typically limited to landmarks~\cite{rit+18}, and as dataset scale increases, ground-truth quality tends to decline~\cite{wac+20,syh+24}. 
This is a consequence of automating the ground-truth creation process to facilitate scaling up.
To address such limitations, we introduce the Instance-Level Image retrieval At Scale (\ours) evaluation dataset.

The creation of our dataset has two key elements. First, query and positive images are manually captured to ensure challenging variations, covering 1,000 objects across diverse domains. 
Second, to expand the dataset size without ground-truth errors or additional annotation effort, we leverage a key technique: distractor images, collected in 2014 from YFCC100M, are combined with query objects verified not to have publicly existed until after 2014. This distractor set includes 100 million images, two orders of magnitude larger than the largest existing dataset~\cite{rit+18}.
Notably, all images have a permissive license, allowing us to ensure long-term online availability to the full extent.

\ours includes both image and text queries. The latter is in the form of detailed descriptions of objects and their distinctive features. 
The dataset is designed to support future research in image-to-image and text-to-image retrieval for particular objects, and additionally serves as a large-scale benchmark for evaluating representations of foundation vision and language models (VLM)~\cite{clip,siglip}. 
To facilitate faster experimentation, we provide a mini, but challenging, version (5M) of the distractor set.

We perform an extensive evaluation comparison, including many foundation image-to-image and text-to-image models, and establish a comprehensive testbed that enables future comparisons.
The provided evaluation includes retrieval with global image representation but also re-ranking techniques that use local representations~\cite{ras+14,pci+07,ski+24} and query expansion~\cite{cps+07,rtc19}.
We observe the following:
\begin{itemize}
\item Performance of standard 10-year-old models, measured by mean Average Precision, is as low as 1.3\%, while the best-performing model achieves 31.3\%, as shown in Fig.~\ref{fig:sota_scatter}. This points out the vast progress of representation models and the high challenging factors of \ours.
\item VLMs are the top-performing models.
\item Smaller (ViT-B) models trained/tested on large resolution (512/724) outperform larger models (ViT-L) trained/tested on small resolution (256/384). 
\item Using Universal Embedding Dataset (UnED)~\cite{ycc+23} to learn a linear adaptation layer on top of frozen models improves performance of most models, making it a candidate training set to couple with \ours. Notably, VLMs demonstrate the largest benefits, presumably because their training stage does not optimize image-to-image relations. 
\item In contrast to the current belief~\cite{sck+23}, local representation is a key ingredient, while global representation, despite being efficient and compact, performs much lower. 
\item The performance gap between image-to-image and text-to-image models is surprisingly small. Therefore, detailed text queries are a reasonable proxy in the absence of image queries, even at the instance level.
\end{itemize}
\section{Related work}

\begin{table*}[t]
  \centering
  \vspace{-10pt}
  \def\sp{\hspace{9pt}}
\def\spp{\hspace{10pt}}
\scalebox{0.80}{
\begin{tabular}{l@{\sp}r@{\sp}r@{\sp}r@{\sp}r@{\sp}r@{\sp}r@{\sp}r@{\sp}r@{\sp}r@{\sp}r@{\sp}r@{\sp}r@{\sp}r@{\sp}}
    \toprule
    \multirow{1}{*}{\textbf{datasets}} & \multirow{1}{*}{\textbf{year}} & \multirow{1}{*}{\textbf{objects}} & \multirow{1}{*}{\textbf{query}} & \multirow{1}{*}{\textbf{positives}} &  \multirow{1}{*}{\textbf{database}}  & \multirow{1}{*}{\textbf{gt}} & \multirow{1}{*}{\textbf{class def.}} & \multirow{1}{*}{\textbf{domain}} & \multirow{1}{*}{\textbf{bbox}} & \multirow{1}{*}{\textbf{online}} & \multirow{1}{*}{\textbf{license}}   \\ 
    \midrule
    UKB~\cite{nister2006scalable}     & 2006 & 2.5K  & 10K   & 10K   & 10K   & FN    & IL         & product   & \ding{55}   & Fully     & N/A               \\
    Holidays~\cite{jed+08}            & 2008 & 500   & 500   & 991   & 1M    & Clean & IL         & landmark  & \ding{55}   & Partially & CC                \\
    Sculptures~\cite{az11}            & 2011 & 10    & 70    & 3.1K  & 3.1K  & Clean & Partial IL & sculpture & \ding{55}   & Partially & Flickr TC         \\
    INSTRE~\cite{wj15}                & 2015 & 200   & 1250  & 27.3K & 27.3K & Clean & Partial IL & multi     & \ding{51}   & Fully     & Flickr TC         \\
    SOP~\cite{sxj+15}                 & 2015 & 11.3K & 60.5K & 60.5K & 60.5K & Clean & Partial IL & product   & \ding{55}   & Fully     & MIT License       \\
    InShop~\cite{liu2016deepfashion}  & 2016 & 3.9K  & 14.2K & 12.6K & 12.6K & Clean & Partial IL & fashion   & \ding{51}   & Fully     & N/A               \\
    R-Oxford~\cite{rit+18}            & 2018 & 11	 & 70    & 5K    & 1M    & FN?   & IL         & landmark  & \ding{55}   & Fully     & Flickr TC, CC     \\
    R-Paris~\cite{rit+18}             & 2018 & 11    & 70    & 6.3K  & 1M    & FN?   & IL         & landmark  & \ding{55}   & Fully     & Flickr TC, CC     \\
    GLDv1~\cite{nas+17}               & 2018 & 30K   & N/A   & N/A   & 1.1M  & Clean & IL         & landmark  & \ding{55}   & Partially & Multiple          \\
    GLDv2~\cite{wac+20}               & 2020 & 318   & 1.1K  & 3.1K  & 762K  & FP    & IL         & landmark  & \ding{55}   & Fully     & CC/ Public-domain \\
    Product1M~\cite{zwd+21}           & 2021 & 392   & 6.5K  & 40K   & 40K   & Clean & Partial IL & product   & \ding{55}   & Partially & N/A               \\
    RP2K~\cite{rp2k}                  & 2021 & 1.2K  & 10.9K & 10.9K & 10.9K & Clean & Partial IL & product   & \ding{51}   & Fully     & N/A               \\
    GPR1200~\cite{gpr1200}            & 2021 & 1.2K  & 12K	 & 12K   & 12K   & Mix   & IL+FG      & multi     & \ding{55}   & Fully     & Multiple          \\
    eProduct~\cite{eproduct}          & 2021 & 206   & 10K   & N/A   & 1.1M  & N/A   & FG         & product   & \ding{55}   & No        & N/A               \\
    FORB~\cite{forb}                  & 2023 & N/A   & 13.9K & 4.5K  & 49.8K & Clean & IL         & planar    & \ding{55}   & Fully     & Snap Inc.         \\
    UnED~\cite{ycc+23}                & 2023 & 21K   & 241K  & 244K  & 1.4M  & Mix   & IL+FG      & multi     & \ding{55}   & Fully     & Multiple          \\ 
    \midrule
    \rowcolor{orange!70}\ours         & 2025 & 1,000 & 1,232 & 4,715 & 100M  & Clean & IL         & multi     & \ding{51}   & Fully     & CC                \\ 
    \bottomrule
\end{tabular}}

  \vspace{-7pt}
  \caption{\textbf{Comparison with other instance-level datasets.} Datasets are compared based on their size (object, query, positives, database), the accuracy of the ground truth (gt), type of class definition, domain, supplementary annotations (bbox) and accessibility (online, license). N/A: not available. FP/FN: false positives/negatives. FN?: possibility of false negatives. Mix: combination of clean and noisy datasets. IL: instance-level. FG: fine-grained. Partial IL: instance-level with subtle variations among same class objects. CC: Creative Commons.
  \label{tab:sota}
  \vspace{-13pt}
  }
\end{table*}

In this section, we review the related work in terms of existing datasets and benchmarks in the literature.

\noindent\textbf{Datasets. }
Tab.~\ref{tab:sota} presents the datasets from the image retrieval literature related to \ours. The datasets can be compared based on five main axes: (i) \emph{Class definition adopted}. Many datasets~\cite{nister2006scalable,jed+08,rit+18,nas+17,wac+20} adopt a strict definition very similar to ours, satisfying instance-level requirements. Others~\cite{sxj+15,wj15,az11,zwd+21} adopt a more relaxed definition, where some minor variations are permitted, \eg color changes in objects of the same class. Even more relaxed are the fine-grained definitions~\cite{eproduct}, where the object of the very same type is considered related, \eg same product with different variant. (ii) \emph{Domain of the dataset}. Most datasets are tailored for a specific domain. Landmarks are among the most popular domains~\cite{az11,jed+08,rit+18,nas+17,wac+20}. Other domains include products~\cite{nister2006scalable,zwd+21,rp2k} and fashion~\cite{sxj+15,liu2016deepfashion}. Some datasets cover multiple domains, either being standalone~\cite{wj15} or bundle of repurposed datasets~\cite{ycc+23,gpr1200}. (iii) \emph{Scale of database}. Most of the datasets are small-scale, counting a few thousand images~\cite{wj15,liu2016deepfashion,zwd+21}. Larger ones~\cite{rit+18,eproduct,ycc+23} expend slightly above a million. None satisfies large-scale requirements. (iv) \emph{Noise in ground truth}. Most datasets consist of clean annotations, except for a few cases that contain inaccuracies, including false positives~\cite{wac+20}, \ie images wrongly annotated as relevant, false negatives~\cite{nister2006scalable,rit+18}, \ie relevant images that have not been annotated as positives, or the possibility of false positives~\cite{rit+18}. (v) \emph{Availability}. Most datasets are publicly available with permissive licenses, with few exceptions of partial~\cite{az11,jed+08} or no~\cite{eproduct,nas+17} availability.
To this end, no publicly available dataset fits the strict instance-level definition, contains objects from multiple domains, ensures error-free labeling and is large scale. This gap is filled with \ours satisfying all the aforementioned requirements.

\noindent\textbf{Evaluation benchmarks.}
Benchmarking~\cite{sws+00} tracks the progress in the field, which is even more necessary with the emergence of foundation models.
Several benchmarks papers~\cite{zheng2017sift,kmo21,ko2019benchmark} exists in the instance-level retrieval literature, investigating the impact of learning scheme, post-processing, model ensembling, query expansion, and whitening.
The most relevant benchmark to \ours is UnED~\cite{ycc+23} that combines existing datasets to create a union that evaluates models performance across various domains.
Due to its wide variety, UnED serves as the training dataset for linear adaptation. 

Regarding the evaluation of foundation models, the most common practice~\cite{odm+24,fwx+23,evaclip} is measuring classification performance on top of frozen models on ImageNet~\cite{dds+09}. 
This is performed either with or without the training of a classifier via linear probing or k-NN search. 
Furthermore, models are usually evaluated on dense prediction tasks~\cite{odm+24} and several multiple-downstream single-domain tasks~\cite{ady+23}. For VLMs, zero-shot classification and retrieval serve as the primary benchmarks~\cite{clip,siglip,metaclip}, utilizing class text labels. 
In this work, we provide similar evaluation protocols tailored for instance-level retrieval. 
One can test the raw model capabilities or adapt for the instance-level task via linear adaptation on UnED.
Text-to-image retrieval is also facilitated for the evaluation of VLMs.

\section{ILIAS dataset}
\label{sec:ilias}

\begin{figure*}[t]
  \centering
  \scalebox{0.83}{
    \input{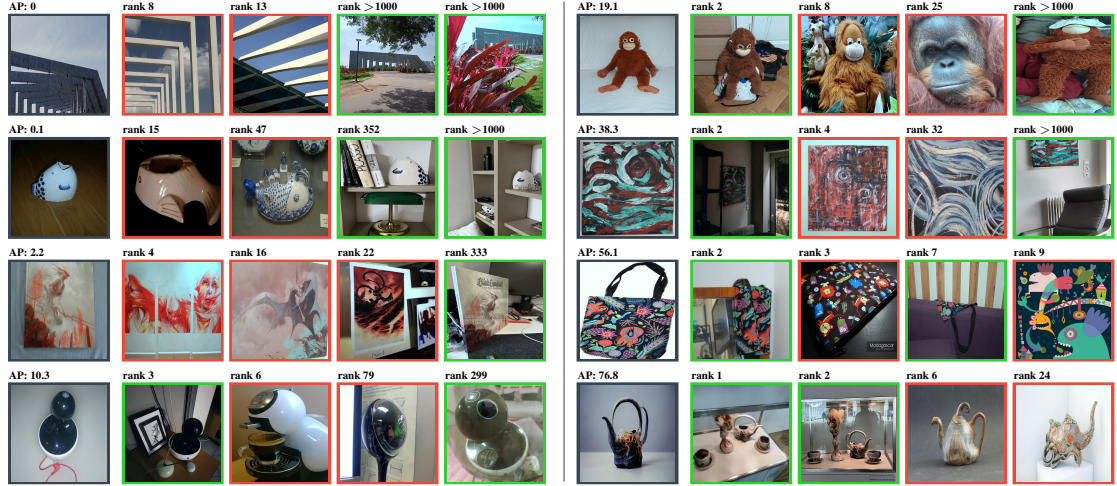}
  }
  \vspace{-21pt}
  \caption{\textbf{Examples of query, positive and hard negatives within the distractor set.} Average Precision per query and rank of the negatives and positives is reported using SigLIP~\cite{siglip} model. {\color{lightblue}\textbf{Gray:}} queries. {\color{lightgreen}\textbf{Green:}} positives. {\color{lightred}\textbf{Red:}} distractors.
  \label{fig:fig_dataset_instances}
  \vspace{-16pt}
  }
\end{figure*}

\subsection{Composition and collection}
\label{sec:composition}

\noindent{\textbf{Instance-level class definition.}} Following an instance-level class definition~\cite{rit+18,ygg+21}, we consider \emph{all indistinguishable object instances of the real world to form their own class}. 
Nevertheless, we add a restriction to consider a pair of images as relevant to each other only if there is a view overlap. Other cases are explicitly not included in the dataset, contrasting the existing work~\cite{liu2016deepfashion,sxj+15,wac+20}. Therefore, models should mostly rely on estimating the visual similarity and less on shortcuts through semantics.

\noindent\textbf{Overview.}
\ours supports both image-to-image (i2i) and text-to-image (t2i) retrieval and follows the standard setup for retrieval datasets, consisting of two main parts: (i) \emph{query} images and text, and (ii) \emph{database} (db) images. The objective is to rank \emph{positives} -- db images relevant to the query -- at the top ranks. 
The collected objects cover a wide range of categories and are not restricted to specific domains. An overview of some collected objects is provided in Fig.~\ref{fig:fig_dataset_instances}.

Queries and positives are created/collected by a group of \emph{collectors} that are well-informed about the task objectives. 
In addition to positives, in the database, we include numerous \emph{distractors} -- irrelevant (negative) images to the queries -- that make retrieval more challenging. Following previous work~\cite{rit+18}, adding a large, uncurated set of random images achieves this. The larger the set, the higher the chances of hard negatives -- images with similar appearance or semantics to the queries. To this end, we select the YFCC100M~\cite{tsf+16} dataset to serve as the source of distractors due to its size and permissive license.

\noindent\textbf{Selected objects.}
Ensuring that distractor images include no false negatives cannot be performed in a scalable or accurate way if one relies on human annotation or metadata. 
Instead, we take advantage of the fact that YFCC100M was crawled from Flickr in 2014.
Hence, an object qualifies in \ours if it could not have appeared on Flickr before 2014. To verify this, we rely either on publicly available information, \eg objects known to be created/manufactured after 2014, or on the collector's knowledge about the object not being publicly available. Additionally, we opt for objects with distinctive and unique features that set them apart from others within the same category. For example, we avoid recent smartphones that look like plain black screens or new objects with distinctive parts closely resembling older ones. \looseness=-1

\noindent\textbf{Queries and positives.}
Query images depict the instance on a clean or uniform background. When this is not feasible (\eg buildings or statues), background blurring or cropping is applied. This is performed to avoid including background objects in the query that do not have corresponding positives in our ground truth information. 
Positives are images featuring the query object in challenging conditions, such as clutter, scale changes, occlusions, and partial views. 
Prior work~\cite{rit+18} reveals that easy positives dominate performance metrics. Thus, we specifically opt for challenging cases that cannot be easily retrieved by the models.
To avoid taking advantage of camera identification, most query and positive images are captured with at least two different camera models to introduce diversity. We also incorporate older camera models that are used in YFCC100M.

Each text query consists of a detailed and fine-grained textual description of an object. 
Descriptions are initially created by a large language model prompted to provide highly detailed depictions of the object shown in query images. Generated descriptions are manually edited to fix errors, insufficient descriptions, or nuances of the model.

\noindent\textbf{Distractors.}
The YFCC100M dataset was chosen for the distractor set due to its large scale and diverse range of concepts. It consists of 100 million Flickr images, collected without specific filtering, aside from being shared under a permissive CC-BY license.

\noindent\textbf{Bounding box annotation.} We include supplementary bounding boxes that specify the precise location of objects in query and positive images. 
They provide statistics about the position and size of object areas, assist our analysis of the dataset challenges, and support future research in instance-level localization.

\noindent\textbf{Evaluation metric.}
Retrieval performance is evaluated via mean Average Precision (mAP), a widely used metric in instance-level image retrieval~\citep{rtc19,pci+07,pci+08}. Specifically, we adopt mAP@1k~\cite{wac+20}, which assesses the ranking of the top-1k nearest neighbors for each query, treating any positive not ranked among the top-1k as not retrieved. 
We estimate the area under the curve using rectangles and not trapezoids.

\subsection{Statistics}
\label{sec:statistics}

\begin{figure}[t]
    \definecolor{appleblue}{RGB}{0,122,255}
\definecolor{calendarRed}{rgb}{1.000, 0.420, 0.420}

\vfill
\centering
\scalebox{0.97}{
\begin{tabular}{ccc}
    \hspace{-25pt}
    \begin{subfigure}{0.22\textwidth}
        \centering
        \begin{tikzpicture}
\small
    \begin{axis}[
        scaled x ticks=true,
        ylabel={\# objects},
        xlabel={\# positives},
        ymin=0, ymax=220,
        xmin=0, xmax=35,
        grid=both,
        grid style={color=lightgray!60, dash pattern=on 2pt off 2pt},
        width=\linewidth,
        height=\linewidth,
        xlabel style={font=\scriptsize, yshift=3pt},
        ylabel style={font=\scriptsize, yshift=-5pt},
        xticklabel style=
        {/pgf/number format/1000 sep=,anchor=north,font=\tiny},
        yticklabel style={/pgf/number format/fixed, font=\tiny},
        ytick={0, 100, 200},
        yticklabels={0, 1, 2},
        xtick={0, 10, 20, 30},
        area style,
        axis on top=false,
        legend pos=north east,
        legend style={cells={anchor=west}, font=\tiny, fill opacity=0.8, row sep=-1pt},
        name=dist
    ]
        \pgfplotstableread[col sep=comma]
        {data/distribution_positives_vs_queries.csv}\loadeddata
        \addplot+[ybar interval, mark=no, fill=appleblue, opacity=0.5] 
            table [x=pos_bin_edge, y=pos_probability]  {\loadeddata};
        \end{axis}
        \path (dist.north west) ++(15pt, -2pt) node[above left, font=\tiny]{${\cdot 10^2}$};
\end{tikzpicture}
        \vspace{-5pt}
        \caption{\label{fig:dist_pos}}
    \end{subfigure}
    &
    \hspace{-40pt}
    \begin{subfigure}{0.22\textwidth}
        \centering
        \begin{tikzpicture}
\small
    \begin{axis}[
        xlabel={\# segments},
        ylabel={\# positives},
        ymin=0, ymax=750,
        xmin=0, xmax=200,
        grid=both,
        grid style={color=lightgray!60, dash pattern=on 2pt off 2pt},
        width=\linewidth,
        height=\linewidth,
        xlabel style={font=\scriptsize, yshift=3pt},
        xticklabel style=
        {/pgf/number format/1000 sep=,anchor=north,font=\tiny},
        ylabel style={font=\scriptsize, yshift=-5pt},
        yticklabel style={/pgf/number format/fixed, font=\tiny},
        ytick={0, 200, 400, 600},
        yticklabels={0, 2, 4, 6},
        xtick={0, 100, 200},
        xticklabels={0, 100, 200+},
        area style,
        axis on top=false,
        legend pos=north east,
        legend style={cells={anchor=west}, font=\tiny, fill opacity=0.8, row sep=-1pt},
        name=dist
    ]
        \pgfplotstableread[col sep=comma]
            {data/distribution_positives_vs_clutter.csv}\loadeddata
        \addplot+[ybar interval, mark=no, fill=appleblue, opacity=0.5] 
            table [x=segment_number_bin_edge, y=segment_number_probability]  {\loadeddata};
    \end{axis}
    \path (dist.north west) ++(15pt, -2pt) node[above left, font=\tiny]{${\cdot 10^2}$};
\end{tikzpicture}
        \vspace{-5pt}
        \caption{\label{fig:dist_seg}}
    \end{subfigure}
    &
    \hspace{-45pt}
    \begin{subfigure}{0.22\textwidth}
        \centering
        \begin{tikzpicture}
    \begin{axis}[
        ymin=0, ymax=600,
        xmin=0, xmax=1,
        xlabel={relative{\color{white} j}bbox{\color{white} j}area},
        ylabel={\# images},
        grid=both,
        grid style={color=lightgray!60, dash pattern=on 2pt off 2pt},
        width=\linewidth,
        height=\linewidth,
        xlabel style={font=\scriptsize, yshift=3pt},
        xticklabel style=
        {/pgf/number format/1000 sep=,anchor=north,font=\tiny},
        ylabel style={font=\scriptsize},
        ylabel style={font=\scriptsize, yshift=-5pt},
        yticklabel style={/pgf/number format/fixed, font=\tiny},
        ytick={0, 200, 400, 600},
        yticklabels={0, 2, 4, 6},
        xtick={0, 0.5, 1.},
        area style,
        axis on top=false,
        legend pos=north east,
        legend style={at={(0.95,0.95)}, cells={anchor=west}, font=\tiny, fill opacity=1., row sep=-1pt},
        name=dist
    ]
        \pgfplotstableread[col sep=comma]
        {data/distribution_bbox_vs_image_type.csv}\loadeddata
        
        \addplot+[ybar interval, mark=no, fill=appleblue, opacity=0.5] 
            table [x=pos_bin_edge, y=pos_probability]  {\loadeddata};
        \addlegendentry{positive}
        \addplot+[ybar interval, mark=no, fill=calendarRed, opacity=0.5] 
            table [x=query_bin_edge, y=query_probability]  {\loadeddata};
        \addlegendentry{query}
        \end{axis}
        \path (dist.north west) ++(15pt, -2pt) node[above left, font=\tiny]{${\cdot 10^2}$};    
\end{tikzpicture}
        \vspace{-5pt}
        \caption{\label{fig:dist_scale}}
    \end{subfigure}
\end{tabular}
}
    \vspace{-15pt}
    \caption{\textbf{\ours statistics}. (a) number of positives per object, (b) positive distribution by the SAM segments outside the bounding box, (c) image distribution by the relative bounding box area.
    \label{fig:statistics}
    \vspace{-15pt}
    }
\end{figure}
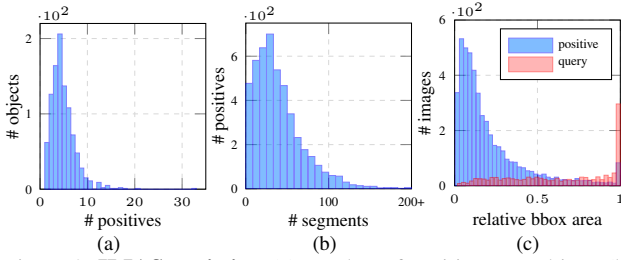

\noindent\textbf{Dataset size.} The final \ours dataset includes 1,000 object instances captured in 5,947 images, of which 1,232 are queries and 4,715 are positives.
Fig.~\ref{fig:dist_pos} shows the distribution of positives per object. Also, 99,144,315 images from YFCC100M are downloaded. All images (queries, positives, distractors) are transferred through Flickr to ensure the same pre-processing.

\noindent\textbf{Taxonomy.} A hierarchical 3-level taxonomy is composed for \ours. All instances are assigned across one to three categories of different granularity levels. The taxonomy consists of 8 categories on the coarser level, 42 on the mid level, and 38 on the finer level. The categories are derived through manual labeling of the objects based on their semantic content. To form the coarser-level categories, we use domain definitions borrowed from prior work~\cite{wac+20,sxj+15,liu2016deepfashion} to align with the literature, \ie art, landmarks, products, fashion. We also define novel categories based on the objects that do not fit into any existing domain. The distribution of objects across categories is uneven, \eg ranging from 168 and 162 for art and landmarks to 83 for products. Each mid- and finer-level category contains at least 4 instances. Note that taxonomy is given to provide statistics about the domains of objects and assist our analysis instead of being leveraged as ground truth. The distribution of taxonomy categories can be inferred by Fig.~\ref{fig:subdomains}, and a detailed figure is provided in the supplementary material.

\noindent\textbf{Bounding box analysis.} A total number of 6,117 bounding boxes are annotated for both queries and positives. Note that positives may display multiple objects of near identical appearance to the query; in such cases, bounding boxes are drawn on all indistinguishable objects. There are 235 images with more than one bounding box.
Based on the annotated bounding boxes, we compute the area covered in the image by the object instances to derive its relative scale. Fig.~\ref{fig:dist_scale} shows the distribution of the scale ratio for queries and positives. Most objects in queries cover the largest area of the images; while in the vast majority of positives, the object covers a small area of less than half the image. It is a result of the severe scale changes and partial views. \looseness=-1
Moreover, we use the Segment Anything Model (SAM)~\cite{kmr+23,rgh+24} to extract object segments from positives. The number of detected segments outside the query object's bounding boxes is computed. This indicates clutter from other items in the positives. Fig.~\ref{fig:dist_seg} shows the segment number distribution, with most images containing multiple segments due to clutter.

\subsection{\miniours}
\label{sec:mini}
We provide a small version of \ours, called \miniours, to facilitate quick experimentations. 
It consists of the query and positive images collected for \ours, and a subset of the YFCC100M distractors.
Instead of randomly subsampling YFCC100M, we construct a challenging subset with the help of VLMs.
We aim at selecting distractors displaying objects of similar categories as the query objects.
We use the text category labels of the taxonomy as text queries. We also extend them with standard templates used for zero-shot recognition~\cite{clip}, which resulted in several thousands. T2i similarity between each text query and each distractor image is estimated. A similarity score for each distractor is derived based on its maximum similarity over the text queries. We ensemble the scores of 3 VLMs to rank images. The top-5M ranked distractors compose the final \miniours. 
Our experiments indicate that this subset is significantly more challenging than a random subset of the same size.

\section{Benchmark methods}
We describe the methods and foundation models we evaluate on \ours, which are grouped according to their type of representations used for retrieval in global (i2i) representations, re-ranking with global (i2i) representations, re-ranking with local (i2i) representations, and text-to-image. A detailed list of models, their performance, and implementation details are in the supplementary materials.

\begin{figure*}[t]
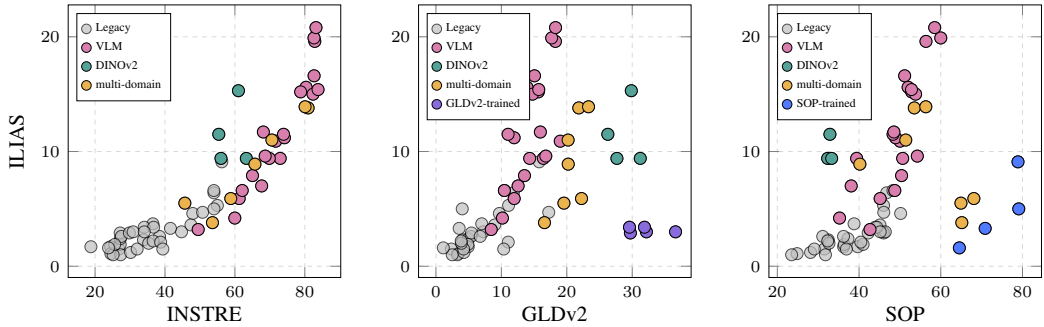

  \vspace{-10pt}
  \centering
  \scalebox{0.9}{
    \hspace{10pt}
    \begin{tikzpicture}
\begin{axis}[
  width=.32\linewidth,
  height=.32\linewidth,
  xlabel={\small INSTRE},
  ylabel={\small ILIAS},
  tick label style={font=\scriptsize},
  ylabel near ticks, xlabel near ticks, 
  xlabel style={yshift=3pt},
  grid=both,
  grid style={color=lightgray!60, dash pattern=on 2pt off 2pt},  
  legend entries={Legacy, VLM, DINOv2, multi-domain},
  legend pos=north west,
  legend style={cells={anchor=west}, font=\tiny, fill opacity=1.0, row sep=-1pt, inner sep=2pt},
  legend image post style={scale=0.8},
  ]

  \input{data/instre_gld_sop}

   \addplot[only marks, mark options={draw=black}, color=lightgray, solid, mark=*, opacity=0.7, mark size=2.5] table[y=ILIAS, x=INSTRE] \instregldsop;

  \addplot[only marks, mark options={draw=black}, color=applepink, solid, mark=*, mark size=2.5] table[y=ILIAS, x expr={\thisrow{INSTRE} / (\thisrow{show} == 1)}] \instregldsop;

  \addplot[only marks, mark options={draw=black}, color=appleteal, solid, mark=*, mark size=2.5] table[y=ILIAS, x expr={\thisrow{INSTRE} / (\thisrow{show} == 2)}] \instregldsop;

  \addplot[only marks, mark options={draw=black}, color=applemustard, solid, mark=*, mark size=2.5] table[y=ILIAS, x expr={\thisrow{INSTRE} / (\thisrow{show} == 5)}] \instregldsop;

\end{axis}
\end{tikzpicture} 
    \hspace{10pt}
    \begin{tikzpicture}
\begin{axis}[
  width=.32\linewidth,
  height=.32\linewidth,
  xlabel={\small GLDv2},
  xmax=39,
  tick label style={font=\scriptsize},
  ylabel near ticks, xlabel near ticks, 
  xlabel style={yshift=3pt},
  grid=both,
  grid style={color=lightgray!60, dash pattern=on 2pt off 2pt},
  legend entries={Legacy, VLM, DINOv2, multi-domain, GLDv2-trained},
  legend pos=north west,
  legend style={cells={anchor=west}, font=\tiny, fill opacity=1.0, row sep=-1pt, inner sep=2pt},
  legend image post style={scale=0.8},
  ]

  \input{data/instre_gld_sop}

  \addplot[only marks, mark options={draw=black}, color=lightgray, solid, mark=*, opacity=0.7, mark size=2.5] table[y=ILIAS, x=GLDv2] \instregldsop;

  \addplot[only marks, mark options={draw=black}, color=applepink, solid, mark=*, mark size=2.5] table[y=ILIAS, x expr={\thisrow{GLDv2} / (\thisrow{show} == 1)}] \instregldsop;

  \addplot[only marks, mark options={draw=black}, color=appleteal, solid, mark=*, mark size=2.5] table[y=ILIAS, x expr={\thisrow{GLDv2} / (\thisrow{show} == 2)}] \instregldsop;   

  \addplot[only marks, mark options={draw=black}, color=applemustard, solid, mark=*, mark size=2.5] table[y=ILIAS, x expr={\thisrow{GLDv2} / (\thisrow{show} == 5)}] \instregldsop;

  \addplot[only marks, mark options={draw=black}, color=applepurple, solid, mark=*, mark size=2.5] table[y=ILIAS, x expr={\thisrow{GLDv2} / (\thisrow{show} == 3)}] \instregldsop;
  
  \end{axis}
\end{tikzpicture}
    \hspace{10pt}
    \begin{tikzpicture}
\begin{axis}[
  width=.32\linewidth,
  height=.32\linewidth,
  xlabel={\small SOP},
  tick label style={font=\scriptsize},
  ylabel near ticks, xlabel near ticks, 
  xlabel style={yshift=3pt},
  grid=both,
  grid style={color=lightgray!60, dash pattern=on 2pt off 2pt},
  legend entries={Legacy, VLM, DINOv2, multi-domain, SOP-trained},
  legend pos=north west,
  legend style={cells={anchor=west}, font=\tiny, fill opacity=1.0, row sep=-1pt, inner sep=2pt},
  legend image post style={scale=0.8},
  ]

  \input{data/instre_gld_sop}

  \addplot[only marks, mark options={draw=black}, color=lightgray, solid, mark=*, opacity=0.7, mark size=2.5] table[y=ILIAS, x=SOP] \instregldsop;

  \addplot[only marks, mark options={draw=black}, color=applepink, solid, mark=*, mark size=2.5] table[y=ILIAS, x expr={\thisrow{SOP} / (\thisrow{show} == 1)}]
  \instregldsop;

  \addplot[only marks, mark options={draw=black}, color=appleteal, solid, mark=*, mark size=2.5] table[y=ILIAS, x expr={\thisrow{SOP} / (\thisrow{show} == 2)}]
  \instregldsop;

  \addplot[only marks, mark options={draw=black}, color=applemustard, solid, mark=*, mark size=2.5] table[y=ILIAS, x expr={\thisrow{SOP} / (\thisrow{show} == 5)}]
  \instregldsop;

  \addplot[only marks, mark options={draw=black}, color=appleblue, solid, mark=*, mark size=2.5] table[y=ILIAS, x expr={\thisrow{SOP} / (\thisrow{show} == 4)}]
  \instregldsop;
  
\end{axis}
\end{tikzpicture} 
  }
  \vspace{-7pt}
  \caption{\textbf{Comparison with other instance-level retrieval datasets} via reporting mAP@1k. INSTRE: 27.3K db size, multi-domain. GLDv2: 762K db size, single-domain. SOP:  60.5K db size, single-domain. Different network types are color-coded. For GLDv2 and SOP, models fine-tuned on these domains with the corresponding training sets are highlighted. No linear adaptation is used.
  \label{fig:instre_gld_sop}
  \vspace{-12pt}
  }
\end{figure*}

\noindent\textbf{Image-to-image retrieval with global representations.} Global representation methods use image encoders to map images to global descriptors and rank db images based on cosine similarity. 
We evaluate legacy and recent foundation models, varying in architecture, descriptor dimensionality, training scheme, training data, and input resolution. Foundation models~\cite{bha+} are the models trained with a training set on the scale of a hundred million. 
Particularly, 23 CNN~\cite{ksh12,sz14,slj+15,hzr+16,zvs+18,tl19,convnext} and 45 ViT~\cite{dbk+21,fwx+23} models, trained with supervision~\cite{wtj+21,vit-augreg,tcd+21,kjk23}, self-supervision~\cite{odm+24,cmm+20,hfw+20,ctm+21}, distillation~\cite{ady+23,tcd+21,swl+24}, or visual-language alignment~\cite{clip,metaclip,siglip,cbw+23,evaclip,siglip2} are benchmarked. 
Most of the non-foundation models are trained on ImageNet~\cite{dds+09}. There are models trained on single specific domains~\cite{sck+23,lsl+22,ptm+22}, \ie landmarks or products on GLDv2 or SOP.
Universal models~\cite{ycc+23,yca+24,ady+23,swl+24} trained on multi-domains or multi-task schemes are included.
The full list of models and results is provided in supplementary materials.

To mitigate the differences in training resolution, we use three widely-used resolutions, \ie 384, 512, and 724 and resize images so that their larger dimension matches one of the three.
The test resolution is defined to be one resolution above the one used for training, \eg a network trained with 224 or 384 is tested with 384 or 512, respectively.
The vast majority of models achieve best performance under this rule. Similar behavior is observed in the literature~\cite{tvd+19,st23}.

\noindent\textbf{Linear adaptation for i2i retrieval.} Pre-trained foundation models, as well as legacy models, are trained to extract representations that are applicable to various tasks; not all encoded features are directly relevant to instance-level retrieval.
To adapt the representation to the task at hand, we propose to train a single linear layer (projection) on top of frozen backbones.
The recently introduced Universal Embeddings (UnED) dataset~\cite{ycc+23} is used for learning the linear adaptation. 
UnED contains images from 8 different domains with fine-grained and/or instance-level class annotation.
In our experiments, the linear layer that converts the backbone output to a 512D descriptor is trained on a uniformly sampled subset of 1M images from UnED.
The linear adaptation layer is trained with the UJCDS~\cite{ycc+23} method. 

\noindent\textbf{Text-to-image retrieval.}
Text-to-image retrieval is performed using Vision-Language Models (VLMs) trained to align the two modalities. 
Retrieval is performed based on cosine similarity between the text query and db image descriptors that are extracted using the textual and visual encoder, respectively.
We evaluate 17 VLM models.

\noindent\textbf{Re-ranking with global representations.} Such methods rely on global descriptors for exhaustive search during the initial ranking, but also for a second refinement stage that issues a new query.
We experiment with $\alpha$QE~\cite{rtc19}, the adaptive variant of average Query Expansion~\cite{cps+07}. After the initial ranking, the descriptors of the top-ranked images are aggregated with the query via weighted average pooling.
The weights are derived from the similarity to the query in the power of $\alpha$. We don't have a validation set; hence, we use a fixed value $\alpha=1$.

\noindent\textbf{Re-ranking with local representations.}
These re-ranking methods rely on global descriptors for exhaustive search during the initial ranking but estimate query-to-db image similarity based on local descriptors for a second refinement stage of the ranked list of images.
We experiment with three methods: (i) Chamfer Similarity (CS)~\cite{rsa+14, btb+77} on the similarity matrix between local descriptors across the image pair. We use the asymmetric variant of CS with max over db descriptors and sum over query descriptors. (ii) Spatial Verification (SP)~\cite{fib+81, pci+07,cas20}, a common re-ranking method where point correspondences are processed with a RANSAC-like process and the number of inliers is used for re-ranking. (iii) AMES~\cite{ski+24}, a recent transformer-based network to estimate the similarity between sets of local descriptors. 
Due to the scale of \ours database, we use only 100 binary local descriptors for each database image and 600 for the query image.
Local descriptors are extracted using the base variant of DINOv2 with registers~\cite{odm+24,doj+23} and selected based the local descriptor detector used in AMES~\cite{ski+24}. Top-1k retrieved images are re-ranked.
\section{Experiments}
\label{sec:experiments}

\begin{table*}[t]
  \vspace{-10pt}
  \centering
  \scalebox{0.83}{
    \newcolumntype{C}{>{\raggedleft\arraybackslash}p{3em}}
\renewcommand{\arraystretch}{.9}
\setlength\tabcolsep{3.5pt}
\small
\begin{tabular}{l@{\lsp}rrrrrr@{\lsp}CCCCC}
\toprule
& & & & & & & \multicolumn{3}{c}{\textbf{image-to-image}} & \multicolumn{2}{c}{\textbf{text-to-image}} \\ \cmidrule(lr){8-10} \cmidrule(lr){11-12}
\multicolumn{1}{c}{\textbf{model}} & \multicolumn{1}{c}{\textbf{arch}} & \multicolumn{1}{c}{\textbf{train}} & \multicolumn{1}{c}{\textbf{dataset}} & \multicolumn{1}{c}{\textbf{data size}} & \multicolumn{1}{c}{\textbf{train res}} & \textbf{test res} & \textbf{5M$^\dagger$} & \textbf{100M$^\dagger$} & \textbf{100M} & \textbf{100M} & \textbf{5M} \\ 
\midrule
ResNet50~\cite{hzr+16}           & R50   & sup & \texttt{in1k}      & 1M   & 224  & 384 & \grca{2.5}  & \grcb{1.8}  & \grcc{1.7}  & -           & -           \\
DINO~\cite{ctm+21}               & R50   & ssl & \texttt{in1k}      & 1M   & 224  & 384 & \grca{4.1}  & \grcb{2.9}  & \grcc{2.9}  & -           & -           \\
ConvNext~\cite{convnext}         & CN-L  & sup & \texttt{in1k}      & 1M   & 288  & 384 & \grca{4.2}  & \grcb{2.9}  & \grcc{2.2}  & -           & -           \\
OAI-CLIP~\cite{hzr+16}           & R50   & vla & \texttt{openai}    & 400M & 224  & 384 & \grca{8.5}  & \grcb{6.0}  & \grcc{3.2}  & \grcd{1.5}  & \grce{2.3}  \\
OpenCLIP~\cite{convnext,iww+21}  & CN-B  & vla & \texttt{laion2b}   & 2B   & 256  & 384 & \grca{18.1} & \grcb{14.0} & \grcc{7.9}  & \grcd{4.6}  & \grce{7.0}  \\
OpenCLIP~\cite{convnext,iww+21}  & CN-L  & vla & \texttt{laion2b}   & 2B   & 320  & 512 & \grca{22.9} & \grcb{18.3} & \grcc{9.6}  & \grcd{8.1}  & \grce{11.5} \\
\midrule
ViT~\cite{dbk+21,vit-augreg}     & ViT-B & sup & \texttt{in1k}      & 1M   & 224  & 384 & \grca{1.9}  & \grcb{1.3}  & \grcc{1.0}  & -           & -           \\
EVA-MIM~\cite{fwx+23}            & ViT-B & ssl & \texttt{in22k}     & 142M & 224  & 384 & \grca{4.7}  & \grcb{3.2}  & \grcc{2.1}  & -           & -           \\
ViT~\cite{dbk+21,vit-augreg}     & ViT-B & sup & \texttt{in21k}     & 14M  & 224  & 384 & \grca{6.2}  & \grcb{4.4}  & \grcc{3.0}  & -           & -           \\
DINO~\cite{ctm+21}               & ViT-B & ssl & \texttt{in1k}      & 1M   & 224  & 384 & \grca{6.6}  & \grcb{4.8}  & \grcc{3.7}  & -           & -           \\
UDON-CLIP~\cite{yca+24}          & ViT-B & sup & \texttt{uned}      & 2.8M & 224  & 384 & \grca{9.2}  & \grcb{6.7}  & \grcc{5.9}  & -           & -           \\
OAI-CLIP~\cite{clip}             & ViT-B & vla & \texttt{openai}    & 400M & 224  & 384 & \grca{10.7} & \grcb{7.9}  & \grcc{4.2}  & \grcd{1.6}  & \grce{2.7}  \\
EVA-CLIP~\cite{evaclip}          & ViT-B & vla & \texttt{merged2b}  & 2B   & 224  & 384 & \grca{11.7} & \grcb{8.7}  & \grcc{5.9}  & \grcd{2.5}  & \grce{4.4}  \\
MetaCLIP~\cite{metaclip}         & ViT-B & vla & \texttt{2pt5b}     & 2.5B & 224  & 384 & \grca{12.7} & \grcb{9.4}  & \grcc{6.6}  & \grcd{4.9}  & \grce{7.6}  \\
DINOv2~\cite{odm+24}             & ViT-B & ssl & \texttt{lvd142m}   & 142M & 518  & 724 & \grca{15.0} & \grcb{12.1} & \grcc{11.5} & -           & -           \\
SigLIP~\cite{siglip}             & ViT-B & vla & \texttt{webli}     & 10B  & 256  & 384 & \grca{20.6} & \grcb{16.7} & \grcc{11.5} & \grcd{7.5}  & \grce{10.3} \\
SigLIP~\cite{siglip}             & ViT-B & vla & \texttt{webli}     & 10B  & 384  & 512 & \grca{26.2} & \grcb{21.5} & \grcc{15.6} & \grcd{11.0} & \grce{14.4} \\
SigLIP~\cite{siglip}             & ViT-B & vla & \texttt{webli}     & 10B  & 512  & 724 & \grca{27.5} & \grcb{23.0} & \grcc{16.6} & \grcd{11.1} & \grce{14.6} \\
SigLIP2~\cite{siglip2}           & ViT-B & vla & \texttt{webli}     & 10B  & 512  & 724 & \grca{28.6} & \grcb{23.5} & \grcc{15.4} & \grcd{10.4} & \grce{14.6} \\
\midrule
EVA-MIM~\cite{fwx+23}            & ViT-L & ssl & \texttt{in22k}     & 14M  & 224  & 384 & \grca{3.9}  & \grcb{2.7}  & \grcc{1.5}  & -           & -           \\
ViT~\cite{dbk+21,vit-augreg}     & ViT-L & sup & \texttt{in21k}     & 14M  & 224  & 384 & \grca{7.3}  & \grcb{5.3}  & \grcc{4.6}  & -           & -           \\
EVA-MIM~\cite{fwx+23}            & ViT-L & ssl & \texttt{merged38m} & 38B  & 224  & 384 & \grca{8.8}  & \grcb{6.1}  & \grcc{4.7}  & -           & -           \\
OAI-CLIP~\cite{clip}             & ViT-L & vla & \texttt{openai}    & 400M & 224  & 384 & \grca{15.8} & \grcb{11.9} & \grcc{7.0}  & \grcd{4.6}  & \grce{6.7}  \\
OpenCLIP~\cite{iww+21,cbw+23}    & ViT-L & vla & \texttt{laion2b}   & 2B   & 224  & 384 & \grca{17.5} & \grcb{13.7} & \grcc{9.4}  & \grcd{7.0}  & \grce{9.4}  \\
Unicom~\cite{ady+23}             & ViT-L & dist& \texttt{laion400m} & 400M & 336  & 512 & \grca{18.6} & \grcb{14.6} & \grcc{13.9} & -           & -           \\
OAI-CLIP~\cite{clip}             & ViT-L & vla & \texttt{openai}    & 400M & 336  & 512 & \grca{19.9} & \grcb{15.2} & \grcc{9.4}  & \grcd{5.8}  & \grce{8.4}  \\
DINOv2~\cite{odm+24}             & ViT-L & ssl & \texttt{lvd142m}   & 142M & 518  & 724 & \grca{18.8} & \grcb{15.3} & \grcc{15.3} & -           & -           \\
EVA-CLIP~\cite{evaclip}          & ViT-L & vla & \texttt{merged2b}  & 2B   & 336  & 512 & \grca{20.9} & \grcb{16.0} & \grcc{10.9} & \grcd{7.2}  & \grce{10.6} \\
MetaCLIP~\cite{metaclip}         & ViT-L & vla & \texttt{2pt5b}     & 2.5B & 224  & 384 & \grca{21.7} & \grcb{16.9} & \grcc{11.7} & \grcd{9.2}  & \grce{13.1} \\
SigLIP~\cite{siglip}             & ViT-L & vla & \texttt{webli}     & 10B  & 256  & 384 & \grca{26.3} & \grcb{21.8} & \grcc{15.2} & \grcd{12.8} & \grce{16.4} \\
SigLIP~\cite{siglip}             & ViT-L & vla & \texttt{webli}     & 10B  & 384  & 512 & \grca{34.3} & \grcb{28.9} & \grcc{19.6} & \grcd{18.1} & \grce{22.2} \\
SigLIP2~\cite{siglip2}           & ViT-L & vla & \texttt{webli}     & 10B  & 512  & 724 & \grca{37.3} & \grcb{31.3} & \grcc{20.8} & \grcd{19.8} & \grce{24.7} \\
\bottomrule
\end{tabular}
  }
  \vspace{-7pt}
  \caption{\textbf{Performance comparison using mAP@1k on \ours and \miniours for global representation models for i2i and t2i}. Comparison of model architecture (arch), training scheme (train), training data, and train/test resolution. $\dagger$ indicates results with the linear adaptation. 5M and 100M correspond to the mini and full versions of the dataset, respectively. sup, ssl, dist, vla: supervised learning, self-supervised learning, distillation and vision-language alignment. R50, CN: ResNet50 and ConvNext.
  \vspace{-15pt}
  \label{tab:foundational}
  }
\end{table*}

We evaluate all the above models and methods, extracting useful insights regarding the factors that boost retrieval performance. \ours is compared with other existing datasets for instance-level image retrieval. We analyze the performance of selected models\footnote{Although the very recent SigLIP2 is the best-performing model, we conduct most experiments with SigLIP.} to break down the impact of different \ours attributes, such as domains, clutter, and scale. Unless stated otherwise, we use the large model variants with the largest resolution available, \eg in our analysis we use SigLIP ViT-L trained with 384 resolution.

\subsection{Comparison with other instance-level datasets}
\label{sec:comparison_datasets}

In Fig.~\ref{fig:instre_gld_sop}, \ours is compared with other instance-level retrieval datasets based on evaluation of the same models. Linear adaptation is not applied as parts of GLDv2 and SOP are included in the UnED dataset. 
Only for the sake of this comparison, and for no other experiment in this work, we use models fine-tuned on specific domains (in-domain models), \ie on the training sets of SOP and GLDv2.
INSTRE, which is also multi-domain, shows a correlation to \ours, but its performance is saturated due to its small size.
For single-domain datasets, in-domain models outperform others by a large margin, with few exceptions, \ie DINOv2, which includes the trainset of GLDv2 in its training data.
Several multi-domain models perform well on SOP since their training set is usually included in the training data. 
However, in-domain and multi-domain models face challenges on \ours, highlighting the diversity of our dataset.

\subsection{Method comparison}
\label{sec:methods_comparison}

\noindent\textbf{Image-to-image retrieval with global representations.}
Tab.~\ref{tab:foundational} presents the performance of global descriptor models on \ours. Selected models are presented to highlight useful comparisons, while many other models are included in the supplementary material. The main factors that improve performance are the size of the training set, training resolution, and model architecture, which aligns with the literature. The impact of dataset size is apparent in various model combinations, \eg CLIP with \texttt{openai} and \texttt{laion2b}. This is also pronounced by the dominance of foundation models. Training with large resolution brings significant gains and consistently improves mAP@1k. In some cases of SigLIP, smaller models trained with large resolutions outperform larger ones trained with small resolutions. For the models of the same resolution, it is a common trend for larger model variants to bring corresponding performance gains. 
In general, VLMs perform the best. From non-VLMs, only DINOv2 and Unicom achieve competitive performance. Masked Image Modeling (MIM) and supervised models are not performing well.
Our linear adaptation scheme is very effective, improving most models. The boost is more pronounced in the case of VLMs. A possible explanation for such improvements is that image-to-image relations are not optimized during the training of VLMs.

\noindent\textbf{Text-to-image retrieval.}
Following results in Tab.~\ref{tab:foundational}, similar conclusions are derived for the t2i case. Retrieval performance improves with the scaling of the training data. The larger model achieves significantly better results, \ie compare the base with large variants. Finally, it is noteworthy that the best performance achieved by SigLIP2 is very close to the i2i performance when no adaptation is used. Note that t2i includes 1k text queries in total, with one query per object, while i2i 1,232 image queries.

\begin{table}[t]
  \centering
  \scalebox{0.85}{
    \small
\begin{tabular}{lYYY}
\toprule
\textbf{model} & \textbf{100M} & \textbf{5M-mini} & \textbf{5M-rand} \\
\midrule
DINOv2$^\dagger$~\cite{odm+24}              & 15.3 & 18.8 & 22.7\scriptsize{$\pm0.2$} \\
EVA-CLIP$^\dagger$~\cite{evaclip}           & 16.0 & 20.9 & 28.8\scriptsize{$\pm0.2$} \\
MetaCLIP$^\dagger$~\cite{metaclip}          & 16.9 & 21.7 & 29.2\scriptsize{$\pm0.1$} \\
OpenCLIP$^\dagger$~\cite{iww+21,convnext}   & 18.3 & 22.9 & 30.9\scriptsize{$\pm0.2$} \\
SigLIP$^\dagger$~\cite{siglip}              & 28.9 & 34.3 & 41.8\scriptsize{$\pm0.1$} \\ 
\bottomrule
\end{tabular}
  }
  \vspace{-7pt}
  \caption{\textbf{A challenging distractor subset for \miniours.} mAP@1k evaluated for different distractor sets, 100M: the full dataset, 5M-mini: \miniours subset, 5M-rand: random subset. We report the mean and std of 3 randomly sampled subsets. $\dagger$ indicates results with the linear adaptation.
  \label{tab:mini}}
  \vspace{-10pt}
\end{table}
\begin{table}[h!]
  \centering
  \scalebox{0.85}{
    \small
\begin{tabular}{lcccc}
\toprule
\multirow{2}{*}{\textbf{reranking}}  & \multicolumn{2}{c}{\textbf{SigLIP}~\cite{siglip}} & \multicolumn{2}{c}{\textbf{SigLIP$^\dagger$}~\cite{siglip}} \\ \cmidrule(lr){2-3} \cmidrule(lr){4-5}
 & \textbf{mAP@1k} & \textbf{oracle}  & \textbf{mAP@1k} & \textbf{oracle} \\ 
\midrule
global                           & 19.6 & 48.7 & 28.9 & 56.0 \\ 
\midrule
$\alpha$QE1~\cite{cps+07,rtc19}  & 22.1 & 44.7 & 33.7 & 56.9 \\
$\alpha$QE2~\cite{cps+07,rtc19}  & 20.4 & 40.8 & 31.5 & 54.4 \\
$\alpha$QE5~\cite{cps+07,rtc19}  & 14.3 & 34.9 & 23.5 & 49.3 \\ 
\midrule
CS~\cite{rsa+14}                 & 22.9 & 48.7 & 32.5 & 56.0 \\
SP~\cite{pci+07}                 & 21.8 & 48.7 & 30.5 & 56.0 \\
AMES~\cite{ski+24}               & 26.4 & 48.7 & 35.6 & 56.0 \\
\bottomrule
\end{tabular}
  }
  \vspace{-7pt}
  \caption{\textbf{Performance comparison for re-ranking methods.} Oracle represents the performance of perfect re-ranking at the top-1k images. Top: query expansion with global descriptors. Bottom: re-ranking with local descriptors. $\dagger$: results with linear adaptation.
  \label{tab:reranking}}
  \vspace{-20pt}
\end{table}

\noindent\textbf{Evaluation of \miniours selection.} Tab.~\ref{tab:mini} shows performance on \miniours for five models with linear adaptation. 
The selected subset is significantly more challenging than a random selection of 5M images. 
More precisely, a set of approximately $\sim$26M random images matches the performance of \miniours. 

\noindent\textbf{Retrieval with re-ranking.} Tab.~\ref{tab:reranking} shows the performance of re-ranking methods applied on top of SigLIP with and without linear adaptation on \ours. Complementary to mAP@1k, an oracle-based top-1k re-ranking metric is reported as the upper bound of a re-ranking method that processes the top 1k images.
Local similarity estimated by a learned model proves to be very effective for re-ranking. Nevertheless, the oracle re-ranking performance indicates that there is a lot more space for improvements. Re-ranking with QE is useful when the number of aggregated neighbors is low and drops below the baseline when the number of neighbors is increased. 
Notably, global re-ranking affects and, interestingly, decreases oracle performance since the whole db is re-ranked; while local re-ranking does not affect it since it is performed only on a shortlist of images.

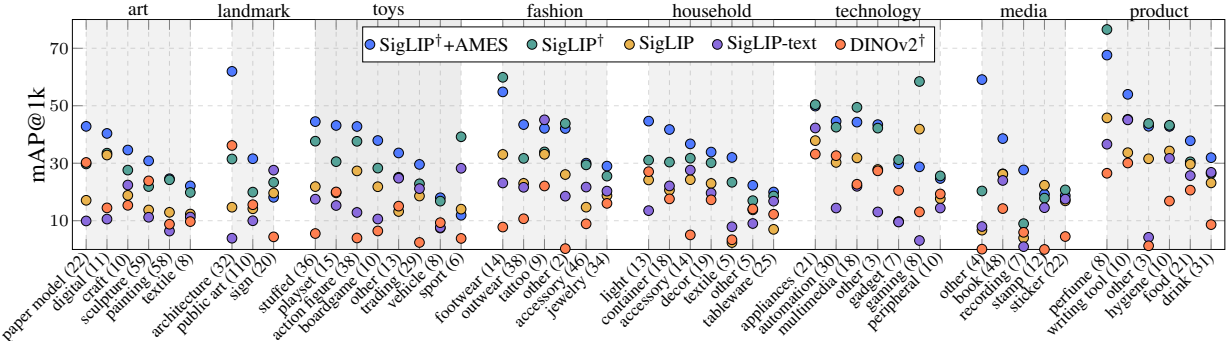
\begin{figure*}[t]
\vspace{-10pt}
\hspace{-15pt}
  \centering
  \scalebox{0.91}{
    \pgfplotsset{every tick label/.append style={font=\scriptsize}}

\newcommand{\addFilling}[4]{
    \addplot [
        fill=gray, fill opacity=0.1, draw=none,
    ] coordinates {
        ({#1}, 100)
        ({#2}, 100)
        ({#2}, -1)
        ({#1}, -1)
    } -- cycle;
}

\begin{tikzpicture}
\begin{axis}[
    width=1.05\linewidth,
    height=0.282\linewidth,
    ymin=0, ymax=80,
    ytick={10,30,50,70,90}, 
    ylabel={mAP@1k},
    ylabel style={yshift=-5pt},
    yticklabel style={font=\scriptsize,xshift=2pt},
    xtick=data,
    xticklabel style={rotate=45, anchor=east, font=\scriptsize, yshift=-2pt, xshift=2pt, name=xlabel\ticknum},
    symbolic x coords=\xcoords,
    symbolic x coords={
        art       - paper model   (22),
        art       - digital       (11),
        art       - craft         (10),
        art       - sculpture     (59),
        art       - painting      (58),
        art       - textile       (8),
        test,
        landmark  - architecture  (32),
        landmark  - public art    (110),
        landmark  - sign          (20),
        test,
        toy       - stuffed       (36),
        toy       - playset       (15),
        toy       - action figure (38),
        toy       - boardgame     (10),
        toy       - other         (13),
        toy       - trading       (29),
        toy       - vehicle       (8),
        toy       - sport         (6),
        test,
        fashion   - footwear      (14),
        fashion   - outwear       (38),
        fashion   - tattoo        (9),
        fashion   - other         (2),
        fashion   - accessory     (46),
        fashion   - jewelry       (34),
        test,
        household - light         (13),
        household - container     (18),
        household - accessory     (14),
        household - decor         (19),
        household - textile       (5),
        household - other         (5),
        household - tableware     (25),
        test,
        technology- appliances    (21),
        technology- automation    (30),
        technology- multimedia    (18),
        technology- other         (3),
        technology- gadget        (7),
        technology- gaming        (8),
        technology- peripheral    (10),
        test,
        media     - other         (4),
        media     - book          (48),
        media     - recording     (7),
        media     - stamp         (12),
        media     - sticker       (22),
        test,
        product   - perfume       (8),
        product   - writing tool  (10),
        product   - other         (3),
        product   - hygiene       (10),
        product   - food          (21),
        product   - drink         (31),
    },
    xticklabels={
         paper model   (22),
         digital       (11),
         craft         (10),
         sculpture     (59),
         painting      (58),
         textile       (8),
         architecture  (32),
         public art    (110),
         sign          (20),
         stuffed       (36),
         playset       (15),
         action figure (38),
         boardgame     (10),
         other         (13),
         trading       (29),
         vehicle       (8),
         sport         (6),
         footwear      (14),
         outwear       (38),
         tattoo        (9),
         other         (2),
         accessory     (46),
         jewelry       (34),
         light         (13),
         container     (18),
         accessory     (14),
         decor         (19),
         textile       (5),
         other         (5),
         tableware     (25),
         appliances    (21),
         automation    (30),
         multimedia    (18),
         other         (3),
         gadget        (7),
         gaming        (8),
         peripheral    (10),
         other         (4),
         book          (48),
         recording     (7),
         stamp         (12),
         sticker       (22),
         perfume       (8),
         writing tool  (10),
         other         (3),
         hygiene       (10),
         food          (21),
         drink         (31),
    },
    grid=major,
    grid style={color=lightgray!60, dash pattern=on 2pt off 2pt},
    enlarge x limits={abs=0.2cm},
    legend style={
        at={(0.5,0.805)}, 
        anchor=south, 
        legend columns=-1,  
        /tikz/every even column/.append style={column sep=0.2cm}, 
        font=\footnotesize,
        inner sep=1pt,  
        outer sep=1pt 
    },
    legend cell align={left},
    name=dist,
]

\pgfplotstableread[col sep=comma]{data/performance_per_subdomain.csv}\loadeddata

\addplot+[
    only marks, 
    mark=*, 
    mark options={fill=appleblue, draw=black}, 
] table [
    x=subdomain, 
    y=avg_ap_ames_siglip_384,
    col sep=comma
] {\loadeddata};
\addlegendentry{SigLIP$^\dagger$+AMES}

\addplot+[
    only marks, 
    mark=*, 
    color=appleteal, 
    mark options={fill=appleteal, draw=black}, 
] table [
    x=subdomain, 
    y=avg_ap_siglip_384_proj,
    col sep=comma
] {\loadeddata};
\addlegendentry{SigLIP$^\dagger$}

\addplot+[
    only marks, 
    mark=*, 
    color=applemustard, 
    mark options={fill=applemustard, draw=black}, 
] table [
    x=subdomain, 
    y=avg_ap_siglip_384_nw,
    col sep=comma
] {\loadeddata};
\addlegendentry{SigLIP}

\addplot+[
    only marks, 
    mark=*, 
    color=applepurple, 
    mark options={fill=applepurple, draw=black}, 
] table [
    x=subdomain, 
    y=avg_ap_siglip_384_text,
    col sep=comma
] {\loadeddata};
\addlegendentry{SigLIP-text}

\addplot+[
    only marks, 
    mark=*, 
    color=appleorange,
    mark options={fill=appleorange, draw=black},
] table [
    x=subdomain, 
    y=avg_ap_dinov2_vitl14,
    col sep=comma
] {\loadeddata};
\addlegendentry{DINOv2$^\dagger$}

\addplot [
    fill=gray, fill opacity=0.1,draw=none,
] coordinates {
    ({art       - paper model   (22)}, 100)
    ({art         - textile      (8)}, 100)
    ({art         - textile      (8)}, -1)
    ({art       - paper model   (22)}, -1)
} -- cycle;

\addplot [
    fill=gray, fill opacity=0.1,  draw=none,
] coordinates {
    ({landmark    - architecture (32)}, 100)
    ({landmark    - sign         (20)}, 100)
    ({landmark    - sign         (20)}, -1)
    ({landmark    - architecture (32)}, -1)
} -- cycle;

\addplot [
    fill=gray, fill opacity=0.15,draw=none,
] coordinates {
    ({toy       - stuffed         (36)}, 100)
    ({toy       - sport       (6)}, 100)
    ({toy       - sport       (6)}, -1)
    ({toy       - stuffed         (36)}, -1)
} -- cycle;

\addplot [
    fill=gray, fill opacity=0.1, draw=none,
] coordinates {
    ({fashion     - footwear     (14)}, 100)
    ({fashion     - jewelry      (34)}, 100)
    ({fashion     - jewelry      (34)}, -1)
    ({fashion     - footwear     (14)}, -1)
} -- cycle;

\addplot [
    fill=gray, fill opacity=0.1,draw=none,
] coordinates {
    ({household - light     (13)}, 100)
    ({household - tableware         (25)}, 100)
    ({household - tableware         (25)}, -1)
    ({household - light     (13)}, -1)
} -- cycle;

\addplot [
    fill=gray, fill opacity=0.1,draw=none,
] coordinates {
    ({technology- appliances    (21)}, 100)
    ({technology- peripheral        (10)}, 100)
    ({technology- peripheral        (10)}, -1)
    ({technology- appliances    (21)}, -1)
} -- cycle;

\addplot [
    fill=gray, fill opacity=0.1,draw=none,
] coordinates {
    ({media     - other         (4)}, 100)
    ({media     - sticker         (22)}, 100)
    ({media     - sticker         (22)}, -1)
    ({media     - other         (4)}, -1)
} -- cycle;

\addplot [
    fill=gray, fill opacity=0.1,draw=none,
] coordinates {
    ({product   - perfume       (8)}, 100)
    ({product   - drink         (31)}, 100)
    ({product   - drink         (31)}, -1)
    ({product   - perfume       (8)}, -1)
} -- cycle;

\end{axis}

\foreach \X/\value in {0.96/art, 2.64/landmark, 4.60/toys, 7.02/fashion, 9.30/household, 11.72/technology,  13.83/media,  15.8/product} {
    \node[anchor=north, font=\footnotesize] at (\X,3.70cm) {\scalebox{1.}{\value}};
}
\end{tikzpicture}
  }
  \vspace{-16pt}
  \caption{\textbf{Performance comparison per category.} mAP@1k  averaged over objects in the same mid-level taxonomy category,   organized by their primary-level category size, with sorting within each group by SigLIP$^\dagger$+AMES performance. Comparison between SigLIP with and without adaptation, SigLIP combined with AMES reranking, SigLIP t2i, and DINOv2. $\dagger$ indicates results with the linear adaptation.
  \label{fig:subdomains}
  \vspace{-16pt}
  }
\end{figure*}

\begin{figure}[t]
  \centering
  \vspace{-3pt}
  \scalebox{0.88}{
    \begin{tabular}{cc}

\hspace{-15pt}
\vspace{-3pt}
\begin{tikzpicture}
    \footnotesize
    \begin{axis}[
        width=0.57\linewidth,
        height=0.57\linewidth,
        xmin=-1,
        xmax=101,
        ymin=-1,
        ymax=101,
        title={\footnotesize w/ vs w/o linear adaptation (0.78)},
        grid=both,
        grid style={color=lightgray!60, dash pattern=on 2pt off 2pt},
        xtick={0, 20, 40, 60, 80, 100},
        ytick={0, 20, 40, 60, 80, 100},
        title style = {yshift = -4pt},
        xlabel = {SigLIP$^\dagger$ (mAP@1k=28.9)},
        ylabel = {SigLIP (mAP@1k=19.6)},
        xlabel style={yshift=2pt},
        ylabel style={yshift=-3pt},
        yticklabel style={font=\scriptsize},
        xticklabel style=
    {/pgf/number format/1000 sep=,anchor=north,font=\scriptsize},
        legend pos=north west,
        legend style={cells={anchor=east}, font=\scriptsize, fill opacity=0.7, row sep=-1pt},
    ]
    \addplot[only marks, mark=*, opacity=0.7, mark size=0.7, color=appleblue] 
    table[x=vit_large_patch16_siglip_384_webli, y=dinov2_vitl14] {./data/pairwise_siglip-dino.csv};
    \end{axis}
\end{tikzpicture}

&

\hspace{-30pt}
\begin{tikzpicture}
    \footnotesize
    \begin{axis}[
        width=0.57\linewidth,
        height=0.57\linewidth,
        xmin=-1,
        xmax=101,
        ymin=-1,
        ymax=101,
        title={\footnotesize i2i vs t2i (0.51)},
        grid=both,
        grid style={color=lightgray!60, dash pattern=on 2pt off 2pt},
        xtick={0, 20, 40, 60, 80, 100},
        ytick={0, 20, 40, 60, 80, 100},
        title style = {yshift = -4pt},
        xlabel = {SigLIP-text (mAP@1k=18.1)},
        ylabel = {SigLIP (mAP@1k=19.6)},
        xlabel style={yshift=0.3pt},
        ylabel style={yshift=-3pt},
        yticklabel style={font=\scriptsize},
        xticklabel style=
    {/pgf/number format/1000 sep=,anchor=north,font=\scriptsize},
        legend pos=north west,
        legend style={cells={anchor=east}, font=\scriptsize, fill opacity=0.7, row sep=-1pt},
    ]
    \addplot[only marks, mark=*, opacity=0.7, mark size=0.7, color=appleblue] 
    table[x=Siglip-text, y=Siglip-image] {./data/pairwise_image-text.csv};
    \end{axis}
\end{tikzpicture}

\\

\hspace{-11.7pt}
\begin{tikzpicture}
    \footnotesize
    \begin{axis}[
        width=0.57\linewidth,
        height=0.57\linewidth,
        xmin=-1,
        xmax=101,
        ymin=-1,
        ymax=101,
        title={\footnotesize global vs re-ranking w/ global  (0.92)},
        grid=both,
        grid style={color=lightgray!60, dash pattern=on 2pt off 2pt},
        xtick={0, 20, 40, 60, 80, 100},
        ytick={0, 20, 40, 60, 80, 100},
        title style = {yshift = -4pt},
        xlabel = {SigLIP$^\dagger$+$\alpha$QE1 (mAP@1k=33.7)},
        ylabel = {SigLIP$^\dagger$ (mAP@1k=28.9)},
        xlabel style={yshift=2pt},
        ylabel style={yshift=-3pt},
        yticklabel style={font=\scriptsize},
        xticklabel style=
    {/pgf/number format/1000 sep=,anchor=north,font=\scriptsize},
        legend pos=north west,
        legend style={cells={anchor=east}, font=\scriptsize, fill opacity=0.7, row sep=-1pt},
    ]
    \addplot[only marks, mark=*, opacity=0.7, mark size=0.7, color=appleblue] 
    table[x=Siglip-qe, y=siglip] {./data/pairwise_siglip-siglipqe.csv};
    \end{axis}
\end{tikzpicture}

&

\hspace{-25pt}
\begin{tikzpicture}
    \footnotesize
    \begin{axis}[
        width=0.57\linewidth,
        height=0.57\linewidth,
        xmin=-1,
        xmax=101,
        ymin=-1,
        ymax=101,
        title={\footnotesize global vs re-ranking w/ local (0.74)},
        grid=both,
        grid style={color=lightgray!60, dash pattern=on 2pt off 2pt},
        xtick={0, 20, 40, 60, 80, 100},
        ytick={0, 20, 40, 60, 80, 100},
        title style = {yshift = -4pt},
        xlabel = {SigLIP$^\dagger$+AMES (mAP@1k=35.6)},
        ylabel = {SigLIP$^\dagger$ (mAP@1k=28.9)},
        xlabel style={yshift=2pt},
        ylabel style={yshift=-3pt},
        yticklabel style={font=\scriptsize},
        xticklabel style=
    {/pgf/number format/1000 sep=,anchor=north,font=\scriptsize},
        legend pos=north west,
        legend style={cells={anchor=east}, font=\scriptsize, fill opacity=0.7, row sep=-1pt},
    ]
    \addplot[only marks, mark=*, opacity=0.7, mark size=0.7, color=appleblue] 
    table[x=SigLIP+AMES, y=SigLIP] {./data/pairwise_siglip-siglipames.csv};
    \end{axis}
\end{tikzpicture}

\end{tabular}
  }
  \vspace{-14pt}
  \caption{\textbf{Performance comparison reporting AP per query for different approaches with SigLIP.} Pearson correlation reported in parenthesis. $\dagger$ indicates results with the linear adaptation.}
  \label{fig:pairwise}
  \vspace{-16pt}
\end{figure}
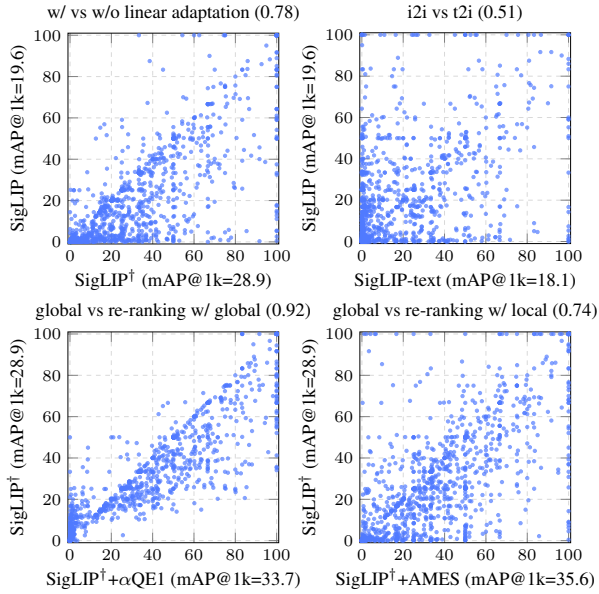

\subsection{Analysis}
\label{sec:analysis}

\noindent\textbf{Performance per domain.} Fig.~\ref{fig:subdomains} shows the performance per taxonomy categories. The taxonomy annotations allow a fine-grained view of the results, which can possibly allow us to capture imbalanced improvements in future work.
For example, DINOv2, despite being overall inferior to SigLIP, is outperforming it in categories like architecture and sculptures or is quite similar in categories like public art and paper art.
This is possibly attributed to the curation and composition of the DINOv2 training set, which includes artwork and landmark datasets. Also, some categories are hurt by re-ranking with AMES, with some demonstrating big drops, \ie sport, gaming, perfume. These categories deviate significantly from the domain AMES is trained, \ie landmarks, which could potentially justify such drops.

\noindent\textbf{Per query comparisons.} Fig.~\ref{fig:pairwise} shows the AP per query for various methods. Linear adaptation boosts most queries, \ie performance drop only for 192 queries. Image- and text-based retrieval are not strongly correlated despite performing similarly, which is good evidence~\cite{tky+24} for the effectiveness of model ensembles. Indeed, ensembling i2i and t2i by averaging similarities brings +6.1 improvement over i2i retrieval. Query expansion improves the queries with at least some positives at top positions, \ie AP greater than 20. However, it harms many low-performing queries by aggregating descriptors irrelevant to the query. AMES improves the majority of the queries; however, many are harmed, indicating that there is plenty of room for improvement.

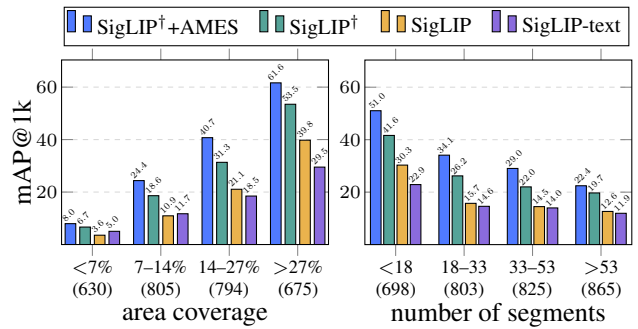
\begin{figure}[t]
    \vfill
    \centering
    \begin{tabular}{cc}
        \hspace{-10pt}
        \begin{subfigure}{0.48\textwidth}
            \begin{tikzpicture}
\begin{axis}[
    width=0.61\linewidth,
    height=0.48\linewidth,
    bar width=4pt, 
    ylabel={mAP@1k},
    xlabel={area coverage},
    ymin=0, ymax=67,
    ytick={20,40,60,80}, 
    ytick distance=5, 
    ybar=1.5,
    enlarge y limits={upper, value=0.05},
    xtick=data,
    symbolic x coords={  
        $<$7\%,
        7--14\%,
        14--27\%,
        $>$27\%,
    },
    ylabel style={yshift=-5pt},
    xlabel style={yshift=-6pt},
    xtick distance = 1,
    xticklabel style={
        font=\scriptsize,
        yshift=2pt,
        },
    yticklabel style={
        font=\scriptsize,
        xshift=2pt,
        },
    enlarge x limits=0.15, 
    nodes near coords,
    every node near coord/.append style={
        scale=.5, 
        anchor=south west, 
        xshift=-3.5pt,
        yshift=-3.5pt,
        color=black,
        rotate=45,
        font=\scriptsize, 
        /pgf/number format/.cd, 
        fixed zerofill, 
        precision=1,
    },
    legend style={
        at={(0.0,1.3)}, 
        anchor=north west, 
        legend columns=-1,  
        /tikz/every even column/.append style={column sep=0.2cm}, 
        font=\footnotesize,
        inner sep=1pt,  
        outer sep=1pt 
    },
    legend cell align={left}, 
    ymajorgrids=true,
    grid style={color=lightgray!60, dash pattern=on 2pt off 2pt},
]

\pgfplotstableread[col sep=comma]{data/performance_per_scale.csv}\loadeddata

\addplot+[ybar, color=black, fill=appleblue] table[x=domain,y=avg_ap_ames_siglip_384] {\loadeddata};
\addplot+[ybar, color=black, fill=appleteal] table[x=domain,y=avg_ap_siglip_384_proj] {\loadeddata};
\addplot+[ybar, color=black, fill=applemustard] table[x=domain,y=avg_ap_siglip_384_nw] {\loadeddata};
\addplot+[ybar, color=black, fill=applepurple] table[x=domain,y=avg_ap_siglip_384_text] {\loadeddata};

\legend{ SigLIP$^\dagger$+AMES,  SigLIP$^\dagger$, SigLIP, SigLIP-text, DINOv2$^\dagger$}
    \end{axis}

\foreach \X/\value in {0.41/(630), 1.313/(805), 2.215/(794), 3.13/(675)} {
    \node[anchor=north, font=\scriptsize] at (\X,-0.375cm) {\scalebox{1.0}{\value}};
}
\end{tikzpicture}
        \end{subfigure}
        &
        \hspace{-130pt}
        \begin{subfigure}{0.48\textwidth}
            \begin{tikzpicture}
\begin{axis}[
    width=0.61\linewidth,
    height=0.48\linewidth,
    bar width=4pt, 
    xlabel={number of segments},
    ymin=0, ymax=67,
    ytick={20,40,60,80}, 
    ytick distance=5, 
    ybar=1,
    enlarge y limits={upper, value=0.05},
    xtick=data,
    symbolic x coords={  
        $<$18,
        18--33,
        33--53,
        $>$53,
    },
    ylabel style={yshift=-5pt},
    xlabel style={yshift=-4pt},
    xtick distance = 1,
    xticklabel style={
        font=\scriptsize,
        yshift=2pt,
        },
    yticklabel style={
        font=\scriptsize,
        xshift=2pt,
        },
    enlarge x limits=0.15, 
    nodes near coords,
    every node near coord/.append style={
        scale=.5, 
        anchor=south west, 
        xshift=-2.5pt,
        yshift=-4pt,
        color=black,
        rotate=45,
        font=\scriptsize, 
        /pgf/number format/.cd, 
        fixed zerofill, 
        precision=1,
    },
    legend style={
        at={(0.0,1.2)}, 
        anchor=north west, 
        legend columns=-1,  
        /tikz/every even column/.append style={column sep=0.2cm}, 
        font=\footnotesize,
        inner sep=1pt,  
        outer sep=1pt 
    },
    legend cell align={left}, 
    ymajorgrids=true,
    grid style={color=lightgray!60, dash pattern=on 2pt off 2pt},
]

\pgfplotstableread[col sep=comma]{data/performance_per_clutter.csv}\loadeddata

\addplot+[ybar, color=black, fill=appleblue] table[x=domain,y=avg_ap_ames_siglip_384] {\loadeddata};
\addplot+[ybar, color=black, fill=appleteal] table[x=domain,y=avg_ap_siglip_384_proj] {\loadeddata};
\addplot+[ybar, color=black, fill=applemustard] table[x=domain,y=avg_ap_siglip_384_nw] {\loadeddata};
\addplot+[ybar, color=black, fill=applepurple] table[x=domain,y=avg_ap_siglip_384_text] {\loadeddata};

\end{axis}

\foreach \X/\value in {0.41/(698), 1.313/(803), 2.215/(825), 3.13/(865)} {
    \node[anchor=north, font=\scriptsize] at (\X,-0.375cm) {\scalebox{1.0}{\value}};
}

\end{tikzpicture}
            \vspace{-0.3pt}
        \end{subfigure}
    \end{tabular}
    \vspace{-10pt}
    \caption{\textbf{Performance evaluation (mAP@1k) across different amounts of object area coverage and background clutter.} Positives across all queries are jointly ranked based on coverage or clutter and split into 4 equal size groups. 
    Queries with no positive in the corresponding group are discarded. No. of queries per group is in parentheses.
    $\dagger$ indicates results with the linear adaptation.
    \vspace{-8pt}
    \label{fig:scale_clutter_groups}}
\end{figure}

\noindent\textbf{Impact of clutter and scale in positives.}
To quantify the impact of background clutter and scale changes, Fig.~\ref{fig:scale_clutter_groups} presents the performance for different groups of positives. Dealing with small objects and multi-object scenes form major weaknesses of existing models. Notably, t2i beats i2i without adaptation in small-scale groups.

\section{Conclusions}
We introduce \ours and conduct an extensive evaluation of current foundational models and retrieval methods, highlighting that instance-level retrieval remains an unsolved problem. 
Our results indicate that off-the-shelf application of foundational models leaves considerable room for improvement, particularly in handling small objects and complex backgrounds. 
While specialized retrieval methods leveraging local descriptors are effective in these cases, their high memory and computational costs become impractical at the scale of \ours or beyond. 
\ours is designed to become a standard benchmark for evaluating foundational representation models and retrieval methods, accommodating both global and local representations, and advancing the field of instance-level recognition.

\newpage

\begin{center}
\Large \textbf{Supplementary materials}
\vspace{25pt}
\end{center}

\renewcommand\thesection{\Alph{section}}

\setcounter{section}{0}
\setcounter{table}{0}
\setcounter{figure}{0}

\appendix
\renewcommand{\thefigure}{\Alph{figure}}
\renewcommand{\thetable}{\Alph{table}}

\section{Implementation details}

\begin{figure}[t]
    \centering
    \centering
\scalebox{0.97}{
\begin{tabular}{ccc}
    \hspace{-23pt}
    \begin{subfigure}{0.2\textwidth}
        \centering
        \includegraphics[height=2.8cm, width=2.8cm]{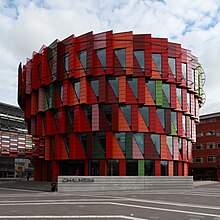}
        \caption{\label{fig:rej_building}}
    \end{subfigure}
    &
    \hspace{-30pt}
    \begin{subfigure}{0.2\textwidth}
        \centering
        \includegraphics[height=2.8cm, width=2.8cm]{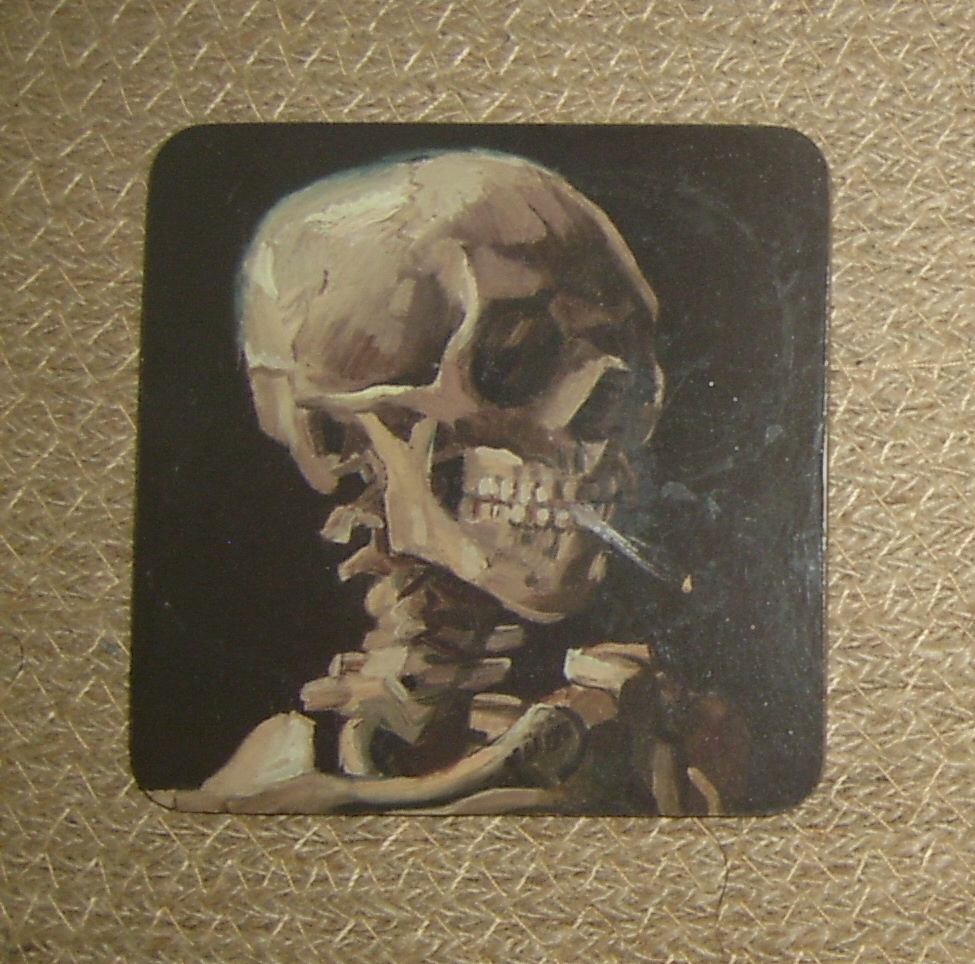}
        \caption{\label{fig:rej_coaster}}
    \end{subfigure}
    &
    \hspace{-30pt}
    \begin{subfigure}{0.2\textwidth}
        \centering
        \includegraphics[height=2.8cm, width=2.8cm]{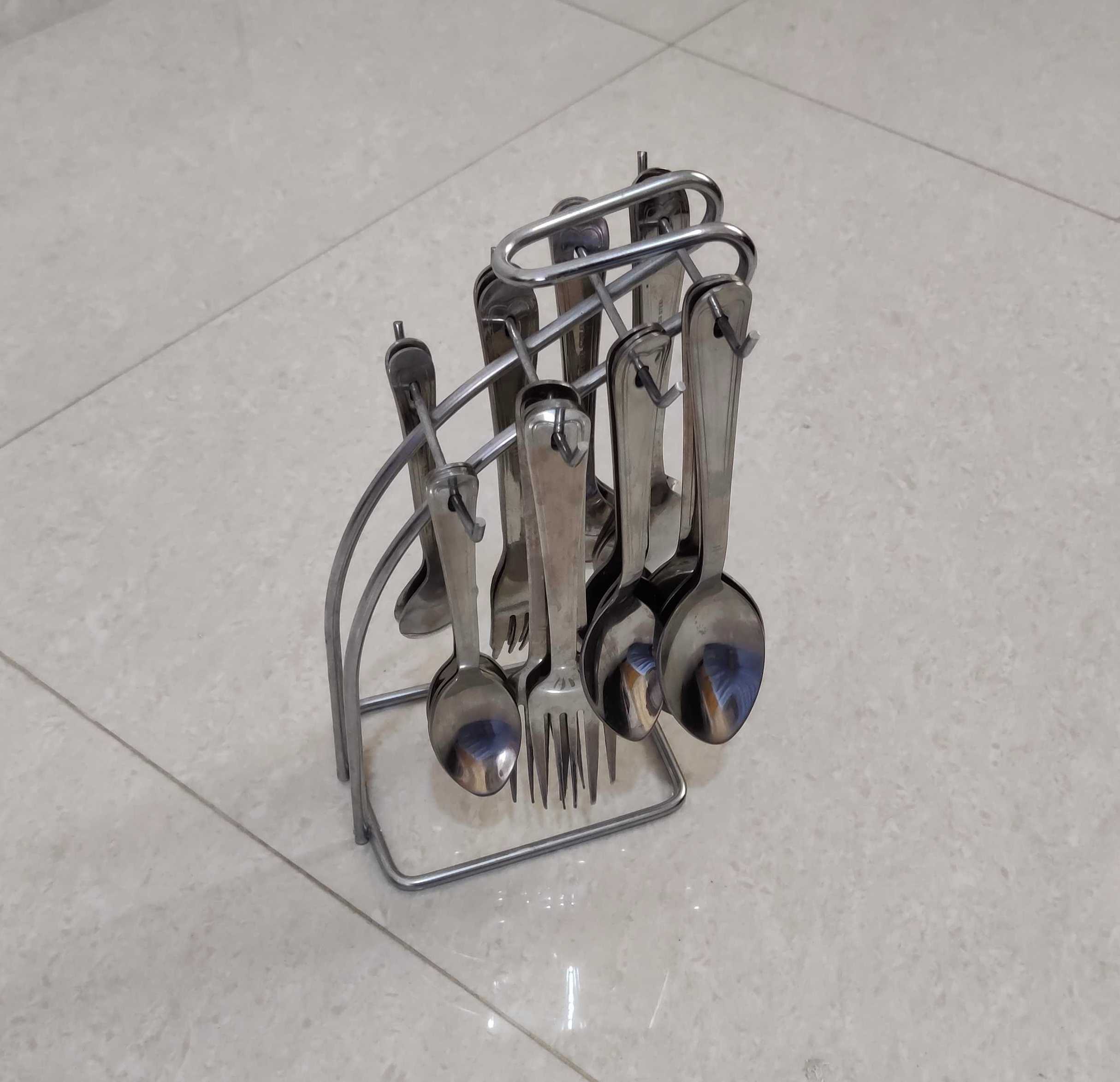}
        \caption{\label{fig:rej_cutlery}}
    \end{subfigure}
\end{tabular}
}
    \vspace{-5pt}
    \caption{\textbf{Rejected objects.} Example of objects that are disregarded during the selection process.
    \label{fig:rejected}
    \vspace{5pt}
    }
\end{figure}

\medskip\noindent{\textbf{Collection process}}. 
Queries and positives are created/collected by a group of 16 collectors who are well-informed about the task objectives. 
Most of the images consist of photographs taken by the collectors for the purpose of this work, while a smaller part is downloaded from online repositories with a permissive license.
All collected images are manually filtered and curated by the authors. Regarding the selection of the objects, the collectors are advised to opt for objects with distinct, uncommon features--such as unique shapes, colors, or textures--that set them apart within their category, \ie prioritize items with rare modifications. 
As mentioned in the main paper, objects that are created or share parts with other objects created before 2014 do not qualify as query objects. Fig.~\ref{fig:rejected} illustrates some of the objects rejected during the selection process. Fig.~\ref{fig:rej_building} is the Kuggen building, whose construction finished in 2011. Fig.~\ref{fig:rej_coaster} is a newly bought coaster that displays a well-known van Gogh painting. Fig.~\ref{fig:rej_cutlery} is a newly bought cutlery holder whose design is rather generic with no distinctive detail; hence, very similar (close to identical) objects may exist in YFCC100M. 
Furthermore, the collectors are provided with older camera models used in YFCC100M. This simulates similar camera distribution for the query and positive images with the distractors. Tab.~\ref{tab:cameras} shows the distribution of the most used cameras. Older-generation cameras are used for the majority of the collected images.
The collectors are instructed to avoid using the same camera for both the query and the positives of an object to avoid any possible shortcuts learned by pre-trained models.

\medskip\noindent\textbf{Downloading and storing images.}
To acquire the YFCC100M~\cite{tsf+16}, we download images based on the Flickr URLs provided by the original authors.
Approximately $\sim$82M images are downloaded. The remaining images are downloaded from the AWS S3 data bucket provided by the authors.
We opt for downloading the images from Flickr to ensure that identical preprocessing has been applied to the distractor dataset and the collected query and positive sets in \ours.
The collected images in \ours are also uploaded to and downloaded from Flickr. We use the ``medium'' option to download all images, which resizes images to 500px based on their larger side.
All images are stored with 90 JPEG compression quality with 4:4:4 chroma subsampling. Following~\cite{gonzalez09,ab19}, white balancing is applied on all images.
All personal details (\eg human faces, license plates) that are displayed in the collected images of \ours are either blurred or cropped.

\begin{table}[t]
  \centering
  \scalebox{0.99}{
    \begin{tabular}{lccr}
\toprule
\textbf{model} & \textbf{year} & \textbf{type} & \textbf{images} \\ 
\midrule
Canon EOS 450D             & 2008 & DSLR   & 770 \\
NIKON D3000                & 2009 & DSLR   & 585 \\
NIKON 3100                 & 2010 & DSLR   & 443 \\
DiMAGE X1                  & 2005 & camera & 286 \\
Xiaomi Poco X5 Pro         & 2023 & phone  & 275 \\
iPhone 14                  & 2022 & phone  & 237 \\
Xiaomi Redmi Note 11 Pro   & 2022 & phone  & 210 \\
Canon EOS 6D Mark II       & 2017 & DSLR   & 208 \\
iPhone SE (3rd generation) & 2022 & phone  & 195 \\
NIKON 5300                 & 2013 & DSLR   & 144 \\
Canon EOS 50D              & 2008 & DSLR   & 141 \\
iPhone 14 Pro              & 2022 & phone  & 122 \\
Canon PowerShot S5 IS      & 2007 & DSLR   & 118 \\
ONEPLUS A6003              & 2018 & phone  & 110 \\
Canon EOS REBEL T2i        & 2010 & DSLR   & 102 \\
\bottomrule
\end{tabular}
  }
  \vspace{-5pt}
  \caption{\textbf{Most frequently used camera models in \ours.} Cameras used for more than 100 images are displayed. Information about release date and type of camera is provided.
  \label{tab:cameras}
  \vspace{-5pt}
  }
\end{table}

\medskip\noindent\textbf{\miniours composition.} We consider the 88 text category labels from \ours taxonomy to generate text queries, manually expanded with 132 terms that are synonyms or fine-grained descriptions of the original labels. The collected labels are combined with 43 templates used in the original CLIP~\cite{clip} to generate a list of 9,976 text queries. Examples of the templates used are in Tab.~\ref{tab:templates}. We do not consider domain-specific templates. We use the large model variants of SigLIP, OpenCLIP, and EVA-CLIP to compute the ensemble text-image similarities between the text queries and each image of YFCC100M.

\medskip\noindent\textbf{Text query generation.} 
We generate text queries using the GPT-4o~\cite{chatgpt}. The prompt displayed in Fig.~\ref{fig:chagpt_prompt} is first provided to the LLM. 
Then, a query image of one of the objects in \ours and its corresponding category is provided to the model to generate a textual description. For object category, we use mid level category from taxonomy. If it is not available, we use the coarser level category. The generated text queries are manually edited by the authors to fix errors, insufficient descriptions, or nuances of the model.

\begin{table}[t]
    \centering
    \begin{tabular}{l}
        \texttt{a close-up photo of the *.}\\
        \texttt{a good photo of the *.}\\
        \texttt{a photo of a cool *.}\\
        \texttt{a low resolution photo of the *.}\\
        \texttt{a bad photo of the *.}\\
        \texttt{a cropped photo of the *.}\\
        \texttt{a photo of a hard to see *.}\\
        \texttt{a bright photo of a *.}\\
        \texttt{a photo of a clean *.}\\
        \texttt{a photo of a dirty *.}\\
        \texttt{a dark photo of the *.}\\
        \texttt{a photo of my *.}\\
        \texttt{a photo of the cool *.}\\
        \texttt{a close-up photo of a *.}\\
        \texttt{a bright photo of the *.}
    \end{tabular}
    \vspace{-6pt}
    \caption{\textbf{Examples of templates} used for the text query generation for the creation of \miniours. The \texttt{*} symbol is replaced with a taxonomy term.
    \label{tab:templates}
    \vspace{-9pt}
    }
\end{table}

\medskip\noindent\textbf{Global representations.}
For the implementation of global representation models, we rely on public resources available on PyTorch~\cite{pgm+19}. We use the timm\footnote{\rurl{github.com/rwightman/pytorch-image-models}} and torchvision\footnote{\rurl{github.com/pytorch/vision}} libraries that provide relevant code and weights for the majority of the models. For the models not included there, we use the relevant code from the official github repositories provided by the authors, \ie
\cite{odm+24}\footnote{\rurl{github.com/facebookresearch/dinov2}},
\cite{ctm+21}\footnote{\rurl{github.com/facebookresearch/dino}},
\cite{cmm+20}\footnote{\rurl{github.com/facebookresearch/swav}},
\cite{hfw+20}\footnote{\rurl{github.com/facebookresearch/moco-v3}},
\cite{ptm+22}\footnote{\rurl{github.com/yash0307/recallatk_surrogate}},
\cite{kjk23}\footnote{\rurl{github.com/tjddus9597/hier-cvpr23}},
\cite{lsl+22}\footnote{\rurl{github.com/sungonce/cvnet}},
\cite{sck+23}\footnote{\rurl{github.com/shihaoshao-gh/superglobal}},
\cite{ady+23}\footnote{\rurl{github.com/deepglint/unicom}},
\cite{swl+24}\footnote{\rurl{github.com/naver/unic}},
\cite{ycc+23}\footnote{\rurl{github.com/nikosips/universal-image-embeddings}},
\cite{yca+24}\footnote{\rurl{github.com/nikosips/udon}}. Model weights that are not publicly available are provided to us by the original authors. 
For t2i, we use the image encoders from timm and the text encoders from huggingface\footnote{\rurl{huggingface.co}} and OpenCLIP\footnote{\rurl{github.com/mlfoundations/open_clip}}. We include only base and large model variants in our benchmark. Tab.~\ref{tab:supp_all_models} and~\ref{tab:supp_all_text_models} contain more information, including model checkpoints. Regarding image preprocessing, following instance-level retrieval literature~\cite{sck+23,rtc19,lsl+22}, the images are resized based on their largest side respecting their aspect ratio, \ie isotropic rescaling. Image resolution is dictated by each model's specifications together with the rule setting resolution one level higher than those used during training. This rule is empirically created based on experiments presented in Sec.~\ref{sec:addan}. We normalize the image tensors with the mean and standard deviation statistics according to model specifications. For all ViT-based models, bicubic interpolation of the position embeddings is performed. Unicom~\cite{ady+23} requires fixed-size tensors in the backbone output, which goes through a projection head; hence, we use adaptive average pooling to fix the spatial dimensions of the output feature tensors. For UDON~\cite{yca+24} and USCRR~\cite{ycc+23} models, we use the representation before projection due to the low dimensionality of the latter. For AlexNet~\cite{ksh12} and VGG~\cite{sz14} models, we extract descriptors based on the feature maps of the last convolutional layer by applying GeM pooling~\cite{rtc19}. For the rest of the models, the extraction process used in the original methods is employed. All global descriptors are \l2 normalized.

\begin{figure}[t]
    \centering
    \begin{tcolorbox}[colback=gray!15, colframe=black, width=\linewidth]
        \hspace{-8pt}
        \begin{tabular}{p{1.03\linewidth}}
            {\scriptsize \texttt{You are a system generating descriptions of objects shown in an image.}} \\
            {\scriptsize \texttt{Provided with an image and a category in which the item shown in the image belongs to, you will describe the main item that you see in the image, giving enough details to unambiguously describe the object.}} \\
            {\scriptsize \texttt{You can describe unambiguously what the item is and its material, color, and style if clearly identifiable.}} \\
            {\scriptsize \texttt{Please do not describe anything about the background.}}
        \end{tabular}
    \end{tcolorbox}
    \vspace{-10pt}
    \caption{\textbf{Prompt} used for the initial generation of text queries.
    \label{fig:chagpt_prompt}
    \vspace{-6pt}
    }
\end{figure}

\noindent\textbf{Linear adaptation.}
The single linear adaptation layer is trained on a 1M random subset of UnED~\cite{ycc+23}.
The training follows the UJCDS~\cite{ycc+23} method that learns a linear classifier on all classes in the UnED subset (191,513 classes). The classifier gets the \l2 normalized features output from the linear adaptation layer.
During training, the Normalized Softmax loss~\cite{zw18} is minimized, and no balancing across UnED domains is performed.
The linear layer and classifier are trained for 2 epochs with 128 batch size. We use Adam~\cite{kb15} optimizer with 10\textsuperscript{-3} learning rate and 10\textsuperscript{-6} weight decay. The scale of the Normalized Softmax loss is 16.

\begin{figure}[t]
    \centering
    \pgfplotsset{every tick label/.append style={font=\scriptsize}}
\begin{tikzpicture}
    \begin{semilogxaxis}[
        width=1\linewidth,
        height=0.7\linewidth,
        xlabel={number of distractors (in millions)},
        ylabel={mAP@1k},
        xlabel style={yshift=3pt},
        ylabel style={yshift=-4pt},
        xmin=90000, xmax=100000000,
        ymin=10, ymax=70,
        grid=both,
        grid style={color=lightgray!60, dash pattern=on 2pt off 2pt},
        log basis x=10,
        ytick={10, 30, 50, 70},
        xtick={100000, 1000000, 10000000, 100000000},
        xticklabels={0, 1, 10, 100}, 
        extra x ticks={100000, 200000, 300000, 400000, 500000, 600000, 700000, 800000, 900000, 1000000, 2000000, 3000000, 4000000, 5000000, 6000000, 7000000, 8000000, 9000000, 10000000, 20000000, 30000000, 40000000, 50000000, 60000000, 70000000, 80000000, 90000000, 100000000},
        extra x tick labels={},
        legend columns=2,
        legend style={
            anchor=north east, 
            at={(1.05,1.13)}, 
            cells={anchor=west}, 
            font =\scriptsize, 
            fill opacity=1pt, 
            inner sep=1pt
            },
    ]

        \addplot[color=applemustard, mark=*, mark size=2.5, opacity=1.0, mark options={draw=black, line width=0.35pt}, line width=1.2pt] coordinates {
            (100000, 57.17) (1000000, 35.42) (5000000, 27.04) (10000000, 23.92)
            (50000000, 17.41) (100000000, 15.25)
        };
        \addlegendentry{OAI-CLIP$^\dagger$}

        \addplot[color=appleteal, mark=*, mark size=2.5, opacity=1.0, mark options={draw=black, line width=0.35pt}, line width=1.2pt] coordinates {
            (100000, 46.86) (1000000, 27.88) (5000000, 22.48) (10000000, 20.46)
            (50000000, 16.91) (100000000, 15.32)
        };
        \addlegendentry{DINOv2$^\dagger$}

        \addplot[color=appleorange, mark=*, mark size=2.5, opacity=1.0, mark options={draw=black, line width=0.35pt}, line width=1.2pt] coordinates {
            (100000, 58.79) (1000000, 37.33) (5000000, 28.61) (10000000, 24.97)
            (50000000, 18.45) (100000000, 16.01)
        };
        \addlegendentry{EVA-CLIP$^\dagger$}

        \addplot[color=applegreen, mark=*, mark size=2.5, opacity=1.0, mark options={draw=black, line width=0.35pt}, line width=1.2pt] coordinates {
            (100000, 56.96) (1000000, 37.13) (5000000, 29.21) (10000000, 25.89)
            (50000000, 19.31) (100000000, 16.94)
        };
        \addlegendentry{MetaCLIP$^\dagger$}

        \addplot[color=applepink, mark=*, mark size=2.5, opacity=1.0, mark options={draw=black, line width=0.35pt}, line width=1.2pt] coordinates {
            (100000, 53.29) (1000000, 32.02) (5000000, 24.51) (10000000, 21.62)
            (50000000, 15.93) (100000000, 13.98)
        };
        \addlegendentry{OpenCLIP$^\dagger_b$}

        \addplot[color=appleblue, mark=*, mark size=2.5, opacity=1.0, mark options={draw=black, line width=0.35pt}, line width=1.2pt] coordinates {
            (100000, 59.80) (1000000, 39.13) (5000000, 30.92) (10000000, 27.46)
            (50000000, 20.72) (100000000, 18.26)
        };
        \addlegendentry{OpenCLIP$^\dagger_l$}
        
        \addplot[color=appleyellow, mark=*, mark size=2.5, opacity=1.0, mark options={draw=black, line width=0.35pt}, line width=1.2pt] coordinates {
            (100000, 61.24) (1000000, 42.33) (5000000, 34.27) (10000000, 31.25)
            (50000000, 25.12) (100000000, 22.96)
        };
        \addlegendentry{SigLIP$^\dagger_b$}

        \addplot[color=applepurple, mark=*, mark size=2.5, opacity=1.0, mark options={draw=black, line width=0.35pt}, line width=1.2pt] coordinates {
            (100000, 65.85) (1000000, 48.92) (5000000, 41.72) (10000000, 38.34)
            (50000000, 31.64) (100000000, 28.95)
        };
        \addlegendentry{SigLIP$^\dagger_l$}

    \end{semilogxaxis}
\end{tikzpicture}
    \vspace{-10pt}
    \caption{\textbf{Impact of the number of distractors.} mAP@1k of five models for varying db size. $\dagger$ indicates results with the linear adaptation. $b$ and $l$: base and large model variants.
    \label{fig:number_of_distractors}
    \vspace{-13pt}
    }
\end{figure}
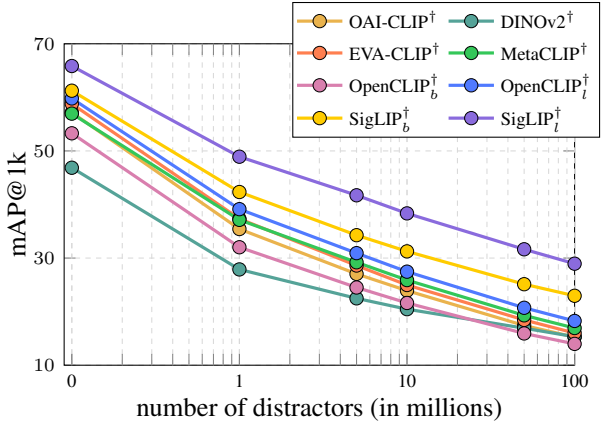

\noindent\textbf{Local representations.}
Following AMES~\cite{ski+24}, local descriptors are extracted based on the base variant of DINOv2 with registers~\cite{odm+24,doj+23}. Local descriptors are selected based on their weights estimated by a feature detector~\cite{cas20}. We use the pre-trained network trained on the corresponding descriptors. The local descriptor extraction, the pre-trained models, and inference configurations are publicly-available\footnote{\rurl{github.com/pavelsuma/ames}}. To ensure a fair comparison between re-ranking methods, we use the same local descriptors for other methods but with  different binarization. AMES consists of a binarization layer initialized with ITQ~\cite{glg+12,ktp+22} and finetuned during model training. Hence, for Chamfer Similarity (CS)~\cite{rsa+14} and Spatial verification (SP)~\cite{pci+07}, the descriptors are binarized with the same ITQ weights.
For SP, we follow the standard practice in retrieval with fast spatial matching~\cite{cas20} and use single correspondence hypotheses, which is translation in our case, and LO-RANSAC~\cite{cmk03} for affine-transformation. Due to the single scale local descriptors, departing from the single correspondence hypothesis and sampling correspondence pairs or triplets can potentially provide better results despite being slower.
Tentative inlier correspondences are extracted based on the nearest neighbor of each query local descriptor, using a threshold of 32 Hamming distance. Local similarity for re-ranking is estimated based on the number of inliers detected by RANSAC, with a minimum threshold of 5 inliers.
Also, the final AMES similarity is an ensemble of local and global similarity. For a fair comparison, the same ensembling scheme is also used for CS and SP, following the same validation process. The ensemble hyper-parameters are tuned on the public split of the GLDv2 dataset.
In the default settings, \ie 100 binarized local descriptors for db images, the total memory requirements for storing local descriptors is $\sim$149GB, which is to be compared with $\sim$95GB needed for 512D global descriptors stored in half-precision.
Note that we do not consider compression techniques for the global descriptors, which can decrease the memory footprint by an order of magnitude with an insignificant performance loss~\cite{gar+17, ski+24}.

\section{Additional experiments}
Similar to the main paper, unless stated otherwise, we use the large ViT model variants with the largest resolution available, \eg we use SigLIP ViT-L trained with 384 resolution. In the case of various architectures for the same method, we use the best-performing one, \eg we use the large variant of ConvNext architecture for OpenCLIP.

\subsection{Additional analysis}
\label{sec:addan}

\begin{table}[t]
  \centering
  \scalebox{0.95}{
  \hspace{-7pt}
  \newcolumntype{C}{>{\raggedleft\arraybackslash}p{2em}}
\small
\begin{tabular}{lrCCCC}
\toprule
\textbf{model} & \textbf{train res} & \textbf{224} & \textbf{384} & \textbf{512} & \textbf{724} \\ 
\midrule
EVA-CLIP~\cite{evaclip,fwx+23}  & 224 & 5.0  & \underline{7.7}  & 5.8  &  3.1 \\
MetaCLIP~\cite{metaclip}        & 224 & 5.1  & \underline{8.8}  & 6.5  &  3.8 \\
OpenCLIP~\cite{convnext,iww+21} & 224 & 8.2  & \underline{10.7} & 6.1  &  2.5 \\
DINOv2~\cite{odm+24}            & 518 & 6.4  & 12.2 & 14.0 & \underline{14.3} \\
SigLIP~\cite{siglip}            & 224 & 9.1  & \underline{14.1} & 10.2 &  6.1 \\
SigLIP~\cite{siglip}            & 512 & 0.1  &  8.9 & 18.9 & \underline{20.2} \\ 
\midrule
EVA-CLIP~\cite{evaclip,fwx+23}  & 336 & 4.7  & 13.1 & \underline{13.4} &  9.5 \\
MetaCLIP~\cite{metaclip}        & 224 & 10.3 & \underline{14.4} & 11.0 &  7.4 \\
OpenCLIP~\cite{convnext,iww+21} & 320 & 10.3 & 16.5 & \underline{12.7} &  7.6 \\
DINOv2~\cite{odm+24}            & 518 & 9.7  & 16.0 & 17.7 & \underline{18.5} \\
SigLIP~\cite{siglip}            & 256 & 8.3  & \underline{18.8} & 15.4 & 10.8 \\ 
SigLIP~\cite{siglip}            & 384 & 1.8  & 21.6 & \underline{24.1} & 20.6 \\ 
\bottomrule
\end{tabular}}
  \vspace{-5pt}
  \caption{\textbf{Impact of resolution.} Performance (mAP@1k) by testing at different resolutions. The underline indicates the resolution selected for each model based on our rule. Linear adaptation is not used. Top: base models. Bottom: large models.
  \label{tab:resolution}
  \vspace{-10pt}
  }
\end{table}

\begin{figure*}[t]
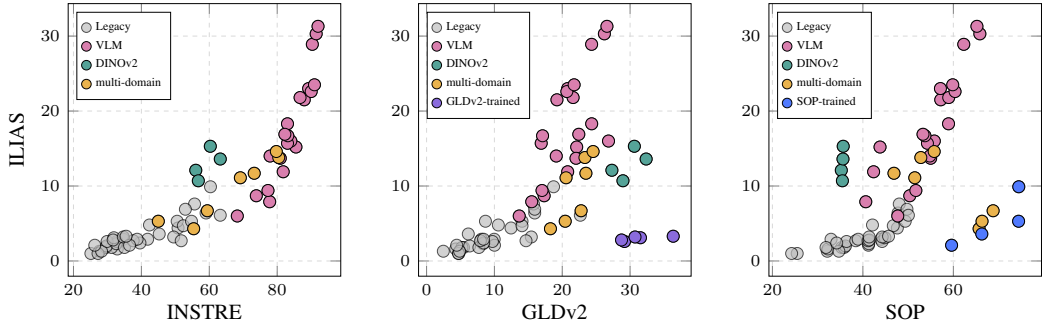

  \centering
  \vspace{-15pt}
  \scalebox{0.9}{
    \hspace{10pt}
    \begin{tikzpicture}
\begin{axis}[
  width=.32\linewidth,
  height=.32\linewidth,
  xlabel={\small INSTRE},
  ylabel={\small ILIAS},
  tick label style={font=\scriptsize},
  ylabel near ticks, xlabel near ticks, 
  xlabel style={yshift=3pt},
  grid=both,
  grid style={color=lightgray!60, dash pattern=on 2pt off 2pt},  
  legend entries={Legacy, VLM, DINOv2, multi-domain},
  legend pos=north west,
  legend style={cells={anchor=west}, font=\tiny, fill opacity=1.0, row sep=-1pt, inner sep=2pt},
  legend image post style={scale=0.8},
  ]

  \input{data/instre_gld_sop_adapt}

   \addplot[only marks, mark options={draw=black}, color=lightgray, solid, mark=*, opacity=0.7, mark size=2.5] table[y=ILIAS, x=INSTRE] \instregldsop;

  \addplot[only marks, mark options={draw=black}, color=applepink, solid, mark=*, mark size=2.5] table[y=ILIAS, x expr={\thisrow{INSTRE} / (\thisrow{show} == 1)}] \instregldsop;

  \addplot[only marks, mark options={draw=black}, color=appleteal, solid, mark=*, mark size=2.5] table[y=ILIAS, x expr={\thisrow{INSTRE} / (\thisrow{show} == 2)}] \instregldsop;

  \addplot[only marks, mark options={draw=black}, color=applemustard, solid, mark=*, mark size=2.5] table[y=ILIAS, x expr={\thisrow{INSTRE} / (\thisrow{show} == 5)}] \instregldsop;

\end{axis}
\end{tikzpicture}
    \hspace{10pt}
    \begin{tikzpicture}
\begin{axis}[
  width=.32\linewidth,
  height=.32\linewidth,
  xlabel={\small GLDv2},
  xmax=39,
  tick label style={font=\scriptsize},
  ylabel near ticks, xlabel near ticks, 
  xlabel style={yshift=3pt},
  grid=both,
  grid style={color=lightgray!60, dash pattern=on 2pt off 2pt},
  legend entries={Legacy, VLM, DINOv2, multi-domain, GLDv2-trained},
  legend pos=north west,
  legend style={cells={anchor=west}, font=\tiny, fill opacity=1.0, row sep=-1pt, inner sep=2pt},
  legend image post style={scale=0.8},
  ]

  \input{data/instre_gld_sop_adapt}

  \addplot[only marks, mark options={draw=black}, color=lightgray, solid, mark=*, opacity=0.7, mark size=2.5] table[y=ILIAS, x=GLDv2] \instregldsop;

  \addplot[only marks, mark options={draw=black}, color=applepink, solid, mark=*, mark size=2.5] table[y=ILIAS, x expr={\thisrow{GLDv2} / (\thisrow{show} == 1)}] \instregldsop;

  \addplot[only marks, mark options={draw=black}, color=appleteal, solid, mark=*, mark size=2.5] table[y=ILIAS, x expr={\thisrow{GLDv2} / (\thisrow{show} == 2)}] \instregldsop;   

  \addplot[only marks, mark options={draw=black}, color=applemustard, solid, mark=*, mark size=2.5] table[y=ILIAS, x expr={\thisrow{GLDv2} / (\thisrow{show} == 5)}] \instregldsop;

  \addplot[only marks, mark options={draw=black}, color=applepurple, solid, mark=*, mark size=2.5] table[y=ILIAS, x expr={\thisrow{GLDv2} / (\thisrow{show} == 3)}] \instregldsop;
  
  \end{axis}
\end{tikzpicture} 
    \hspace{10pt}
    \begin{tikzpicture}
\begin{axis}[
  width=.32\linewidth,
  height=.32\linewidth,
  xlabel={\small SOP},
  tick label style={font=\scriptsize},
  ylabel near ticks, xlabel near ticks, 
  xlabel style={yshift=3pt},
  grid=both,
  grid style={color=lightgray!60, dash pattern=on 2pt off 2pt},
  legend entries={Legacy, VLM, DINOv2, multi-domain, SOP-trained},
  legend pos=north west,
  legend style={cells={anchor=west}, font=\tiny, fill opacity=1.0, row sep=-1pt, inner sep=2pt},
  legend image post style={scale=0.8},
  ]

  \input{data/instre_gld_sop_adapt}

  \addplot[only marks, mark options={draw=black}, color=lightgray, solid, mark=*, opacity=0.7, mark size=2.5] table[y=ILIAS, x=SOP] \instregldsop;

  \addplot[only marks, mark options={draw=black}, color=applepink, solid, mark=*, mark size=2.5] table[y=ILIAS, x expr={\thisrow{SOP} / (\thisrow{show} == 1)}]
  \instregldsop;

  \addplot[only marks, mark options={draw=black}, color=appleteal, solid, mark=*, mark size=2.5] table[y=ILIAS, x expr={\thisrow{SOP} / (\thisrow{show} == 2)}]
  \instregldsop;

  \addplot[only marks, mark options={draw=black}, color=applemustard, solid, mark=*, mark size=2.5] table[y=ILIAS, x expr={\thisrow{SOP} / (\thisrow{show} == 5)}]
  \instregldsop;

  \addplot[only marks, mark options={draw=black}, color=appleblue, solid, mark=*, mark size=2.5] table[y=ILIAS, x expr={\thisrow{SOP} / (\thisrow{show} == 4)}]
  \instregldsop;

\end{axis}
\end{tikzpicture} 
  }
  \vspace{-10pt}
  \caption{\textbf{Comparison with other instance-level retrieval datasets} via reporting mAP@1k. Results with linear adaptation. INSTRE: 27.3K db size, multi-domain. GLDv2: 762K db size, single-domain. SOP:  60.5K db size, single-domain. Different network types are color-coded. For GLDv2 and SOP, models fine-tuned on these domains with the corresponding training sets are highlighted.
  \label{fig:instre_gld_sop_adapt}
  \vspace{-15pt}
  }
\end{figure*}

\noindent{\textbf{Impact of number of distractors}}.
Fig.~\ref{fig:number_of_distractors} presents the performance of five models under varying numbers of distractors. Performance declines as more distractors are added; however, significantly increasing the dataset's difficulty requires an exponential growth in the number of distractors. Notably, the ranking of models changes considerably when comparing performance with no distractors to that with 100M distractors. For example, DINOv2 demonstrates strong robustness to distractor increases, ranking last with no distractors but surpassing two models at 100M distractors and reaching others. Also, several crossings between models are observed. Therefore, evaluation at a large scale, provided by \ours, is important.

\noindent\textbf{Impact of image resolution.}
In Tab.~\ref{tab:resolution}, we investigate the impact of resolution and validate the rule of using as test resolution one up from the training one. Linear adaptation is not used in this experiment. It is clear that the vast majority of models achieve the best performance following the imposed rule; test at a resolution one level larger than the training resolution. Interestingly, SigLIP collapses when used with a resolution much lower than the training one. 

\noindent{\textbf{Impact of background clutter}}. To quantify the impact of background clutter, we experiment with masking out areas outside object bounding boxes in the positives during descriptor extraction. This approach improves SigLIP$^\dagger$ performance from 28.9 to 62.4.
Fig.~\ref{fig:heatmap_db_rank_vs_clutter} also presents the impact of clutter, \ie number of segments detected by SAM in a positive image outside of the object bounding box, on the ranking of this positive.
This experiment provides insight about the type of positives, according to clutter, that populate the top and bottom ranks.
Positives with less clutter, \ie low number of segments, are the most common in the higher ranks; while, positives with more clutter, \ie high number of segments, are the most common in the lower ranks.

\noindent{\textbf{Impact of object scale}}. Flowing the same strategy as above, but object bounding boxes are cropped and rescaled instead of masking, performance further improves to 69.4. However, although this does not reflect solely the impact of scale changes due to potential partial views and drastic viewpoint changes, it still gives a good insight into the limitations of the current models regarding scale changes.
Fig.~\ref{fig:heatmap_db_rank_vs_scale} presents the impact of relative scale, \ie percentage of the bounding box area within the image area, on the ranking of positive. This experiment provides insight into the type of positives, according to relative image coverage, that populate the corresponding rank ranges. Positives where the object covers a large area are the most common in the higher ranks; while, positives with a small area coverage are the most common in lower.

\begin{figure}[t]
    \centering
    \scalebox{0.82}{
    \begin{tabular}{cc}
        \hspace{-10pt}
        \begin{subfigure}{0.28\textwidth}
            \centering
            \newcommand{\mycolorbar}[8]
{  
    \foreach \x [count=\c] in {#3}{ \xdef\numcolo{\c}}
    \pgfmathsetmacro{\pieceheight}{#1/(\numcolo-1)}
    \xdef\lowcolo{}
    \foreach \x [count=\c] in {#3}
    {   \ifthenelse{\c = 1}
        {}
        {   
            \fill[bottom color=\lowcolo,top color=\x] 
            (#7,{#8+(\c-2)*\pieceheight}) rectangle (#7+#2,{#8+(\c-1)*\pieceheight});
        }
        \xdef\lowcolo{\x}
    }
    \draw (#7,#8) rectangle (#7+#2,#8+#1);
    \pgfmathsetmacro{\secondlabel}{#4+#6}
    \pgfmathsetmacro{\lastlabel}{#5+0.01}
    \pgfkeys{/pgf/number format/.cd,fixed,precision=2}
    \foreach \x in {#4,\secondlabel,...,\lastlabel}{
        \draw (#7+#2,{(\x-#4)/(#5-#4)*#1+#8}) -- ++ (0.05,0.) node[right, font=\tiny] {
            \pgfmathprintnumber{\x}
        };
    }
}

\newcommand{\viridisColor}[1]{%
    \pgfmathsetmacro{\temp}{#1}
    \pgfmathsetmacro{\rgbcol}{colormap={viridis}{\temp}}%
    \definecolor{cellColor}{rgb}{\rgbcol}%
}
\begin{tikzpicture}[scale=0.6]
    \pgfplotstableread[col sep=comma]{data/heatmap_db_rank_vs_clutter.csv}\datatable
    
    \pgfplotstablegetrowsof{\datatable}
    \pgfmathsetmacro{\numrows}{\pgfplotsretval}
    \pgfplotstablegetcolsof{\datatable}
    \pgfmathsetmacro{\numcols}{\pgfplotsretval}
    
    \pgfmathsetmacro{\numrowsi}{\numrows-1}
    \pgfmathsetmacro{\numcolsi}{\numcols-1}
    
    \pgfmathsetmacro{\cellspacing}{1.75}
    \pgfmathsetmacro{\cellsize}{10mm}
    \pgfmathsetmacro{\maxValue}{39.1}
    \pgfmathsetmacro{\minValue}{20}

    \pgfmathsetmacro{\height}{\cellspacing*\numrows-0.1}

    \foreach \y in {0,...,\numrowsi} {
        \foreach \x in {0,...,\numcolsi} {
          \pgfplotstablegetelem{\y}{\x}\of{\datatable}
          \let\cellvalue\pgfplotsretval
    
          \pgfmathsetmacro{\cellColorValue}{((\cellvalue-\minValue)/(\maxValue-\minValue)*100}
          \colorlet{cellColor}{highe!\cellColorValue!white}
    
          \node[fill=cellColor, draw=black, minimum size=\cellsize, font=\scriptsize] at (\x*\cellspacing+1, -\y*\cellspacing-1) {\cellvalue};
        };
    }

    \node[font=\small] at (3.7, 0.8) {number of segments};
    \node[font=\small, rotate=90] at (-0.8, -3.7) {positive rank};
    
    \foreach \x [count=\i from 1] in {{(0-18]}, {(18-33]}, {(33-53]}, {(53-558]}} {             \node[minimum size=\cellsize, font=\scriptsize, scale=0.9] at (\i*\cellspacing-0.7, 0.1) {\x};
    }
    
    \node[minimum size=\cellsize, font=\scriptsize, anchor= east, align=right,rotate=90, scale=0.9] at (-0.15, -0.27) {{[0-10]}};
    \node[minimum size=\cellsize, font=\scriptsize, anchor= east, align=right,rotate=90, scale=0.9] at (-0.15, -1.9) {{(10-100]}};
    \node[minimum size=\cellsize, font=\scriptsize, anchor= east, align=right,rotate=90, scale=0.9] at (-0.15, -3.47) {{(100-1000]}};
    \node[minimum size=\cellsize, font=\scriptsize, anchor= east, align=right,rotate=90, scale=0.9] at (-0.15, -5.45) {{(1000+]}};
\end{tikzpicture}
            \vspace{-12pt}
            \caption{\label{fig:heatmap_db_rank_vs_clutter}}
        \end{subfigure}
    &
        \hspace{-5pt}
        \begin{subfigure}{0.28\textwidth}
            \centering
            \providecommand{\mycolorbar}[8]
{  
    \foreach \x [count=\c] in {#3}{ \xdef\numcolo{\c}}
    \pgfmathsetmacro{\pieceheight}{#1/(\numcolo-1)}
    \xdef\lowcolo{}
    \foreach \x [count=\c] in {#3}
    {   \ifthenelse{\c = 1}
        {}
        {   
            \fill[bottom color=\lowcolo,top color=\x] 
            (#7,{#8+(\c-2)*\pieceheight}) rectangle (#7+#2,{#8+(\c-1)*\pieceheight});
        }
        \xdef\lowcolo{\x}
    }
    \draw (#7,#8) rectangle (#7+#2,#8+#1);
    \pgfmathsetmacro{\secondlabel}{#4+#6}
    \pgfmathsetmacro{\lastlabel}{#5+0.01}
    \pgfkeys{/pgf/number format/.cd,fixed,precision=2}
    \foreach \x in {#4,\secondlabel,...,\lastlabel}{
        \draw (#7+#2,{(\x-#4)/(#5-#4)*#1+#8}) -- ++ (0.05,0.) node[right, font=\tiny] {
            \pgfmathprintnumber{\x}
        };
    }
}

\begin{tikzpicture}[scale=0.6]
    \pgfplotstableread[col sep=comma]{data/heatmap_db_rank_vs_positive_scale.csv}\datatable
    
    \pgfplotstablegetrowsof{\datatable}
    \pgfmathsetmacro{\numrows}{\pgfplotsretval}
    \pgfplotstablegetcolsof{\datatable}
    \pgfmathsetmacro{\numcols}{\pgfplotsretval}
    
    \pgfmathsetmacro{\numrowsi}{\numrows-1}
    \pgfmathsetmacro{\numcolsi}{\numcols-1}
    
    \pgfmathsetmacro{\cellspacing}{1.75}
    \pgfmathsetmacro{\cellsize}{10mm}
    \pgfmathsetmacro{\maxValue}{43}
    \pgfmathsetmacro{\minValue}{20.}

    \pgfmathsetmacro{\height}{\cellspacing*\numrows-0.1}

    \foreach \y in {0,...,\numrowsi} {
        \foreach \x in {0,...,\numcolsi} {
          \pgfplotstablegetelem{\y}{\x}\of{\datatable}
          \let\cellvalue\pgfplotsretval
    
          \pgfmathsetmacro{\cellColorValue}{((\cellvalue-\minValue)/(\maxValue-\minValue)*100}
          \colorlet{cellColor}{highe!\cellColorValue!white}
    
          \node[fill=cellColor, draw=black, minimum size=\cellsize, font=\scriptsize] at (\x*\cellspacing+1, -\y*\cellspacing-1) {\cellvalue};
        };
    }
    
    \node[font=\small] at (3.7, 0.8) {area coverage};
    \node[font=\small, rotate=90] at (-0.8, -3.7) {positive rank};
    
    \foreach \x [count=\i from 1] in {{(0.3-7]}, {(7-14]}, {(14-27]},  {(27-100]}} {             \node[minimum size=\cellsize, font=\scriptsize, scale=0.9] at (\i*\cellspacing-0.8, 0.1) {\x};
    }
    
    \node[minimum size=\cellsize, font=\scriptsize, anchor= east, align=right,rotate=90, scale=0.9] at (-0.15, -0.27) {{[0-10]}};
    \node[minimum size=\cellsize, font=\scriptsize, anchor= east, align=right,rotate=90, scale=0.9] at (-0.15, -1.9) {{(10-100]}};
    \node[minimum size=\cellsize, font=\scriptsize, anchor= east, align=right,rotate=90, scale=0.9] at (-0.15, -3.47) {{(100-1000]}};
    \node[minimum size=\cellsize, font=\scriptsize, anchor= east, align=right,rotate=90, scale=0.9] at (-0.15, -5.45) {{(1000+]}};
\end{tikzpicture} 
            \vspace{-12pt}
            \caption{\label{fig:heatmap_db_rank_vs_scale}}
        \end{subfigure}
    \end{tabular}
    }
    \vspace{-10pt}
    \caption{\textbf{Impact of clutter and area coverage.} Percentage of images per ranking range based on SigLIP$^\dagger$ and grouped based on (a) clutter, \ie number of segments detected by SAM, (b) scale, \ie area of object bounding box in images. Column bins contain the same number of positives. Normalization per row is applied.
    \label{fig:heatmaps_scale}
    \vspace{-6pt}
    }
\end{figure}

\begin{figure*}[t]
    \centering
    \scalebox{1.}{
       \begin{tikzpicture}
\begin{axis}[
    width=1\linewidth,
    height=0.3\linewidth,
    bar width=6pt, 
    ylabel={mAP@1k},
    ymin=0, ymax=50,
    ytick={10,20,30,40,50,60}, 
    ytick distance=5, 
    ybar=2,
    enlarge y limits={upper, value=0.05},
    xtick=data,
    symbolic x coords={  
        product,
        technology,
        toy,
        fashion,
        landmark,
        household,
        media,
        art,
    },
    ylabel style={yshift=-2pt},
    xtick distance = 0.4,
    xticklabel style={
        font=\footnotesize},
    enlarge x limits=0.05, 
    nodes near coords,
    every node near coord/.append style={
        scale=0.56, 
        color=black,        
        font=\scriptsize, 
        /pgf/number format/.cd, 
        fixed zerofill, 
        precision=1,
    },
    legend style={
        at={(0.4,0.97)}, 
        anchor=north west, 
        legend columns=-1,  
        /tikz/every even column/.append style={column sep=0.2cm}, 
        font=\footnotesize,
        inner sep=1pt,  
        outer sep=1pt 
    },
    legend cell align={left}, 
    ymajorgrids=true,
    grid style={color=lightgray!60, dash pattern=on 2pt off 2pt},
]

\pgfplotstableread[col sep=comma]{data/performance_per_domain.csv}\loadeddata

\addplot+[ybar, color=black, fill=appleblue] table[x=domain,y=avg_ap_ames_siglip_384] {\loadeddata};
\addplot+[ybar, color=black, fill=appleteal] table[x=domain,y=avg_ap_siglip_384_proj] {\loadeddata};
\addplot+[ybar, color=black, fill=applemustard] table[x=domain,y=avg_ap_siglip_384_nw] {\loadeddata};
\addplot+[ybar, color=black, fill=applepurple] table[x=domain,y=avg_ap_siglip_384_text] {\loadeddata};
\addplot+[ybar, color=black, fill=appleorange] table[x=domain,y=avg_ap_dinov2_vitl14] {\loadeddata};

\legend{ SigLIP$^\dagger$+AMES,  SigLIP$^\dagger$, SigLIP, SigLIP-text, DINOv2$^\dagger$}
\end{axis}
\end{tikzpicture}
    }
    \vspace{-11pt}
    \caption{\textbf{Performance comparison per primary category.} mAP@1k  averaged over objects in the same primary-level category size, sorted by SigLIP$^\dagger$+AMES performance. Comparison between SigLIP with and without adaptation, SigLIP combined with AMES reranking, SigLIP t2i, and DINOv2. $\dagger$ indicates results with the linear adaptation.
    \label{fig:performance_barplot}
    \vspace{-10pt}
    }
\end{figure*}
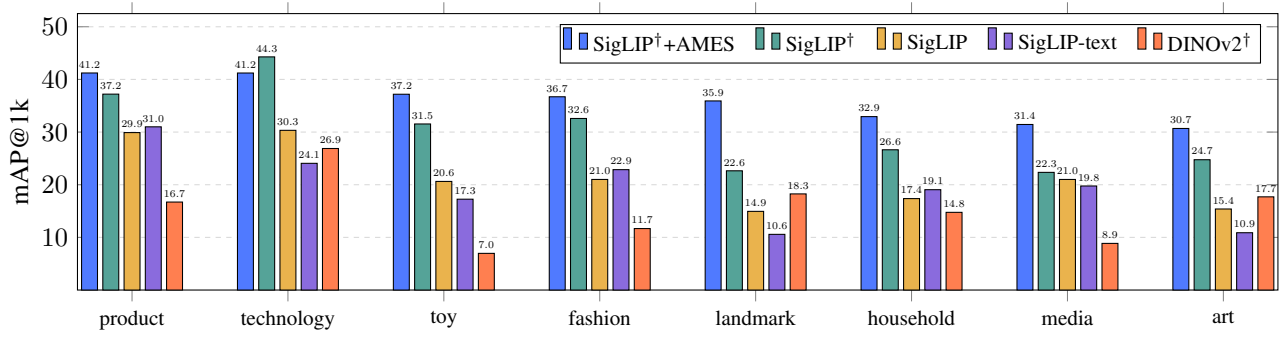

\noindent{\textbf{Multi-scale and multi-rotation extraction}}. 
A common approach to address scale variation is multi-scale feature extraction, as widely adopted in the literature~\cite{rtc19,sck+23}.
Applying multi-scale extraction asymmetrically, \ie only on the queries, yields an average 0.4 performance improvement across benchmarked models. SigLIP is marginally improved by 0.1. Multi-rotation is also tested in a similar manner, which, however, leads to an average drop of 0.3. Yet, SigLIP is marginally improved by 0.2.

\noindent{\textbf{Comparison with other datasets using linear adaptation}}. Fig.~\ref{fig:instre_gld_sop_adapt} presents the performance of global representation models with linear adaptation. Similar conclusions derive as in the case without adaptation. Only SigLIP achieves a competitive performance in SOP datasets out of the models not trained in-domain.

\noindent{\textbf{Performance per domain}}. Fig.~\ref{fig:performance_barplot} shows the average performance of objects grouped based on coarser taxonomy level.

\noindent{\textbf{Qualitative examples}}. Fig.~\ref{fig:fig_dataset_supplementary_i2i} and \ref{fig:fig_dataset_supplementary_t2i} show examples of retrieved images based on i2i and t2i retrieval, respectively.

\subsection{Linear adaptation}

\noindent\textbf{Comparison with other approaches.}
Tab.~\ref{tab:adaptation} compares the proposed linear adaptation with other linear projection methods trained on UnED for three models. 
All methods project the off-the-shelf descriptors to 512D ones.
The unsupervised PCA whitening (PCA$_w$)~\cite{jc12} and the supervised learnable whitening (L$_w$)~\cite{rtc19} approaches are evaluated. 
The proposed linear adaptation scheme achieves the best performance, typically with a large margin. It is the only one that does not drop off-the-shelf DINOv2 performance.

\begin{table}[t]
  \centering
  \small
\begin{tabular}{lcccc}
\toprule
\textbf{model} & \textbf{labels} & \textbf{DINOv2} & \textbf{OpenCLIP} & \textbf{SigLIP} \\ \midrule
\textbf{no adaptation}        & -         & 15.3 & 9.6  & 19.6 \\
\midrule
\textbf{PCA$_w$}~\cite{jc12}  & \ding{55} & 14.8 & 12.5 & 22.2 \\
\textbf{L$_w$}~\cite{rtc19}   & \ding{51} & 14.0 & 9.1  & 15.1 \\
\midrule
\textbf{ours}                 & \ding{51} & \textbf{15.3} & \textbf{18.3} & \textbf{28.9}\\
\bottomrule
\end{tabular}
  \vspace{-7pt}
  \caption{\textbf{Performance comparison for linear adaptation via mAP@1k.} Label requirement is indicated. Performance before adaptation is provided for reference.
  \vspace{-7pt}
  \label{tab:adaptation}}
\end{table}

\noindent\textbf{Impact of multi-domain linear adaptation.}
Tab.~\ref{tab:uned_subdomains} illustrates the performance of several models with linear adaptation trained on the four largest single-domain datasets of UnED, as well as the entire UnED. Training on a single domain typically increases the performance of VLMs, except in the case of Met, where performance drops dramatically. DINOv2 performance decreases consistently with single-domain training. Nevertheless, the margin with multi-domain training is significant, indicating that multi-domain training on the whole UnED is best suited for \ours.

\begin{table}[t]
  \centering
  \scalebox{0.87}{
    \small
\begin{tabular}{llccc}
\toprule
\textbf{dataset} & \textbf{domain} & \textbf{DINOv2} & \textbf{OpenCLIP} &  \textbf{SigLIP}  \\ 
\midrule
\textbf{no adaptation}               & -             & 15.3 & 9.6  & 19.6 \\ 
\midrule
\textbf{GLDv2}~\cite{wac+20}         & landmarks     & 14.6 & 14.2 & 25.6 \\
\textbf{Food2k}~\cite{mwl+23}        & food          & 12.6 & 13.6 & 22.6 \\ 
\textbf{Met}~\cite{ygg+21}           & artworks      & 14.7 & 5.1  & 7.6  \\ 
\textbf{iNaturalist}~\cite{vms+18}   & natural world & 14.2 & 16.3 & 26.4 \\ 
\midrule
\textbf{UnED}~\cite{ycc+23}          & multi-domain  & \textbf{15.3} & \textbf{18.3} & \textbf{28.9} \\
\bottomrule
\end{tabular}
  }
  \vspace{-5pt}
  \caption{\textbf{Performance comparison of single- and multi-domain linear adaptation.} mAP@1k of models with linear adaptation trained on different dataset setups based on UnED. Performance before adaptation is provided for reference.
  \vspace{-5pt}
  \label{tab:uned_subdomains}}
\end{table}

\begin{figure}[t]
    \centering
    \vspace{-12pt}
    \pgfplotsset{every tick label/.append style={font=\footnotesize}}
\begin{tikzpicture}
\begin{axis}[%
  width=\linewidth,
  height=0.64\linewidth,
  ylabel={mAP@1k},
  xlabel={descriptor dimensionality},
  grid=both,
  grid style={color=lightgray!60, dash pattern=on 2pt off 2pt},
  xlabel style={yshift=3pt},
  ylabel style={yshift=-4pt},
  log basis x=2,
  xmode=log,
  ytick={10,15,20,25,30},
  xtick={64,128,256,512,1024},
  xticklabels={64,128,256,512,1024},
  legend columns=3, 
  legend style={
    anchor=north east, 
    at={(0.92,1.16)}, 
    cells={anchor=west}, 
    font =\scriptsize, 
    fill opacity=1pt, 
    inner sep=1pt
    },
  ]

        \addplot[color=appleteal, mark=*, mark size=2.5, opacity=1.0, mark options={draw=black, line width=0.35pt}, line width=1.2pt] coordinates {
            (64, 9.93) (128, 13.02) (256, 14.67) (384, 14.95)
            (512, 15.26) (768, 15.56) (1024, 15.55)
        };
        \addlegendentry{DINOv2}

        \addplot[color=appleorange, mark=*, mark size=2.5, opacity=1.0, mark options={draw=black, line width=0.35pt}, line width=1.2pt] coordinates {
            (64, 10.11) (128, 13.99) (256, 15.84) (384, 15.77)
            (512, 16.07) (768, 15.90) (1024, 15.92)
        };        \addlegendentry{EVA-CLIP}

       \addplot[color=applegreen, mark=*, mark size=2.5, opacity=1.0, mark options={draw=black, line width=0.35pt}, line width=1.2pt] coordinates {
            (64, 9.46) (128, 14.90) (256, 16.30) (384, 16.91)
            (512, 17.10) (768, 17.00) (1024, 17.10)
        };
        \addlegendentry{MetaCLIP}

        \addplot[color=appleblue, mark=*, mark size=2.5, opacity=1.0, mark options={draw=black, line width=0.35pt}, line width=1.2pt] coordinates {
            (64, 10.65) (128, 15.95) (256, 18.08) (384, 18.08)
            (512, 18.17) (768, 18.59) (1024, 18.29)
        };
        \addlegendentry{OpenCLIP}
        
        \addplot[color=applepurple, mark=*, mark size=2.5, opacity=1.0, mark options={draw=black, line width=0.35pt}, line width=1.2pt] coordinates {
            (64, 17.98) (128, 25.69) (256, 28.46) (384, 28.58)
            (512, 29.17) (768, 28.84) (1024, 29.11)
        };
        \addlegendentry{SigLIP}
               
\end{axis}
\end{tikzpicture}
    \vspace{-11pt}
    \caption{\textbf{Impact of descriptor dimensionality.} mAP@1k of five models with the linear adaptation of various dimensionalities.
    \label{fig:descriptor_dimensionality}
    \vspace{-15pt}
    }
\end{figure}
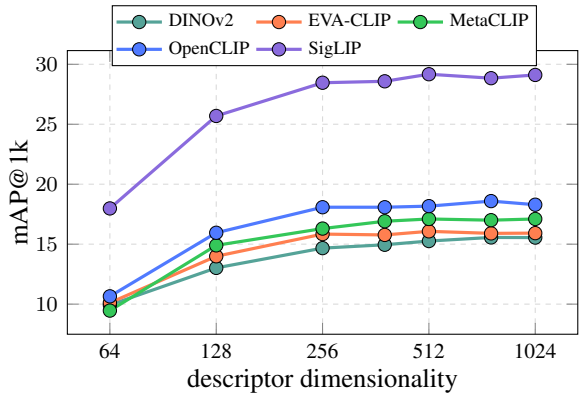

\noindent\textbf{Impact of descriptor dimensionality.}
Fig.~\ref{fig:descriptor_dimensionality} illustrates the performance of five models linearly adapted on UnED with varying descriptor dimensionalities. For all models, performance saturates at a descriptor dimensionality of 256D, with only marginal improvements observed for most models beyond this point.

\noindent\textbf{Robustness.} 
We conduct three independent runs using different random seeds to evaluate the robustness of the linear adaptation. 
Across five global descriptors, the proposed scheme exhibits strong robustness, with a maximum standard deviation of 0.2 and a minimum of 0 across runs.

\subsection{Re-ranking with local representations}

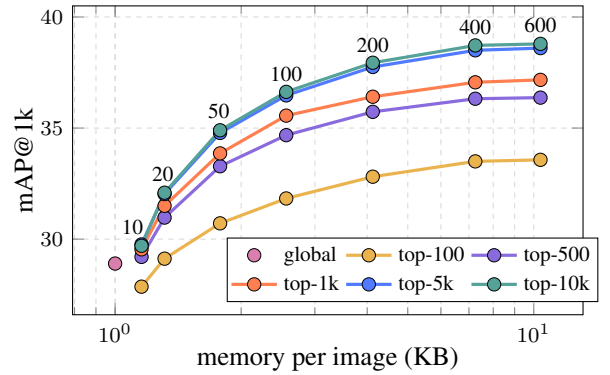
\begin{figure}[t]
    \centering
    \pgfplotsset{every tick label/.append style={font=\footnotesize}}
\begin{tikzpicture}
\begin{axis}[%
  width=\linewidth,
  height=0.68\linewidth,
  ylabel={mAP@1k},
  xlabel={memory per image (KB)},
  grid=both,
  grid style={color=lightgray!60, dash pattern=on 2pt off 2pt},
  xlabel style={yshift=5pt},
  ylabel style={yshift=-4pt},
  xmode=log,
  ymax=40.5,
  legend columns=3, 
  legend style={
    anchor=north east, 
    at={(1.03,0.25)}, 
    cells={anchor=west}, 
    font =\footnotesize, 
    fill opacity=1pt, 
    inner sep=1pt
    },
  ]
        \addplot[only marks, color=applepink, mark=*, mark size=2.5, opacity=1.0, mark options={draw=black, line width=0.35pt}, line width=1.2pt] coordinates {(1.0, 28.9)};

        \addplot[color=applemustard, mark=*, mark size=2.5, opacity=1.0, mark options={draw=black, line width=0.35pt}, line width=1.2pt] coordinates {
                                (1.15625,       27.86)          
                                (1.3125,        29.12)          
                                (1.78125,       30.71)          
                                (2.5625,        31.83)          
                                (4.125,         32.81)          
                                (7.25,          33.50)          
                                (10.375,        33.57)          
        };

       \addplot[color=applepurple, mark=*, mark size=2.5, opacity=1.0, mark options={draw=black, line width=0.35pt}, line width=1.2pt] coordinates {
                                (1.15625,       29.20)          
                                (1.3125,        30.97)          
                                (1.78125,       33.28)          
                                (2.5625,        34.68)          
                                (4.125,         35.73)          
                                (7.25,          36.32)          
                                (10.375,        36.37)          
        };

        \addplot[color=appleorange, mark=*, mark size=2.5, opacity=1.0, mark options={draw=black, line width=0.35pt}, line width=1.2pt] coordinates {
                                (1.15625,       29.54)          
                                (1.3125,        31.50)          
                                (1.78125,       33.86)          
                                (2.5625,        35.56)          
                                (4.125,         36.41)          
                                (7.25,          37.06)          
                                (10.375,        37.17)          
        };

        \addplot[color=appleblue, mark=*, mark size=2.5, opacity=1.0, mark options={draw=black, line width=0.35pt}, line width=1.2pt] coordinates {
                                (1.15625,       29.76)          
                                (1.3125,        32.03)          
                                (1.78125,       34.78)          
                                (2.5625,        36.47)          
                                (4.125,         37.75)          
                                (7.25,          38.50)          
                                (10.375,        38.60)          
        };

        \addplot[color=appleteal, mark=*, mark size=2.5, opacity=1.0, mark options={draw=black, line width=0.35pt}, line width=1.2pt] coordinates {
                                (1.15625,       29.70)          
                                (1.3125,        32.09)          
                                (1.78125,       34.90)          
                                (2.5625,        36.63)          
                                (4.125,         37.94)          
                                (7.25,          38.72)          
                                (10.375,        38.79)          
        };
        
        \addlegendentry{global}
        \addlegendentry{top-100}
        \addlegendentry{top-500}
        \addlegendentry{top-1k}
        \addlegendentry{top-5k}
        \addlegendentry{top-10k}

        \node [above] at (axis cs:  1.1,29.80) {\footnotesize \textcolor{black}{10}};
        \node [above] at (axis cs:  1.3,32.2) {\footnotesize \textcolor{black}{20}};
        \node [above] at (axis cs:  1.77,35.0) {\footnotesize \textcolor{black}{50}};
        \node [above] at (axis cs:  2.56,36.7) {\footnotesize \textcolor{black}{100}};
        \node [above] at (axis cs:  4.125,38.0) {\footnotesize \textcolor{black}{200}};
        \node [above] at (axis cs:  7.25,38.8) {\footnotesize \textcolor{black}{400}};
        \node [above] at (axis cs:  10.375,38.9) {\footnotesize \textcolor{black}{600}};

    \end{axis}
\end{tikzpicture}
    \vspace{-7pt}
    \caption{\textbf{Impact of the re-ranking shortlist size and required memory for local descriptors.} Text above each point denotes the number of local descriptors per DB image. The shortlist size is indicated in the legend. Results are with the linear adaptation.
    \label{fig:ames_num_desc}
    \vspace{-7pt}
    }
\end{figure}

\noindent\textbf{Impact of top-M re-ranked images and number of local descriptors.}
Fig.~\ref{fig:ames_num_desc} illustrates the performance of SigLIP with re-ranking when an increasing number of re-ranked images and local descriptors, translated to memory per image, are used. Performance increases as both variables increase. In the default scenario of top-1k and 100 descriptors, the performance is 35.6, which requires 0.6sec per query and approximately 150GB of memory. In an unconstrained scenario, the top performance is 38.8, requiring 20sec and almost 900GB.

\noindent\textbf{Combination with various global representations.}
Tab.~\ref{tab:ames_models} presents the performance with and without re-ranking on \ours and \miniours using various models for global representation. mAP@1k is improved by more than 6 when re-ranking is applied for all models and datasets.

\begin{table}[t]

  \centering
  \small
\begin{tabular}{lcccc}
\toprule
\multirow{2}{*}{\textbf{model}} & \multicolumn{2}{c}{\textbf{\ours}} & \multicolumn{2}{c}{\textbf{\miniours}} \\ \cmidrule{2-3} \cmidrule{4-5}
 & \textbf{mAP@1k} & \textbf{oracle} & \textbf{mAP@1k} & \textbf{oracle} \\ 
\midrule
\textbf{DINOv2}$^\dagger$   & 15.3 & 34.0 & 18.8 & 41.8 \\
\quad\textbf{+ AMES}        & 21.8 & 34.0 & 26.5 & 41.8 \\
\midrule
\textbf{OpenCLIP}$^\dagger$ & 18.3 & 48.0 & 22.9 & 56.3 \\
\quad\textbf{+ AMES}        & 27.1 & 48.0 & 32.9 & 56.3 \\
\midrule
\textbf{SigLIP}$^\dagger$   & 28.9 & 56.0 & 34.3 & 63.9 \\ 
\quad\textbf{+ AMES}        & 35.6 & 56.0 & 41.4 & 63.9 \\
\bottomrule
\end{tabular}
  \vspace{-7pt}
  \caption{\textbf{Re-ranking on top of different global representations.} mAP@1k and oracle re-ranking on \ours and \miniours. + indicates re-ranking with AMES. $\dagger$ indicates results with the linear adaptation.
  \label{tab:ames_models}
  \vspace{-7pt}
  }
\end{table}

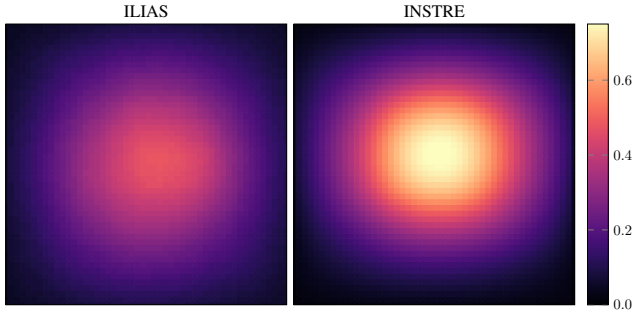
\begin{figure}[t]
  \centering
  \scalebox{0.55}{
    \begin{tikzpicture}
    \begin{axis}[
        name=heatmap1,
        width=1.0\linewidth,
        height=1.0\linewidth,
        view={0}{90}, 
        ymin=0, ymax=49,
        ticks=none,
        colormap/magma,
        enlargelimits=false,
        axis on top,
        point meta min=0,
        point meta max=0.80,
        mesh/cols=50,
        mesh/rows=50,
        title={\ours},
        title style={font=\large, yshift=-5pt}
    ]
    \addplot [
        matrix plot*,
        point meta=explicit,
      ] file {data/heatmap_spatial_location.dat};
    \end{axis}
    \hspace{5pt}
    \begin{axis}[
        at={(heatmap1.south east), anchor=north west},
        width=1.0\linewidth,
        height=1.0\linewidth,
        view={0}{90}, 
        ymin=0, ymax=49,
        ticks=none,
        colormap/magma,
        colorbar,
        colorbar style={
            yticklabel style={
                /pgf/number format/.cd,
                fixed,
                precision=1,
                fixed zerofill,
            },
        },
        enlargelimits=false,
        axis on top,
        point meta min=0,
        point meta max=0.75,
        mesh/cols=50,
        mesh/rows=50,
        title={INSTRE},
        title style={font=\large, yshift=-5pt}
    ]
    \addplot [
        matrix plot*,
        point meta=explicit,
      ] file {data/heatmap_spatial_location_instre.dat};
    \end{axis}
\end{tikzpicture}
  }
  \caption{\textbf{Distribution of object bounding boxes in positives.}
  \label{fig:heatmap_spatial_location}
  \vspace{-10pt}
  }
\end{figure}

\begin{figure}[t]
  \centering
  \vspace{10pt}
  \input{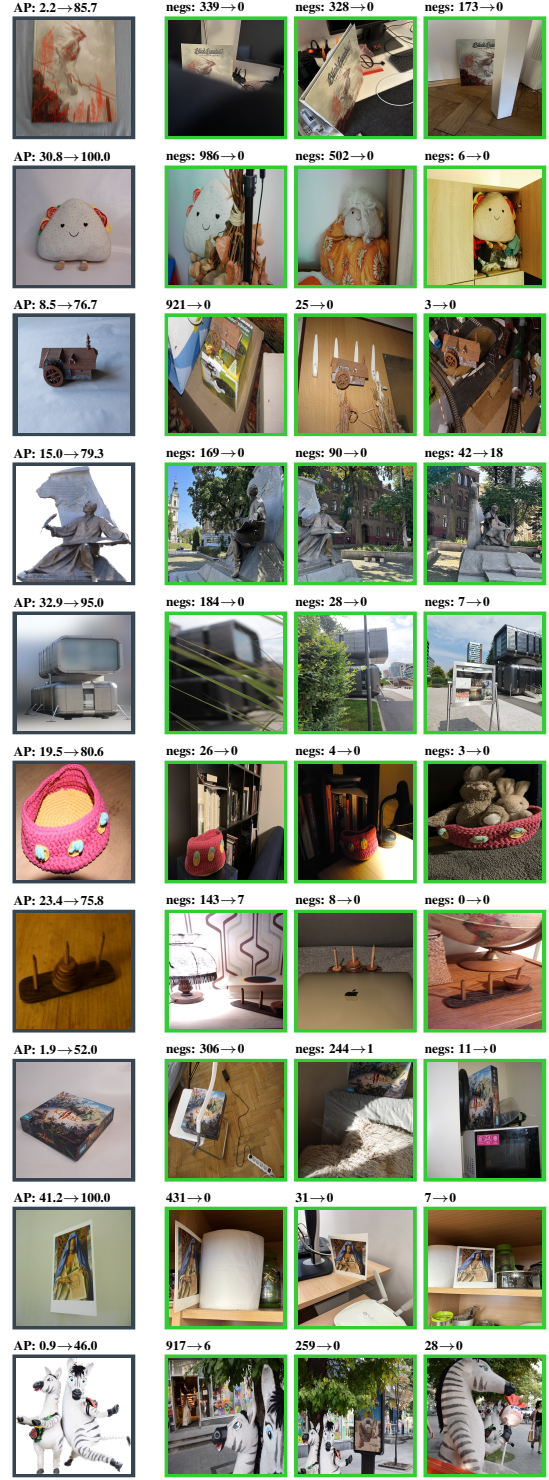}
  \vspace{-15pt}
  \caption{\textbf{Re-ranking with AMES.} Queries with the most significant AP increase from re-ranking. The number of negatives ranked above positives is reported on top, as before $\rightarrow$ after re-ranking.
  \label{fig:ames_rerank}
  }
\end{figure}

\noindent\textbf{Qualitative examples.} Fig.~\ref{fig:ames_rerank} presents some queries with the largest AP improvement from re-ranking with AMES. Several cases of severe clutter, scale changes, and partial views are successfully retrieved with re-ranking.

\section{Dataset extras}

\noindent{\textbf{Spatial location of objects in positives}}. 
Fig.~\ref{fig:heatmap_spatial_location} illustrates the spatial location of the object in the positives. Center bias in \ours is much less prominent in comparison with INSTRE~\cite{wj15} dataset. 

\noindent{\textbf{Taxonomy}}. Fig.~\ref{fig:taxonomy} illustrate the defined categories for the three taxonomy levels.

\begin{figure*}[t]
  \centering
  \input{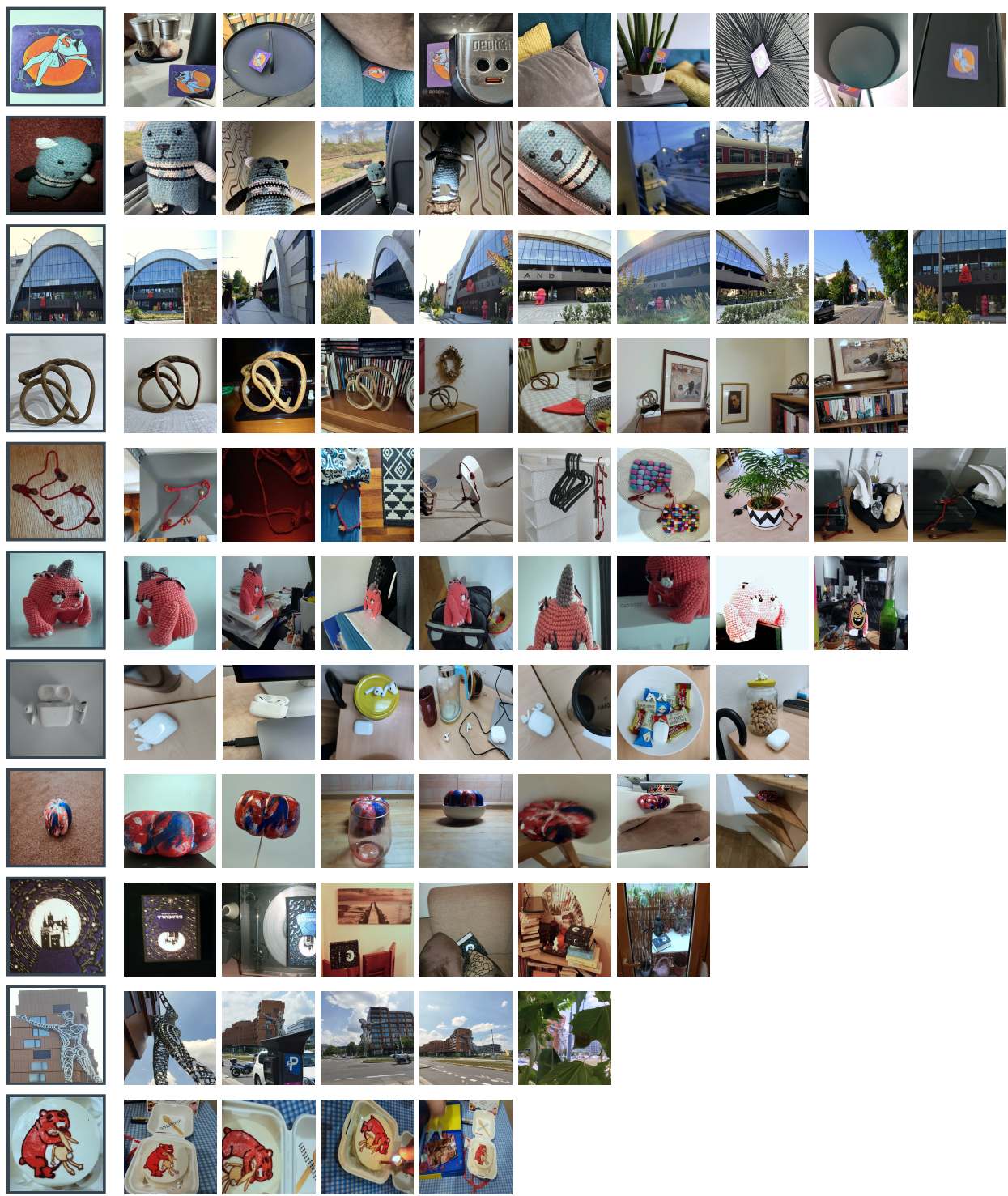}
  \caption{\textbf{Examples of collected query objects.} Queries and multiple positives are displayed. 
  {\color{lightblue}\textbf{Gray:}} queries.
  \label{fig:fig_dataset_supplementary_obj}
  }
\end{figure*}

\noindent{\textbf{Query and positive examples}}. Fig.~\ref{fig:fig_dataset_supplementary_obj} provides visual examples of the collected queries and positives of several query objects.

\noindent{\textbf{Benchmarked models}}. Tab.~\ref{tab:supp_all_models} and \ref{tab:supp_all_text_models} provide details and performance on \ours and \miniours of all models.

\section{Dataset hosting, sharing and license}
\ours is hosted in our servers in its entirety (\ie collected images and the downloaded YFCC100M) to assert its long-term availability to the broader public. All collected images are shared under the permissive CC-BY 4.0 license. The downloaded images are distributed under their original license. All collectors have signed a consent form for the distribution of their images under this license.

\newpage

\begin{table*}[t]
  \centering
  \scalebox{0.67}{
    \newcolumntype{R}{>{\raggedleft\arraybackslash}p{2.2em}}
\setlength{\tabcolsep}{4pt}
\begin{tabular}{llllrrrrrrrRRRR} \toprule
\textbf{checkpoint} & \textbf{year} & \textbf{cite} & \textbf{repo} & \textbf{arch} & \textbf{train} & \textbf{dims} & \textbf{dataset} & \textbf{data size} & \textbf{train res} & \textbf{test res} & \textbf{5M} & \textbf{100M} & \textbf{5M$^\dagger$} & \textbf{100M$^\dagger$} \\ \midrule
alexnet.tv\_in1k                                  & 2012 & \cite{ksh12}             & torchvision & CNN   & sup      & 256  & \texttt{in1k}      & 1M   & 224 & 384 & 2.0  & 1.5 & 1.9   & 1.3  \\
vgg16.tv\_in1k                                    & 2014 & \cite{sz14}              & torchvision & CNN   & sup      & 512  & \texttt{in1k}      & 1M   & 224 & 384 & 3.0  & 2.3 & 2.3   & 1.6  \\
resnet50.tv\_in1k                                 & 2015 & \cite{hzr+16}            & torchvision & R50   & sup      & 2048 & \texttt{in1k}      & 1M   & 224 & 384 & 2.3  & 1.7 & 2.5   & 1.8  \\
resnet101.tv\_in1k                                & 2015 & \cite{hzr+16}            & torchvision & R101  & sup      & 2048 & \texttt{in1k}      & 1M   & 224 & 384 & 2.7  & 1.9 & 2.7   & 1.8  \\
densenet169.tv\_in1k                              & 2016 & \cite{hlv+17}            & torchvision & CNN   & sup      & 2048 & \texttt{in1k}      & 1M   & 224 & 384 & 3.2  & 2.4 & 2.9   & 2.0  \\
inception\_v4.tf\_in1k                            & 2017 & \cite{slj+15}            & torchvision & CNN   & sup      & 1536 & \texttt{in1k}      & 1M   & 299 & 512 & 1.7  & 1.1 & 1.5   & 1.0  \\
nasnetalarge.tf\_in1k                             & 2018 & \cite{zvs+18}            & torchvision & CNN   & sup      & 4032 & \texttt{in1k}      & 1M   & 331 & 512 & 1.7  & 1.0 & 1.6   & 1.0  \\
tf\_efficientnet\_b4.ns\_jft\_in1k                & 2019 & \cite{tl19}              & timm        & CNN   & sup+dist & 1792 & \texttt{in1k}      & 1M   & 380 & 512 & 3.8  & 2.6 & 4.3   & 2.9  \\
vit\_base\_patch16\_224.augreg\_in1k              & 2020 & \cite{dbk+21,vit-augreg} & timm        & ViT-B & sup      & 768  & \texttt{in1k}      & 1M   & 224 & 384 & 1.4  & 1.0 & 1.9   & 1.3  \\
vit\_base\_patch16\_224.augreg\_in21k             & 2020 & \cite{dbk+21,vit-augreg} & timm        & ViT-B & sup      & 768  & \texttt{in21k}     & 14M  & 224 & 384 & 4.2  & 3.0 & 6.2   & 4.4  \\
vit\_large\_patch16\_224.augreg\_in21k            & 2020 & \cite{dbk+21,vit-augreg} & timm        & ViT-L & sup      & 1024 & \texttt{in21k}     & 14M  & 224 & 384 & 6.0  & 4.6 & 7.3   & 5.3  \\
vit\_large\_patch16\_224.augreg\_in21k\_ft\_in1k  & 2020 & \cite{dbk+21,vit-augreg} & timm        & ViT-L & sup      & 1024 & \texttt{in1k}      & 1M   & 224 & 384 & 5.1  & 3.6 & 6.6   & 4.7  \\
vit\_large\_patch16\_384.augreg\_in21k\_ft\_in1k  & 2020 & \cite{dbk+21,vit-augreg} & timm        & ViT-L & sup      & 1024 & \texttt{in1k}      & 1M   & 384 & 512 & 7.2  & 5.3 & 8.7   & 6.4  \\
deit3\_base\_patch16\_224.fb\_in1k                & 2021 & \cite{tcd+21}            & timm        & ViT-B & sup+dist & 768  & \texttt{in1k}      & 1M   & 224 & 384 & 1.9  & 1.2 & 2.7   & 1.8  \\
deit3\_large\_patch16\_224.fb\_in1k               & 2021 & \cite{tcd+21}            & timm        & ViT-L & sup+dist & 1024 & \texttt{in1k}      & 1M   & 224 & 384 & 2.0  & 1.5 & 3.3   & 2.4  \\
RN50.openai                                       & 2021 & \cite{clip}              & github      & R50   & vla      & 1024 & \texttt{opanai}    & 400M & 224 & 384 & 4.4  & 3.2 & 8.5   & 6.0  \\
vit\_base\_patch16\_clip\_224.openai              & 2021 & \cite{clip}              & timm        & ViT-B & vla      & 512  & \texttt{opanai}    & 400M & 224 & 384 & 5.9  & 4.2 & 10.7  & 7.9  \\
vit\_large\_patch14\_clip\_224.openai             & 2021 & \cite{clip}              & timm        & ViT-L & vla      & 768  & \texttt{opanai}    & 400M & 224 & 384 & 9.0  & 7.0 & 15.8  & 11.9 \\
vit\_large\_patch14\_clip\_336.openai             & 2021 & \cite{clip}              & timm        & ViT-L & vla      & 768  & \texttt{opanai}    & 400M & 336 & 512 & 12.1 & 9.4 & 19.9  & 15.2 \\
vit\_large\_patch14\_clip\_224.laion2b            & 2021 & \cite{cbw+23,iww+21}     & timm        & ViT-L & vla      & 768  & \texttt{laion2b}   & 2B   & 224 & 384 & 11.8 & 9.4 & 17.5  & 13.7 \\
swav\_resnet50                                    & 2021 & \cite{cmm+20}            & github      & R50   & ssl      & 2048 & \texttt{in1k}      & 1M   & 224 & 384 & 2.2  & 1.7 & 2.9   & 2.1  \\
dino\_resnet50                                    & 2021 & \cite{ctm+21}            & github      & R50   & ssl      & 2048 & \texttt{in1k}      & 1M   & 224 & 384 & 3.8  & 2.9 & 4.1   & 2.9  \\
dino\_vitb16                                      & 2021 & \cite{ctm+21}            & github      & ViT-B & ssl      & 768  & \texttt{in1k}      & 1M   & 224 & 384 & 5.0  & 3.7 & 6.6   & 4.8  \\
moco\_v3\_resnet50                                & 2021 & \cite{hfw+20}            & github      & R50   & ssl      & 2048 & \texttt{in1k}      & 1M   & 224 & 384 & 3.3  & 2.6 & 3.4   & 2.6  \\
moco\_v3\_vitb                                    & 2021 & \cite{hfw+20}            & github      & ViT-B & ssl      & 768  & \texttt{in1k}      & 1M   & 224 & 384 & 2.5  & 1.9 & 3.2   & 2.3  \\
convnext\_base.fb\_in1k                           & 2022 & \cite{convnext}          & timm        & CN-B  & sup      & 1024 & \texttt{in1k}      & 1M   & 288 & 384 & 2.8  & 2.0 & 3.9   & 2.7  \\
convnext\_base.fb\_in22k                          & 2022 & \cite{convnext}          & timm        & CN-B  & sup      & 1536 & \texttt{in22k}     & 14M  & 224 & 384 & 3.2  & 6.4 & 9.9   & 7.6  \\
convnext\_large.fb\_in1k                          & 2022 & \cite{convnext}          & timm        & CN-L  & sup      & 1024 & \texttt{in1k}      & 1M   & 288 & 384 & 8.9  & 2.2 & 4.2   & 2.9  \\
convnext\_large.fb\_in22k                         & 2022 & \cite{convnext}          & timm        & CN-L  & sup      & 1536 & \texttt{in22k}     & 14M  & 288 & 384 & 8.6  & 6.6 & 9.1   & 6.9  \\
convnext\_base.clip\_laion2b\_augreg              & 2022 & \cite{iww+21,convnext}   & timm        & CN-B  & vla      & 640  & \texttt{laion2b}   & 2B   & 256 & 384 & 10.7 & 7.9 & 18.1  & 14.0 \\
convnext\_large\_mlp.clip\_laion2b\_ft\_soup\_320 & 2022 & \cite{iww+21,convnext}   & timm        & CN-L  & vla      & 768  & \texttt{laion2b}   & 2B   & 320 & 512 & 12.7 & 9.6 & 22.9  & 18.3 \\
recall\_512-resnet50                              & 2022 & \cite{ptm+22}            & github$^*$  & R50   & sup      & 512  & \texttt{sop}       & 60k  & 224 & 384 & 2.3  & 1.6 & 3.1   & 2.1  \\
recall\_512-vit\_base\_patch16\_224\_in21k        & 2022 & \cite{ptm+22}            & github$^*$  & ViT-B & sup      & 512 & \texttt{sop}        & 60k  & 224 & 384 & 6.8  & 5.0 & 7.3   & 5.3  \\
cvnet\_resnet50                                   & 2022 & \cite{lsl+22}            & github      & R50   & sup      & 2048 & \texttt{gldv2}     & 1M   & 512 & 724 & 3.7  & 2.9 & 3.5   & 2.6  \\
cvnet\_resnet101                                  & 2022 & \cite{lsl+22}            & github      & R101  & sup      & 2048 & \texttt{gldv2}     & 1M   & 512 & 724 & 3.9  & 3.0 & 4.2   & 3.1  \\
superglobal\_resnet50                             & 2023 & \cite{sck+23}            & github      & R50   & sup      & 2048 & \texttt{gldv2}     & 1M   & 512 & 724 & 4.3  & 3.4 & 3.8   & 2.8  \\
superglobal\_resnet101                            & 2023 & \cite{sck+23}            & github      & R101  & sup      & 2048 & \texttt{gldv2}     & 1M   & 512 & 724 & 4.5  & 3.4 & 4.5   & 3.2  \\
hier\_dino\_vits16\_sop                           & 2023 & \cite{kjk23}             & github$^*$  & ViT-S & sup      & 384  & \texttt{sop}       & 60k  & 224 & 384 & 4.6  & 3.3 & 5.1   & 3.6  \\
eva02\_base\_patch14\_224.mim\_in22k              & 2023 & \cite{fwx+23}            & timm        & ViT-B & ssl      & 768  & \texttt{in22k}     & 14M  & 224 & 384 & 3.1  & 2.1 & 4.7   & 3.2  \\
eva02\_large\_patch14\_224.mim\_in22k             & 2023 & \cite{fwx+23}            & timm        & ViT-L & ssl      & 1024 & \texttt{in22k}     & 14M  & 224 & 384 & 2.5  & 1.5  & 3.9  & 2.7  \\
eva02\_large\_patch14\_224.mim\_m38m              & 2023 & \cite{fwx+23}            & timm        & ViT-L & ssl      & 1024 & \texttt{merged38m} & 38M  & 224 & 384 & 6.7  & 4.7  & 8.8  & 6.1  \\
eva02\_base\_patch16\_clip\_224.merged2b          & 2023 & \cite{fwx+23,evaclip}    & timm        & ViT-B & vla      & 512  & \texttt{merged2b}  & 2B   & 224 & 384 & 7.8  & 5.9  & 11.7 & 8.7  \\
eva02\_large\_patch14\_clip\_336.merged2b         & 2023 & \cite{fwx+23,evaclip}    & timm        & ViT-L & vla      & 768  & \texttt{merged2b}  & 2B   & 336 & 512 & 13.6 & 10.9 & 20.9 & 16.0 \\
unicom\_vit\_base\_patch16\_224                   & 2023 & \cite{ady+23}            & github      & ViT-B & dist     & 768  & \texttt{laion400m} & 400M & 224 & 384 & 13.8 & 11.0 & 13.8 & 11.1 \\
unicom\_vit\_large\_patch14\_224                  & 2023 & \cite{ady+23}            & github      & ViT-L & dist     & 768  & \texttt{laion400m} & 400M & 224 & 384 & 18.0 & 13.8 & 17.7 & 13.8 \\
unicom\_vit\_large\_patch14\_336                  & 2023 & \cite{ady+23}            & github      & ViT-L & dist     & 768  & \texttt{laion400m} & 400M & 336 & 512 & 17.8 & 13.9 & 18.6 & 14.6 \\
unicom\_vit\_base\_patch16\_gldv2                 & 2023 & \cite{ady+23}            & github$^*$  & ViT-B & sup      & 768  & \texttt{gldv2}     & 400M & 512 & 724 & 3.7  & 3.0  & 4.1  & 3.3  \\
unicom\_vit\_base\_patch16\_sop                   & 2023 & \cite{ady+23}            & github$^*$  & ViT-B & sup      & 768  & \texttt{sop}       & 400M & 224 & 384 & 12.2 & 9.1  & 12.8 & 9.9  \\
uscrr\_64-vit\_base\_patch16\_clip\_224.openai    & 2023 & \cite{ycc+23}            & github      & ViT-B & sup      & 768  & \texttt{uned}      & 2.8M & 224 & 384 & 5.7  & 3.8  & 6.4  & 4.3  \\
dinov2\_vitb14                                    & 2023 & \cite{odm+24}            & github      & ViT-B & ssl      & 768  & \texttt{lvd142m}   & 142M & 518 & 724 & 14.3 & 11.5 & 15.0 & 12.1 \\
dinov2\_vitl14                                    & 2023 & \cite{odm+24}            & github      & ViT-L & ssl      & 1024 & \texttt{lvd142m}   & 142M & 518 & 724 & 18.5 & 15.3 & 18.8 & 15.3 \\
vit\_base\_patch16\_siglip\_224.webli             & 2023 & \cite{siglip}            & timm        & ViT-B & vla      & 768  & \texttt{webli}     & 10B  & 224 & 384 & 14.1 & 11.2 & 19.4 & 15.7 \\
vit\_base\_patch16\_siglip\_256.webli             & 2023 & \cite{siglip}            & timm        & ViT-B & vla      & 768  & \texttt{webli}     & 10B  & 256 & 384 & 14.6 & 11.5 & 20.6 & 16.7 \\
vit\_base\_patch16\_siglip\_384.webli             & 2023 & \cite{siglip}            & timm        & ViT-B & vla      & 768  & \texttt{webli}     & 10B  & 384 & 512 & 19.3 & 15.6 & 26.2 & 21.5 \\
vit\_base\_patch16\_siglip\_512.webli             & 2023 & \cite{siglip}            & timm        & ViT-B & vla      & 768  & \texttt{webli}     & 10B  & 512 & 724 & 20.1 & 16.6 & 27.5 & 23.0 \\
vit\_large\_patch16\_siglip\_256.webli            & 2023 & \cite{siglip}            & timm        & ViT-L & vla      & 1024 & \texttt{webli}     & 10B  & 256 & 384 & 18.8 & 15.2 & 26.3 & 21.8 \\
vit\_large\_patch16\_siglip\_384.webli            & 2023 & \cite{siglip}            & timm        & ViT-L & vla      & 1024 & \texttt{webli}     & 10B  & 384 & 512 & 24.2 & 19.6 & 34.3 & 28.9 \\
vit\_base\_patch16\_clip\_224.metaclip\_2pt5b     & 2024 & \cite{metaclip}          & timm        & ViT-B & vla      & 768  & \texttt{2pt5b}     & 2.5B & 224 & 384 & 8.8  & 6.6  & 12.7 & 9.4  \\
vit\_large\_patch14\_clip\_224.metaclip\_2pt5b    & 2024 & \cite{metaclip}          & timm        & ViT-L & vla      & 1024 & \texttt{2pt5b}     & 2.5B & 224 & 384 & 14.4 & 11.7 & 21.7 & 16.9 \\
dinov2\_vitb14\_reg                               & 2024 & \cite{odm+24,doj+23}     & github      & ViT-B & ssl      & 768  & \texttt{lvd142m}   & 142M & 518 & 724 & 11.8 & 9.4  & 13.5 & 10.7 \\
dinov2\_vitl14\_reg                               & 2024 & \cite{odm+24,doj+23}     & github      & ViT-L & ssl      & 1024 & \texttt{lvd142m}   & 142M & 518 & 724 & 15.9 & 12.7 & 17.1 & 13.6 \\
unic\_l                                           & 2024 & \cite{swl+24}            & github      & ViT-L & dist     & 1024 & \texttt{in1k}      & 1M   & 518 & 512 & 11.4 & 8.9  & 15.3 & 11.7 \\
udon\_64-vitb\_in21k\_ft\_in1k                    & 2024 & \cite{yca+24}            & github$^*$  & ViT-B & sup      & 768  & \texttt{uned}      & 2.8M & 224 & 384 & 7.5  & 5.5  & 7.3  & 5.3  \\
udon\_64-vitb\_clip\_openai                       & 2024 & \cite{yca+24}            & github$^*$  & ViT-B & sup      & 768  & \texttt{uned}      & 2.8M & 224 & 384 & 8.3  & 5.9  & 9.2  & 6.7  \\
vit\_base\_patch16\_siglip\_384.v2\_webli         & 2025 & \cite{siglip2}           & timm        & ViT-B & vla      & 768  & \texttt{webli}     & 10B  & 384 & 512 & 18.4 & 15.0 & 27.5 & 22.6 \\
vit\_base\_patch16\_siglip\_512.v2\_webli         & 2025 & \cite{siglip2}           & timm        & ViT-B & vla      & 768  & \texttt{webli}     & 10B  & 512 & 724 & 18.6 & 15.4 & 28.6 & 23.5 \\
vit\_large\_patch16\_siglip\_384.v2\_webli        & 2025 & \cite{siglip2}           & timm        & ViT-L & vla      & 1024 & \texttt{webli}     & 10B  & 384 & 512 & 24.6 & 19.9 & 36.3 & 30.3 \\
vit\_large\_patch16\_siglip\_512.v2\_webli        & 2025 & \cite{siglip2}           & timm        & ViT-L & vla      & 1024 & \texttt{webli}     & 10B  & 512 & 724 & 25.3 & 20.8 & 37.3 & 31.3 \\

\bottomrule
\end{tabular}
  }
  \caption{\textbf{Benchmarked model details and mAP@1k on \ours and \miniours for global representation models for i2i}. Model details include the year of publication, repository used, architecture (arch), model descriptor dimensions (dims), training scheme (train), training data, and train/test resolution. 5M and 100M correspond to the mini and full versions of the dataset, respectively. For fine-tuned models, only the fine-tuning dataset is considered. Repo indicates the framework used to acquire model weights, \ie torchvision, timm, or official github. $*$ indicates non-publicly available models provided by the original author. $\dagger$ indicates results with the linear adaptation. sup, ssl, dist, vla: supervised learning, self-supervised learning, distillation, vision-language alignment. R50, R101, CN: ResNet50, ResNet101 and ConvNext.
  \label{tab:supp_all_models}
  }
\end{table*}

\newpage

\begin{table*}[t]
  \centering
  \scalebox{0.7}{
    \newcolumntype{R}{>{\raggedleft\arraybackslash}p{2em}}
\setlength{\tabcolsep}{4pt}
\begin{tabular}{llllrrrrrrRR}  \toprule 
\textbf{checkpoint}                               & \textbf{year} & \textbf{cite} & \textbf{repo} & \textbf{arch} & \textbf{dims} & \textbf{dataset}  & \textbf{data size} & \textbf{train res} & \textbf{test res} & \textbf{5M} & \textbf{100M} \\
\midrule
RN50.openai                                       & 2021 & \cite{clip}            & oc      & R50   & 1024 & \texttt{opanai}   & 400M  & 224 & 384 & 2.3  & 1.5  \\
vit\_base\_patch16\_clip\_224.openai              & 2021 & \cite{clip}            & timm+oc & ViT-B & 512  & \texttt{opanai}   & 400M  & 224 & 384 & 2.7  & 1.6  \\
vit\_large\_patch14\_clip\_224.openai             & 2021 & \cite{clip}            & timm+oc & ViT-L & 768  & \texttt{opanai}   & 400M  & 224 & 384 & 6.7  & 4.6  \\
vit\_large\_patch14\_clip\_336.openai             & 2021 & \cite{clip}            & timm+oc & ViT-L & 768  & \texttt{opanai}   & 400M  & 336 & 512 & 8.4  & 5.8  \\
vit\_large\_patch14\_clip\_224.laion2b            & 2021 & \cite{cbw+23,iww+21}   & timm+oc & ViT-L & 768  & \texttt{laion2b}  & 2B    & 224 & 384 & 9.4  & 7.0  \\
convnext\_base.clip\_laion2b\_augreg              & 2022 & \cite{iww+21,convnext} & timm+oc & CN-B  & 640  & \texttt{laion2b}  & 2B    & 256 & 384 & 7.0  & 4.6  \\
convnext\_large\_mlp.clip\_laion2b\_ft\_soup\_320 & 2022 & \cite{iww+21,convnext} & timm+oc & CN-L  & 768  & \texttt{laion2b}  & 2B    & 320 & 512 & 11.5 & 8.1  \\
eva02\_base\_patch16\_clip\_224.merged2b          & 2023 & \cite{fwx+23,evaclip}  & timm+oc & ViT-B & 512  & \texttt{merged2b} & 2B    & 224 & 384 & 4.4  & 2.5  \\
eva02\_large\_patch14\_clip\_336.merged2b         & 2023 & \cite{fwx+23,evaclip}  & timm+oc & ViT-L & 768  & \texttt{merged2b} & 2B    & 336 & 512 & 10.6 & 7.2  \\
vit\_base\_patch16\_siglip\_224.webli             & 2023 & \cite{siglip}          & timm+hf & ViT-B & 768  & \texttt{webli}    & 10B   & 224 & 384 & 10.1 & 7.1  \\
vit\_base\_patch16\_siglip\_256.webli             & 2023 & \cite{siglip}          & timm+hf & ViT-B & 768  & \texttt{webli}    & 10B   & 224 & 384 & 10.3 & 7.5  \\
vit\_base\_patch16\_siglip\_384.webli             & 2023 & \cite{siglip}          & timm+hf & ViT-B & 768  & \texttt{webli}    & 10B   & 384 & 512 & 14.4 & 11.0 \\
vit\_base\_patch16\_siglip\_512.webli             & 2023 & \cite{siglip}          & timm+hf & ViT-B & 768  & \texttt{webli}    & 10B   & 512 & 724 & 14.6 & 11.1 \\
vit\_large\_patch16\_siglip\_256.webli            & 2023 & \cite{siglip}          & timm+hf & ViT-L & 1024 & \texttt{webli}    & 10B   & 256 & 384 & 16.4 & 12.8 \\
vit\_large\_patch16\_siglip\_384.webli            & 2023 & \cite{siglip}          & timm+hf & ViT-L & 1024 & \texttt{webli}    & 10B   & 384 & 512 & 22.2 & 18.1 \\
vit\_base\_patch16\_clip\_224.metaclip\_2pt5b     & 2024 & \cite{metaclip}        & timm+oc & ViT-B & 768  & \texttt{2pt5b}    & 2.5B  & 224 & 384 & 7.6  & 4.9  \\
vit\_large\_patch14\_clip\_224.metaclip\_2pt5b    & 2024 & \cite{metaclip}        & timm+oc & ViT-L & 1024 & \texttt{2pt5b}    & 2.5B  & 224 & 384 & 13.1 & 9.2  \\
vit\_base\_patch16\_siglip\_384.v2\_webli         & 2025 & \cite{siglip2}         & timm+hf & ViT-B & 768  & \texttt{webli}    & 10B   & 384 & 512 & 15.1 & 11.1 \\
vit\_base\_patch16\_siglip\_512.v2\_webli         & 2025 & \cite{siglip2}         & timm+hf & ViT-B & 768  & \texttt{webli}    & 10B   & 512 & 724 & 14.6 & 10.4 \\
vit\_large\_patch16\_siglip\_384.v2\_webli        & 2025 & \cite{siglip2}         & timm+hf & ViT-L & 1024 & \texttt{webli}    & 10B   & 384 & 512 & 23.7 & 18.6 \\
vit\_large\_patch16\_siglip\_512.v2\_webli        & 2025 & \cite{siglip2}         & timm+hf & ViT-L & 1024 & \texttt{webli}    & 10B   & 512 & 724 & 24.7 & 19.8 \\
\bottomrule
\end{tabular}
  }
  \caption{\textbf{Benchmarked model details and mAP@1k on \ours and \miniours for global representation models for t2i}. Model details include the year of publication, repository used, architecture (arch), model descriptor dimensions (dims), training data, and train/test resolution. 5M and 100M correspond to the mini and full versions of the dataset, respectively. Repo indicates the framework used to acquire model weights, \ie timm for the image encoders and huggingface (hf) or OpenCLIP (oc) for the text encoders. R50, CN: ResNet50 and ConvNext.
  \label{tab:supp_all_text_models}
  }
\end{table*}

\newpage

\begin{figure*}[p]
    \centering
    \scalebox{1.8}{
        \begin{tikzpicture}
[define rgb/.code={\definecolor{mycolor}{rbg}{#1}}, rgb color/.style={define rgb={#1},mycolor},
scale=1.]

\newcommand{\arctext}[8]{
    \draw[
        color=white,
        thick,
        line width=1.3pt,
        fill=#2
    ]
    (#5:#4cm+#3) coordinate (above #1) arc (#5:#6:#4cm+#3)
    -- (#6:#4) coordinate (right #1) -- (#6:#4cm-#3) coordinate (below right #1) 
    arc (#6:#5:#4cm-#3) coordinate (below #1)
    -- (#5:#4) coordinate (left #1) -- cycle;

    \pgfmathsetmacro{\angleStart}{#5}
    \pgfmathsetmacro{\angleEnd}{#6}
    \pgfmathsetmacro{\averageAngle}{(\angleStart + \angleEnd) / 2}
    \pgfmathsetmacro{\adjustedAngle}{\averageAngle}

    \path (#5:#4 ) coordinate (start) -- (#6:#4) coordinate (end);
    \path (start) -- (end) coordinate[pos=0.5] (mid);
    \pgfmathsetmacro{\isBold}{#4 < 3 ? 1 : 0}
    \pgfmathsetmacro{\scaleTex}{#4 == 1.3 ? 0.8 : (#4 == 2.7 ? 0.6 : 0.6)}

    \message{Here: #7 \adjustedAngle}
    \pgfmathsetmacro{\rotationAngle}{\adjustedAngle > 90 && \adjustedAngle < 270 ? \adjustedAngle+180: \adjustedAngle}
    \pgfmathsetmacro{\shiftSide}{\adjustedAngle > 90 && \adjustedAngle < 270 ? "west": "east"}
    \pgfmathsetmacro{\shiftAngle}{#4 == 1.3 ? 0.7 : 0.7}
    \pgfmathsetmacro{\angleTitle}{
        #4 == 1.3 && \adjustedAngle > 90 && \adjustedAngle < 270 ? "#7" : (#4 == 1.3 ? "#7": "#7")
    }
    
    \node[anchor=\shiftSide, font=\tiny, rotate=\rotationAngle] 
    at (\adjustedAngle:#4+\shiftAngle)
    {\scalebox{\scaleTex}{\ifnum\isBold=1 \angleTitle \else \angleTitle \fi}};
}

\newcommand{\addsunburstplot}[1]{
    \node[circle, fill=white, align=center, font=\tiny, inner sep=1pt] at (0,0) {\textbf{\ours}};
    \csvreader[head to column names]{#1}{}
    {
        \arctext{\fid}{\fcolor}{20pt}{\fradius}{\astart}{\aend}{\flabel}{\fcount}
    }
}

\addsunburstplot{./data/taxonomy.csv}

\end{tikzpicture}
    }
    \caption{\textbf{The \ours taxonomy} with a 3 level hierarchy.
     The number of objects is displayed for categories with more than 5 objects.
    The taxonomy is used to summarize the objects' diversity and distribution and to report performance per category without affecting the ground truth, which is defined at the instance level.
    \label{fig:taxonomy}
    }
\end{figure*}

\newpage
\begin{figure*}[t]
  \centering
  \scalebox{1.0}{
    \input{fig/fig_dataset_supplementary_i2i}
  }
  \caption{\textbf{Additional examples of queries, positives, and hard negatives within the distractor set based on i2i retrieval.} Average Precision per query and rank of the negatives and positives are reported using SigLIP$^\dagger$. {\color{lightblue}\textbf{Gray:}} queries. {\color{lightgreen}\textbf{Green:}} positives. {\color{lightred}\textbf{Red:}} distractors.
  \label{fig:fig_dataset_supplementary_i2i}
  }
\end{figure*}

\newpage

\begin{figure*}[t]
  \centering
  \hspace{-55pt}
  \scalebox{0.9}{
    \input{fig/fig_dataset_supplementary_t2i}
  }
  \vspace{-10pt}
  \caption{\textbf{Examples of text queries, positives, and hard negatives within the distractor set based on t2i retrieval.} Average Precision per text query, and rank of the negatives and positives is reported using SigLIP. 
  {\color{lightblue}\textbf{Gray:}} text queries. {\color{lightgreen}\textbf{Green:}} positives. {\color{lightred}\textbf{Red:}} distractors.
  \label{fig:fig_dataset_supplementary_t2i}
  }
\end{figure*}

\newpage

\section*{Acknowledgments}

\noindent\textbf{Funding acknowledgement.} This work was supported by the Junior Star GACR (grant no. GM 21-28830M), Horizon MSCA-PF (grant no. 101154126), Programme Johannes Amos Comenius (grant no. CZ.02.01.01/00/22\_010/0003405), CTU in Prague (grant no. SGS23/173/OHK3/3T/13), the CTU institutional support (Future fund). JM was supported by Ministry of the Interior of the Czech Republic (project no. VJ02010041).
We acknowledge VSB – Technical University of Ostrava, IT4Innovations National Supercomputing Center, Czech Republic, for awarding this project access to the LUMI supercomputer, owned by the EuroHPC Joint Undertaking, hosted by CSC (Finland) and the LUMI consortium through the Ministry of Education, Youth and Sports of the Czech Republic through the e-INFRA CZ (grant ID: 90254).

\medskip

\noindent\textbf{Contributors acknowledgments.} We want to thank Larysa Ivashechkina for her work on the annotation of object bounding boxes and masks and the initial filtering of text queries. We also want to thank all contributors for the collection of \ours dataset. Additionally to all authors, the list of external contributor in alphabetical order: Aggeliki Tserota, Anna Nisyraiou, Celeste Abreu, Charalambos Tzamos, Christina Tserota, Dimitris Karageorgiou, Dmytro Mishkin, Eleni Karantali, Eleni Papadopoulou, Eva Tsiliakou, Kelly Kordopati-Zilou, Markos Zampoglou, Noa Garcia, Panagiotis Tassis, Paraskevas Kordopatis, Pavlos Alexantonakis, Ruslan Rozumnyi, Sne\v{z}ana \'{C}urguz, Tereza Nejedl\'{a}, Tom\'{a}\v{s} Jel\'{i}nek, Yannis Kalantidis, Vasilis Alexiadis, Yankun Wu.

{   
    \small
    \bibliographystyle{ieeenat_fullname}
    \bibliography{main}
}

\end{document}